\documentclass[10pt,twocolumn,letterpaper]{article}

\usepackage[pagenumbers]{cvpr} %

\usepackage{times}
\usepackage{epsfig}
\usepackage{graphicx}
\usepackage{amssymb}
\usepackage{booktabs}
\usepackage{multirow}
\usepackage{float}
\usepackage{multicol}
\usepackage{tikz}
\usepackage{comment}
\usepackage{wrapfig}
\usepackage{graphicx}
\usepackage[nopar]{lipsum}
\usepackage{stfloats}
\usepackage{amsmath,amssymb} %
\usepackage{amsfonts}
\usepackage{siunitx}

\usepackage{color}
\usepackage[hypcap=true]{subcaption}
\usepackage[misc]{ifsym}

\usepackage{array}     %
\usepackage{stfloats}
\usepackage{etoolbox}
\usepackage{tocloft}
\usepackage{setspace}

\usepackage{tabularx}  %

\usepackage{xcolor}

\usepackage{afterpage}
\usepackage{stfloats}
\usepackage{lineno}

\usepackage{stfloats}
\usepackage[hypcap=true]{caption}
\usepackage{amssymb}%
\usepackage{pifont}%
\usepackage{subcaption}
\usepackage{enumitem}

\newcommand{\myparagraph}[1]{\vspace{0.75ex}\par\noindent \textbf{#1}}

\newcommand{\notation}[1]{\ensuremath{#1}}
\newcommand{\src}{\notation{s}}
\newcommand{\tgt}{\notation{t}}

\newcommand{\srcpixel}{\notation{n}}
\newcommand{\tgtpixel}{\notation{n}}
\newcommand{\modelname}{Selfi}

\usepackage{dsfont}

\DeclareMathSymbol{@}{\mathord}{letters}{"3B}

\def\latex/{\LaTeX}
\def\bibtex/{\hologo{BibTeX}}

\newcommand{\RN}[1]{%
  \textup{\uppercase\expandafter{\romannumeral#1}}%
}

\definecolor{cvprblue}{rgb}{0.21,0.49,0.74}
\usepackage[pagebackref,breaklinks,colorlinks,allcolors=cvprblue]{hyperref}

\newcommand{\cmark}{\ding{51}}%
\newcommand{\xmark}{\ding{55}}%
\def\paperID{1458} %
\def\confName{CVPR}
\def\confYear{2026}

\title{\emph{Selfi}: Self Improving Reconstruction Engine \\ via 3D Geometric Feature Alignment}

\author{
    Youming Deng\textsuperscript{1,2}%
    \hspace{1cm} 
    Songyou Peng\textsuperscript{2}
    \hspace{1cm} 
    Junyi Zhang\textsuperscript{2,3} \hspace{1cm} 
    Kathryn Heal\textsuperscript{2}  \and
    Tiancheng Sun\textsuperscript{2} \hspace{1cm} 
    John Flynn\textsuperscript{2} \hspace{1cm} 
    Steve Marschner\textsuperscript{1} \hspace{1cm}
    Lucy Chai\textsuperscript{2} \and
    \textsuperscript{1}Cornell University \hspace{1cm} 
    \textsuperscript{2}Google \hspace{1cm} 
    \textsuperscript{3}UC Berkeley
}

\begin{document}

\twocolumn[{%
\renewcommand\twocolumn[1][]{#1}%
\maketitle
\begin{center}
\vspace{-6mm}
    \includegraphics[width=0.99\textwidth]{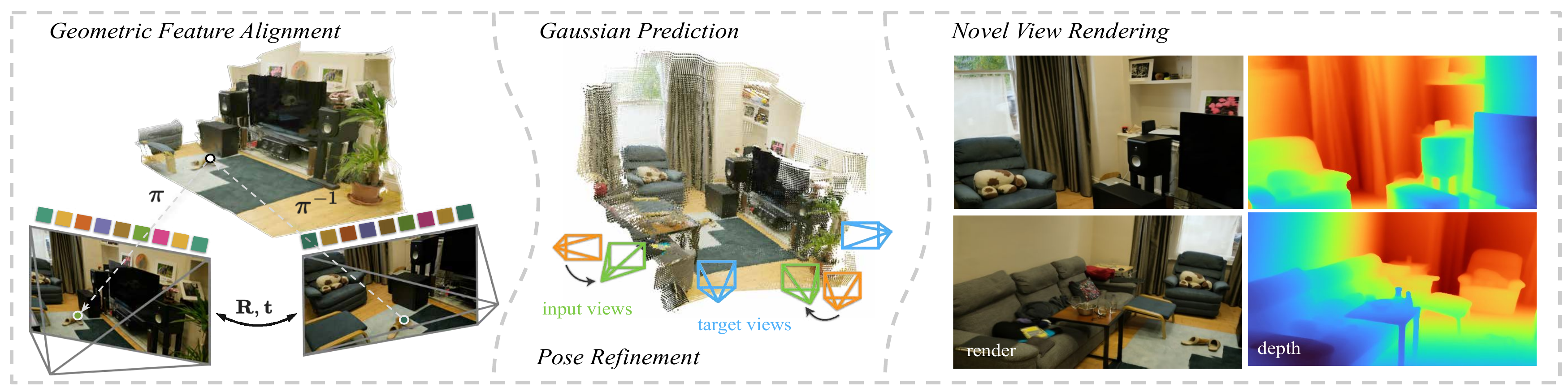}
    \vspace{-2mm}
    \captionof{figure}{\textbf{Self Improving Reconstruction Engine.} We introduce \emph{\modelname}, a self-improving pipeline for novel view synthesis from unposed images. 
    We start by learning geometrically aligned features using consistency losses and self-labelled pseudo ground truths from a 3D foundation model (\textit{e.g.,} VGGT~\cite{wang2025vggt}). 
    These features can be used to predict Gaussian primitives~\cite{kerbl20233d}, and also refine initial poses via bundle adjustment.
    The improved poses are used to further adjust the initial 3D representation, resulting in an even higher quality final rendering.
    }
    \label{fig:teaser}
\end{center}%
}]

\renewcommand{\thefootnote}{\fnsymbol{footnote}}

\begin{abstract}

\vspace{-3mm}
Novel View Synthesis (NVS) has traditionally relied on models with explicit 3D inductive biases combined with known camera parameters from Structure-from-Motion (SfM) beforehand. 
Recent vision foundation models like VGGT take an orthogonal approach -- 3D knowledge is gained implicitly through training data and loss objectives, enabling feed-forward prediction of both camera parameters and 3D representations directly from a set of uncalibrated images. 
While flexible, VGGT features lack explicit multi-view geometric consistency, and we find that improving such 3D feature consistency benefits both NVS and pose estimation tasks. 
We introduce \modelname, a self-improving 3D reconstruction pipeline via feature alignment, transforming a VGGT backbone into a high-fidelity 3D reconstruction engine by leveraging its own  outputs as pseudo-ground-truth. 
Specifically, we train a lightweight feature adapter using a reprojection-based consistency loss, which distills VGGT outputs into a new geometrically-aligned feature space that captures spatial proximity in 3D.
This enables state-of-the-art performance in both NVS and camera pose estimation, demonstrating the benefits of feature alignment for downstream 3D reasoning.
More details on our project page: \url{https://denghilbert.github.io/selfi}

\end{abstract}

\section{Introduction}
Novel view synthesis (NVS) has long relied on known camera parameters or those recovered from an SfM-first pipeline: detect keypoints, match them, and solve for cameras before optimizing a scene representation for rendering~\cite{schoenberger2016sfm,pan2024glomap,schoenberger2016mvs,mildenhall2021nerf,kerbl20233d}. 
While effective, this decoupling between the cameras and scene representation is not only computationally intensive but also fragile -- NVS quality is highly dependent on the accuracy of SfM poses.
Feed-forward NVS methods remove per-scene optimization and move toward direct prediction. Given calibrated images, they extract image features and lift them to 3D primitives or pixels in one forward pass~\cite{flynn2016deepstereo, kalantari2016learning, zhou2018stereo,flynn2019deepview,yu2021pixelnerf,wang2021ibrnet,szymanowicz2024splatter,xu2025resplat,ziwen2024long,xu2024grm,
chen2024mvsplat,jin2024lvsm}. However, most approaches still assume known cameras from SfM, so the quality also degrades when calibration is inaccurate or even fails without SfM estimation.

The advent of 3D Vision Foundation Models (VFMs) \cite{wang2024dust3r,mast3r_eccv24,wang2025vggt} has offered a paradigm shift. Trained on vast, diverse datasets with 3D annotations, these models can predict camera poses, dense depth, and 3D structure from uncalibrated images in a single forward pass, effectively bypassing SfM. 
A promising recent direction is to leverage these VFMs for NVS by directly decoding VFM features into 3D representations like 3D Gaussians~\cite{jiang2025anysplat,ye2024no}. However, this approach suffers from a significant drawback: the resulting NVS quality is substantially lower than that of optimization-based methods. We hypothesize this is because VFM features, while powerful for the geometric prediction tasks they were trained on, are not explicitly optimized to be geometrically consistent across different views, which is crucial for high-fidelity NVS.

This paper overcomes that limitation: we propose a surprisingly simple feature alignment strategy to transform a pre-trained 3D VFM into a state-of-the-art NVS and pose estimation engine \emph{without using any 3D ground-truth annotations}. Our key insight is to leverage the VFM's own outputs as a dense, self-supervised signal to learn a new, geometrically-aligned feature space.
We freeze VGGT as a backbone foundation model and train a lightweight feature adapter by constructing a self-supervised task.
Using pseudo-ground-truth depth and cameras from VGGT, we reproject query points from one view to other views to serve as a correspondence signal. The feature adapter head produces per-pixel features for both images; we then enforce a simple reprojection-based feature consistency loss between features at their corresponding locations.
This yields geometrically aligned features where feature similarity captures both semantic content and proximity in 3D space without any camera annotations or 3D supervision beyond the model outputs themselves. 

The resulting features are powerful and versatile.
First, when used to predict parameters for 3D Gaussian Splatting~\cite{kerbl20233d}, our learned features achieve state-of-the-art NVS quality from unposed images, dramatically outperforming methods that use the original VGGT features.
Second, these geometrically-aligned features help establish robust correspondences, allowing us to achieve better performance on pose estimation with a few extra bundle adjustment (BA) steps.
Both of these are achieved in a fully self-supervised manner. In summary, our contributions are:
\begin{itemize}
\item \textbf{Self-improving geometric feature learning from unposed RGB}. We introduce a reprojection-based feature consistency loss to learn geometrically aligned 3D features, without any annotations.
\item \textbf{State-of-the-art NVS and pose estimation results with a frozen backbone}. Despite training only small heads on top of a frozen VFM, we set a new SOTA on unposed NVS and pose estimation benchmarks.
\end{itemize}

\section{Related Work}
\label{sec:related_work}
\myparagraph{3D Reconstruction and Pose Estimation.} 
Earlier 3D reconstruction pipelines rely on decoupled two-stage processes, with Structure-from-Motoin (SfM) for geometry and cameras followed by Multi-view Stereo (MVS) for depth
~\cite{schoenberger2016sfm,pan2024glomap,detone2018superpoint,sarlin2020superglue,sarlin2021back,yao2018mvsnet,karaev2024cotracker,schoenberger2016mvs}.
While these methods are grounded in multi-view geometry~\cite{hartley2003multiple}, they are slow and lack robustness. More recent learning-based approaches use diffusion models~\cite{wang2023posediffusion,zhao2025diffusionsfm,zhang2024cameras} and transformers~\cite{wang2023pf,vaswani2017attention,wang2024vggsfm}
to directly infer geometric quantities like camera pose, depth, and 3D point maps in an end-to-end manner~\cite{wang2025continuous,murai2025mast3r,wang20243d,wang2025pi,wang2025moge,wang2024dust3r,mast3r_eccv24,yang2025fast3r,zhang2024monst3r}.
These 3D Vision Foundation Models (VFMs), trained on large, diverse datasets, have demonstrated impressive generalization capabilities. 
Our method builds on VGGT~\cite{wang2025vggt}, a VFM designed to jointly predict all necessary geometric quantities, which we use as our geometric pseudo-label teacher.

\myparagraph{Feed-Forward Novel View Synthesis.} 
NVS aims to generate realistic images from arbitrary viewpoints. 
One class of NVS methods, such as those built on NeRF~\cite{mildenhall2021nerf}, 3D Gaussian Splatting (3DGS)~\cite{kerbl20233d}, or voxels~\cite{fridovich2022plenoxels,sun2022direct,muller2022instant}, optimize a representation individually per scene and typically require calibrated inputs from SfM or a fixed camera rig~\cite{xu2023vr,broxton2020immersive,lin2021deep,flynn2019deepview}.
In contrast, generalizable NVS approaches %
train a network to directly regress a 3D scene representation (e.g., volumetric fields or Gaussian parameters) from one or more input images in a feed-forward fashion, but many still assume known camera parameters. 
These methods may incorporate inductive biases like cost volumes~\cite{chen2024mvsplat,flynn2016deepstereo}, depths~\cite{xu2025depthsplat}, epipolar constraints~\cite{wang2021ibrnet,suhail2022generalizable}, or multi-plane images~\cite{zhou2018stereo,flynn2016deepstereo,flynn2024quark,tucker2020single}. More recently some methods directly feed the inputs into a general-purpose transformer architecture~\cite{xu2024grm,zhang2024gs}, resulting in a renderable representation~\cite{yu2021pixelnerf,charatan2024pixelsplat,szymanowicz2024splatter,szymanowicz2025flash3d,tang2024lgm}.
Variations of this task aim to reconstruct entire scenes in a single inference pass rather than localized viewing angles~\cite{ziwen2024long,imtiaz2025lvt}. Given the recent rise in models that also predict camera poses or pointmaps in a feed-forward fashion, several works jointly regress Gaussian splat parameters with cameras and geometry, allowing for NVS from unposed images~\cite{zhang2025flare,jiang2025anysplat,ye2024no}. An orthogonal approach taken by LVSM~\cite{jin2024lvsm} and Rayzer~\cite{jiang2025rayzer} avoids a 3D representation entirely, and instead directly outputs the rendered image. Our approach follows the per-pixel Gaussian splat parameterization; we train a lightweight adapter on top of the features derived from VGGT that yields an explicit 3D scene representation for rendering to new camera viewpoints. 

\myparagraph{Vision Foundation Models as a Feature Backbone.}
Feature representations learned from large-scale VFMs like DINO, CLIP, and more~\cite{oquab2023dinov2,simeoni2025dinov3,dosovitskiy2020image,radford2021learning} have proven invaluable for various downstream 3D tasks, including correspondence matching and depth estimation~\cite{bochkovskii2024depth,yang2024depth,wang2024dust3r,mast3r_eccv24,wang2025vggt}. 
Prior works have also leveraged diffusion features for semantic correspondences~\cite{zhang2023tale,tang2023emergent,hamilton2022unsupervised,luo2023dhf}, while Feat2GS~\cite{chen2025feat2gs} probes the representation space of various visual foundation models using NVS as a proxy task for 3D understanding. 
However, these features are primarily semantic and lack the dense geometric consistency required for high-fidelity 3D reconstruction.
Rather than simply reusing existing VFM features, we use features from VGGT to learn a geometrically aligned representation space, and we demonstrate that these features not only benefit NVS, but can be used to refine predicted poses and achieve even better rendering.

\begin{figure*}[t]
    \centering
    \includegraphics[width=\textwidth]{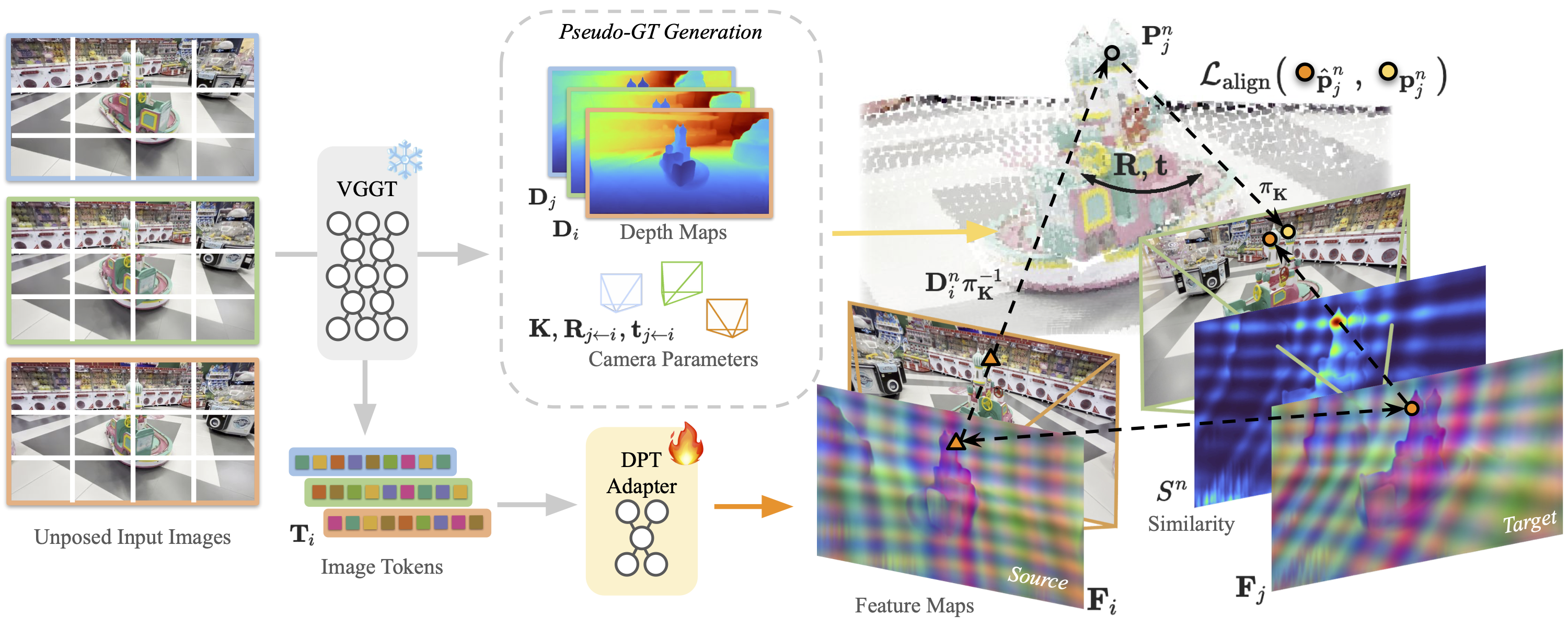}
    \caption{\textbf{Geometric Feature Alignment with Self-Labeled Pseudo-Ground-Truth}. Using a pretrained VGGT~\cite{wang2025vggt} backbone, we use predicted depth and camera parameters as pseudo-ground-truth to align features obtained from a DPT adapter on top of VGGT image tokens. We sample query points and reproject these points to a target view using depth and camera parameters. Our loss function encourages the features at these two corresponding locations from source and target frames to be similar.
    }
    \label{fig:feature_alignment_w_pseudo_gt}
\end{figure*}

\section{Method}
\label{sec:method}
In this section, we first introduce our training strategy for geometric feature alignment to obtain a powerful 3D-aware representation in~\cref{sec:feature_alignment}.
Next, we feed these features into an efficient 3D Gaussian predictor for Novel View Synthesis (NVS) in~\cref{sec:forward_gs}. Finally, we leverage our features for matching and dense bundle adjustment (BA) to further improve pose estimation and novel view synthesis in~\cref{sec:dense_ba}.

\subsection{Geometric Feature Alignment}
\label{sec:feature_alignment}
While NVS methods with explicit 3D inductive biases~\cite{mildenhall2021nerf,kerbl20233d,flynn2019deepview,flynn2024quark,chen2024mvsplat,yu2021pixelnerf} are able to maintain geometric and appearance consistency when rendering to novel views by design, this property is not guaranteed in transformer-based foundation models. Since 3D VFMs like DUSt3R~\cite{wang2024dust3r} and VGGT~\cite{wang2025vggt} learn 3D priors implicitly, their feature spaces are not natively multi-view consistent. To bridge this gap, we propose learning a geometrically aligned feature space as a prerequisite for high-fidelity NVS~\cite{ye2024no,jiang2025anysplat,liu2025worldmirror}.

\myparagraph{Feature Correspondence Prediction.} 
To produce our aligned features, we take the feature backbone and aggregator from pre-trained VGGT~\cite{wang2025vggt} and append a spatial feature adapter DPT~\cite{ranftl2021vision}:
\begin{equation}
    \mathbf{F}_i=\text{DPT}_\text{adapter}(\mathbf{T}_i),
\end{equation}
where image tokens $\mathbf{T}_{i} \in \mathbb{R}^{4\times(H_p \times W_p) \times D}$ are taken from four intermediate feature maps from the VGGT backbone with $H_p$, $W_p$, and $D$ as the feature spatial resolution and feature dimension. $\mathbf{F}_i \in \mathbb{R}^{H\times W\times C}$ is the output feature map for an arbitrary frame $i$, with $H$ and $W$ the same as the output image resolution, and $C=24$.

The goal of our feature alignment is to train the feature adapter such that features corresponding to spatially proximal 3D locations exhibit high similarity. %
Given a sampled query location $n$, source frame $s$, and target frame $t$, we take the pixel-aligned feature $\mathbf{F}_{\src}^{\srcpixel} \in \mathbb{R}^C$ from the 2D source feature map $\mathbf{F}_{\src}$, and the 2D target feature map $\mathbf{F}_{\tgt}$. We compute a similarity map in~\cref{eq:similarity_map} and apply a softmax with temperature $\tau$ in ~\cref{eq:softmax_norm}:
\begin{gather}
\label{eq:similarity_map}
S^{\srcpixel}(u,v) = \frac{ \mathbf{F}_{\src}^{\srcpixel} \cdot \mathbf{F}_{\tgt}(u,v) }{ \| \mathbf{F}_{\src}^{\srcpixel} \|_2 \, \| \mathbf{F}_{\tgt}(u,v) \|_2 }, \\
\label{eq:softmax_norm}
w^{\srcpixel}(u,v) = \frac{ \exp(S^{\srcpixel}(u,v)/\tau) }{ \sum_{u',v'} \exp(S^{\srcpixel}(u',v')/\tau) }.
\end{gather}

The predicted 2D correspondence $\hat{\mathbf{p}}_{\tgt}^{\tgtpixel}$ (shown as an orange dot \textcolor{orange}{$\bullet$} in~\cref{fig:feature_alignment_w_pseudo_gt}) is obtained as a weighted average over the $(u,v)$ target coordinates. 
\begin{equation}
\label{eq:predicted_correspondence}
\hat{\mathbf{p}}_{\tgt}^{\tgtpixel} = \sum_{u,v} w^{\srcpixel}(u,v) \,[u, v],\\
\end{equation}

We find that using a weighted average over all pixels enables the model to leverage information from all target features during alignment, providing denser supervision compared to contrastive training methods~\cite{mast3r_eccv24}. 

\myparagraph{Pseudo-Ground-Truth Supervision from VGGT.} 
To supervise the predicted correspondence, we establish pseudo-ground-truth 2D-2D correspondences using VGGT.
Specifically, given a set of images, we run VGGT~\cite{wang2025vggt} to obtain per-frame depth maps $\mathbf{D}_i$, shared camera intrinsics $\mathbf{K}$, and relative poses $[\mathbf{R}_{j\leftarrow i}|\mathbf{t}_{j\leftarrow i}]$. 
For a query pixel $\mathbf{p}_{\src}^{\srcpixel}$ in source frame $\src$, we compute its corresponding location in a target frame $\tgt$ via 3D unprojection and reprojection.  We unproject the pixel using its depth $\mathbf{D}_{\src}^{\srcpixel}$ to 3D space (orange triangle {\color{orange}$\blacktriangle$} in~\cref{fig:feature_alignment_w_pseudo_gt}), transform it to the target coordinate system, and project it back to 2D to obtain the pseudo-GT correspondence $\mathbf{p}_{\tgt}^{\tgtpixel}$ (yellow dot {\color{Goldenrod} $\bullet$} in~\cref{fig:feature_alignment_w_pseudo_gt}):
\begin{gather}\label{eq:pseudogt}
\mathbf{P}_{\tgt}^\tgtpixel = \mathbf{R}_{\tgt \leftarrow \src} \mathbf{D}^{\srcpixel}_{\src}\pi_{\mathbf{K}}^{-1} \mathbf{p}_{\src}^{\srcpixel} + \mathbf{t}_{\tgt\leftarrow \src},\\
\mathbf{p}_{\tgt}^{\tgtpixel} = \pi_\mathbf{K}   \mathbf{P}_{\tgt}^{\tgtpixel} ,
\end{gather}
where $\pi_\mathbf{K}$ denotes the projection from 3D camera coordinates to the 2D pixel space with intrinsic $\mathbf{K}$, and $\pi_{\mathbf{K}}^{-1}$ denotes the inverse projection. 
To handle possible occlusions during reprojection, we also maintain a hard visibility map $\mathbf{V}_{\tgt}^{\tgtpixel}$, filtering by the difference between the z-coordinate of the 3D point backprojected from the source view, and the target-view depth map at the corresponding re-projected location (which we refer to as $\mathbf{D}_{\tgt}^{\tgtpixel}$, the value of $\mathbf{D}_{\tgt}$ at pixel location $\mathbf{p}_{\tgt}^{\tgtpixel}$):
\begin{equation}\label{eq:occlusion}
    \mathbf{V}_{\tgt}^{\tgtpixel} = \left[ \left| \mathbf{P}_{\tgt}^{\tgtpixel} \cdot [0\;0\;1]^T - \mathbf{D}_{\tgt}^{\tgtpixel} \right| < \alpha \right],
\end{equation}
where $[\cdot]$ denotes the Iverson bracket.%

\myparagraph{Objective for Alignment.}
Given our predicted correspondence $\hat{\mathbf{p}}_{\tgt}^{\tgtpixel}$ and reprojection-based pseudo-ground truth $\mathbf{p}_{\tgt}^{\tgtpixel}$, we optimize the features to produce high similarity at the corresponding points, weighted by visibility:
\begin{equation}
\label{eq:dense_loss}
    \mathcal{L}_{\text{align}} = 
    \frac{1}{TN}\sum_{t=1}^{T}\sum_{n=1}^{N} 
    \mathbf{V}_{\tgt}^{\tgtpixel}\big\| \hat{\mathbf{p}}_{\tgt}^{\tgtpixel} - \mathbf{p}_{\tgt}^{\tgtpixel} \big\|_2^2.
\end{equation}
We demonstrate in our baseline experiments~\cref{sec:exp_nvs} and ablations~\cref{sec:exp_ablation} that our alignment strategy benefits downstream NVS performance. Moreover, we show that our aligned features can further refine the initial camera poses from VGGT via bundle adjustment, despite training on pseudo-GT with zero projection error in~\cref{sec:exp_pose}.

\subsection{Feed-Forward Gaussian Prediction}
\label{sec:forward_gs}

Using the aligned feature maps from the previous step, we now predict 3D Gaussian parameters~\cite{kerbl20233d} to enable feed-forward novel view rendering. We freeze the $\text{DPT}_\text{adapter}$ feature head and train a new U-Net~\cite{ronneberger2015u} decoder. This U-Net takes the aligned feature $\mathbf{F}_s$ concatenated with the source input image $I_s$, and predicts the parameters for each Gaussian primitive relative to the source camera coordinates.
Specifically, the decoder outputs: quaternions $\mathbf{q}_s$, scales $\mathbf{s}_s$, color $\mathbf{c}_s$, opacity $\boldsymbol{\sigma}_s$, and a depth residual $\Delta\mathbf{D}_s$ to be added to the initial depth map  $\mathbf{D}_s$:
\begin{align}
    \mathbf{F}_s^{\text{dec}}&=\text{U-Net}(\text{cat}(\mathbf{F}_s, I_s)), \\
    \mathbf{q}_s, \Delta\mathbf{D}_s&=\text{Conv}_{q}(\mathbf{F}_s^{\text{dec}}), \text{Conv}_{\mathbf{D}}(\mathbf{F}_s^{\text{dec}}), \\
    \{\boldsymbol{\sigma}_s,\mathbf{s}_s,\mathbf{c}_s\}&=\text{MLP}(\mathbf{F}_s^{\text{dec}}),
\end{align}
For the 3D positions of the Gaussians, we add the predicted depth residual to the initial depth map and unproject the 2D pixels into 3D locations $\boldsymbol{\mu}_s$ relative to the source images' coordinate frames:
\begin{align}
    \boldsymbol{\mu}_s&=(\mathbf{D}_s+\Delta\mathbf{D}_s)\pi_{\mathbf{K}}^{-1}\mathbf{p}_s.
    \label{eq:unproj}
\end{align}
We predict colors $\mathbf{c}_s$ using spherical harmonics coefficients to model view-dependent effects~\cite{fridovich2022plenoxels,kerbl20233d}. 
One key modification in our work is that we also enable spherical harmonics on the density $\boldsymbol{\sigma}_s$~\cite{imtiaz2025lvt}, rather than using a single scalar density. 
As the geometry prediction from VGGT may not be fully accurate, in particular in low-confidence regions, 
we find this view-dependent density crucial for overcoming possible occlusions and misalignments. 
In effect, the density spherical harmonics serves as a learned confidence metric; for any given render viewpoint, it modulates opacity so that unreliable Gaussians become transparent. 
This also enables us to prune the low-confidence Gaussians for rendering efficiency, a distinct approach from the voxelization pruning used by AnySplat~\cite{jiang2025anysplat} and WorldMirror~\cite{liu2025worldmirror}. We quantify this design choice in our ablation (\cref{sec:exp_ablation} and~\cref{tab:ablation_small_table}) and visualize the learned densities in~\cref{fig:view_dependent}.

Finally, the collection of 3D Gaussian parameters produced from all source images $s$ are then rasterized to obtain target renderings $\hat{I}_t$ using the predicted relative camera parameters $\mathbf{C}_{t\leftarrow s}$.
The model is trained solely with an RGB reconstruction loss:
\begin{gather}
    \hat{I}_t = \text{Rasterizer}(\{(\boldsymbol{\sigma}_s,\mathbf{s}_s,\mathbf{c}_s,\boldsymbol{\mu}_s,\mathbf{q}_s, \mathbf{C}_{t\leftarrow s})\}_{s=1}^S),\\
    \mathcal{L}_{\text{RGB}} =\frac{1}{T}\sum_{t=1}^{T}
    \big\| \hat{I}_t -I_t \big\|.
\end{gather}

\subsection{Dense Bundle Adjustment with Depth Shift}
\label{sec:dense_ba}

Compared to other feed-forward 3DGS methods~\cite{jiang2025anysplat,ye2024no,liu2025worldmirror} that perform post-optimization on both camera parameters and 3D Gaussians, our aligned features offer a more classic and efficient alternative to improve camera parameters with a quickly-converging bundle adjustment (BA).

\begin{figure}[t]
    \centering
    \setlength{\tabcolsep}{1pt}
    \resizebox{\columnwidth}{!}{%
    \begin{tabular}{ccc}
        \includegraphics[height=0.22\columnwidth]{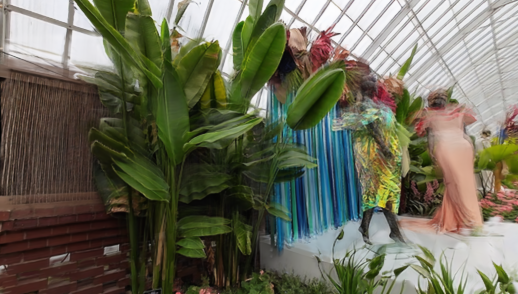} &
        \includegraphics[height=0.22\columnwidth]{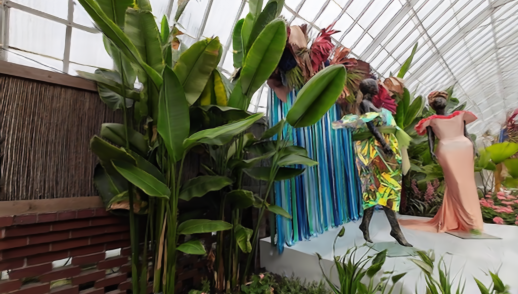} &
        \multirow[t]{2}{*}{%
          \centering
          \raisebox{-0.24\columnwidth}[\dimexpr.44\columnwidth\relax][0pt]{%
            \includegraphics[height=.44\columnwidth]{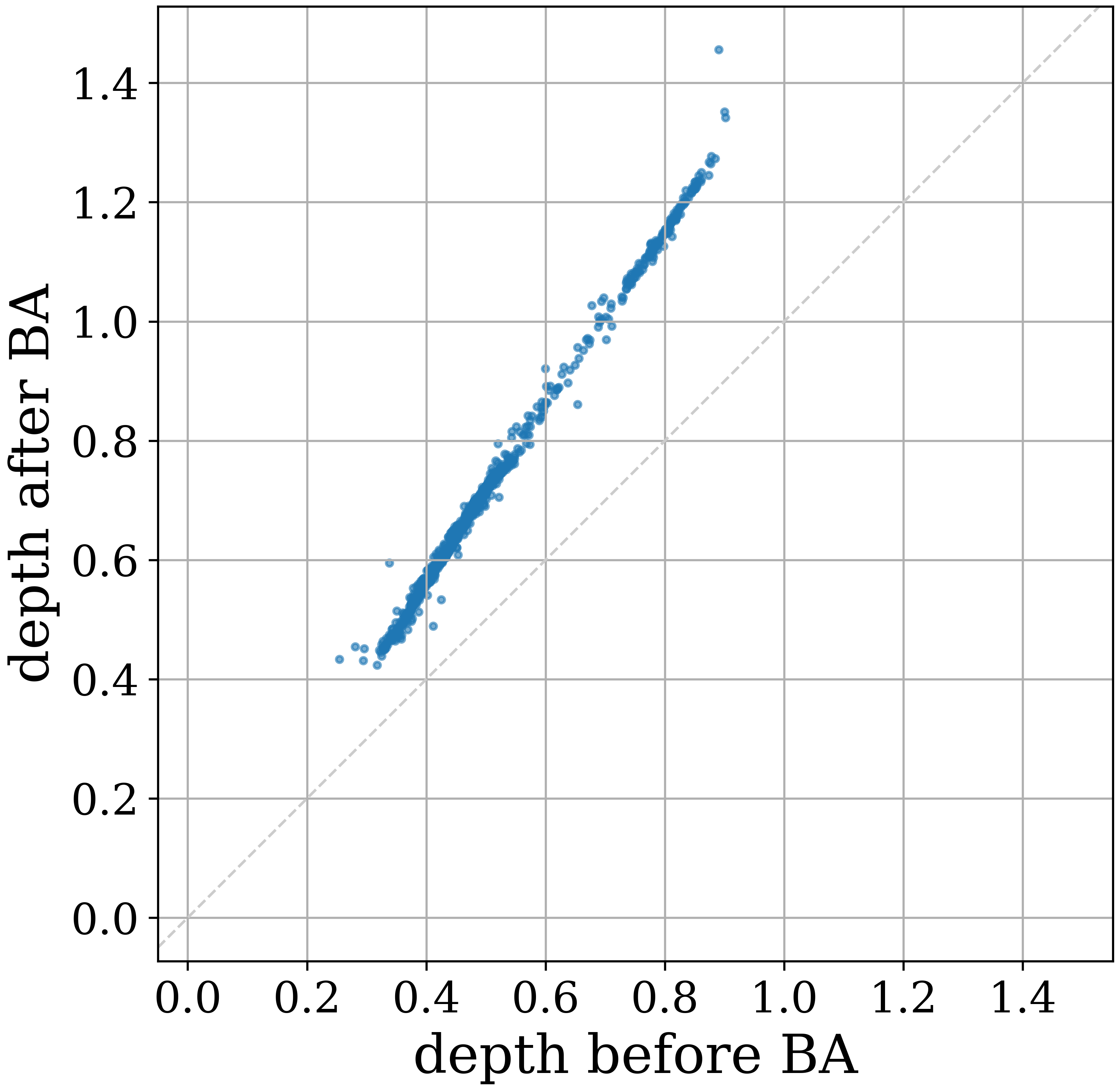}%
          }%
        }\\
        \includegraphics[height=0.22\columnwidth]{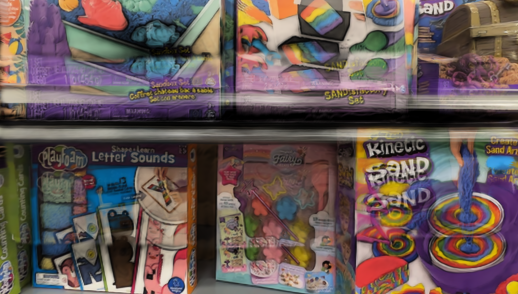} &
        \includegraphics[height=0.22\columnwidth]{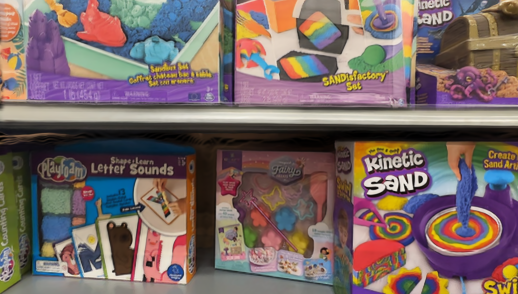} &
        \\

        (a) w/o Depth Shift \phantomsubcaption\label{fig:depth_shift_a} &
        (b) w/ Depth Shift \phantomsubcaption\label{fig:depth_shift_b} &
        (c) Shift Function \phantomsubcaption\label{fig:depth_shift_c} \\
    \end{tabular}%
    }
    \caption{\textbf{Bundle Adjustment with Depth Shift.} (a) After refining the camera poses with bundle adjustment, naively rendering the predicted Gaussian primitives with the new poses results in misalignment. (b) Propagating the adjustments in sparse 3D points during BA to the dense depth maps results in improved rendering. (c) We plot the sparse point depths before and after BA, and observe that a linear fit suffices for this adjustment.
    }
    \label{fig:ablation_depth_shift}
\end{figure}

While BA using correspondences from our aligned features improves the estimated poses (\cref{sec:exp_pose}), it also changes the position of sparse 3D points associated with 2D correspondences. This requires us to also adjust the Gaussian centers to be consistent with this new geometry. 
We find that, because the gradient from BA may change the estimated intrinsics, the 3D points will move along each camera's z-axis (in the depth direction) to compensate. Ignoring this change in depth before rasterizing the Gaussian primitives leads to rendering misalignment shown in~\cref{fig:depth_shift_a}.

We observe that the change in depth is primarily a linear function, as visualized in~\cref{fig:depth_shift_c}. Therefore, it can be easily modeled with an affine transformation $\phi(\cdot)$ estimated from the sparse BA points and applied to all densely predicted Gaussians. We apply $\phi(\cdot)$ to the per-frame depth maps $\mathbf{D}_s + \Delta\mathbf{D}_s$, with Gaussian scales adjusted accordingly:
\begin{align}
    \boldsymbol{\mu}_s'&=\phi(\mathbf{D}_s+\Delta\mathbf{D}_s)\pi_{\mathbf{K}'}^{-1}\mathbf{p}_s\\
    \mathbf{s}_s' &= \frac{\phi(\mathbf{D}_s + \Delta\mathbf{D}_s)}{\mathbf{D}_s + \Delta\mathbf{D}_s}\cdot \mathbf{s}_s.
\end{align}
where $\boldsymbol{\mu}_s'$ is the new 3D position scaled by the new affine-corrected depth and $\mathbf{K}'$ is the new intrinsics. The scale $\mathbf{s}_s'$ is adjusted proportionally. As shown in~\cref{fig:depth_shift_b} and in the experiments, this simple correction successfully bridges the geometric gap, allowing for pose improvements from BA and enhancing NVS quality.

\begin{figure*}[t]
    \centering
    \setlength{\tabcolsep}{1pt}
    \resizebox{\textwidth}{!}{
    \begin{tabular}{cccc}
    \includegraphics[width=0.25\textwidth]{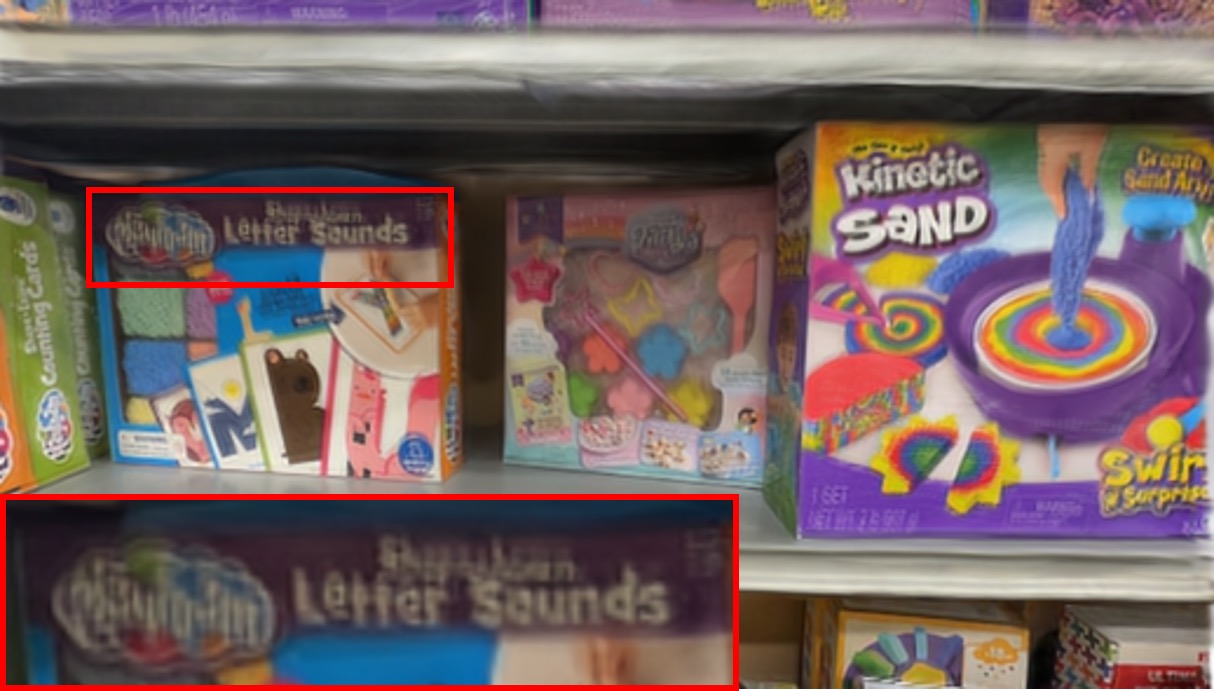} &
    \includegraphics[width=0.25\textwidth]{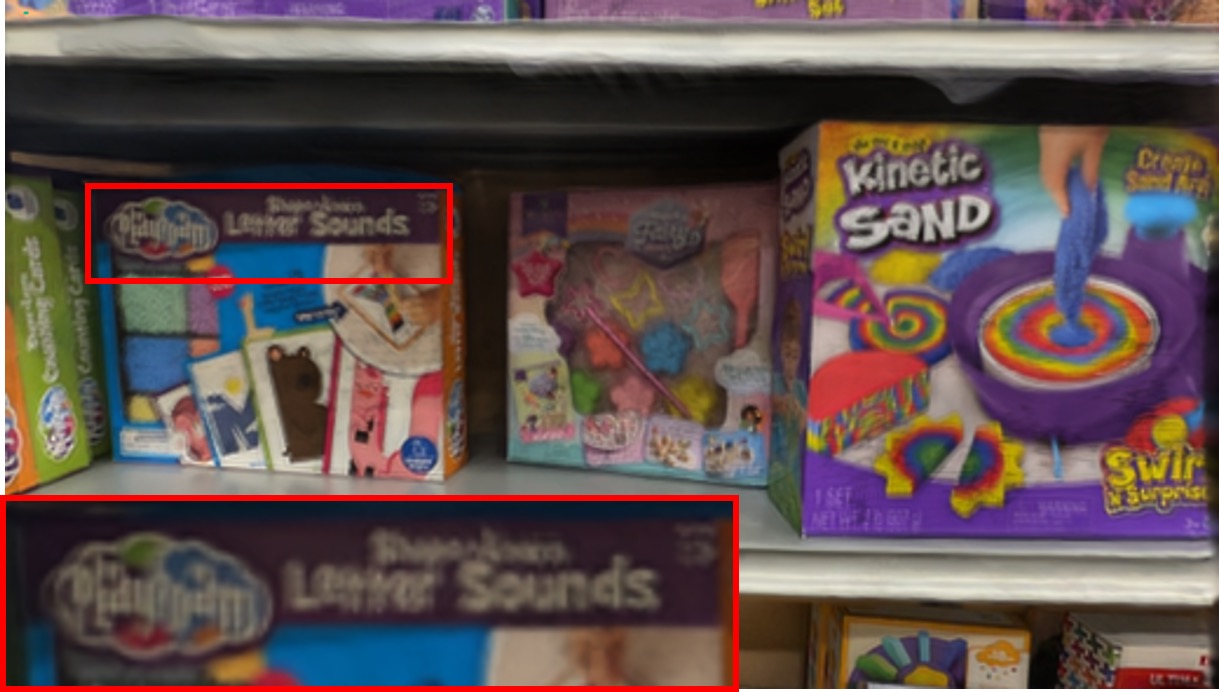} &
    \includegraphics[width=0.25\textwidth]{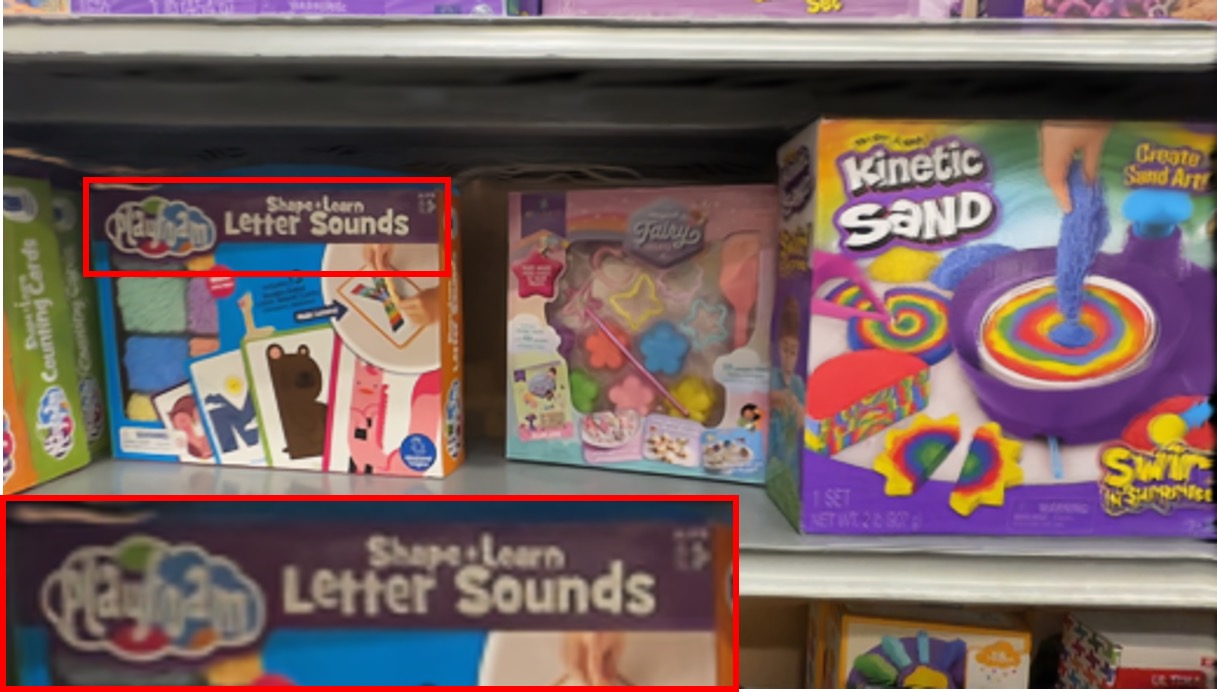} &
    \includegraphics[width=0.25\textwidth]{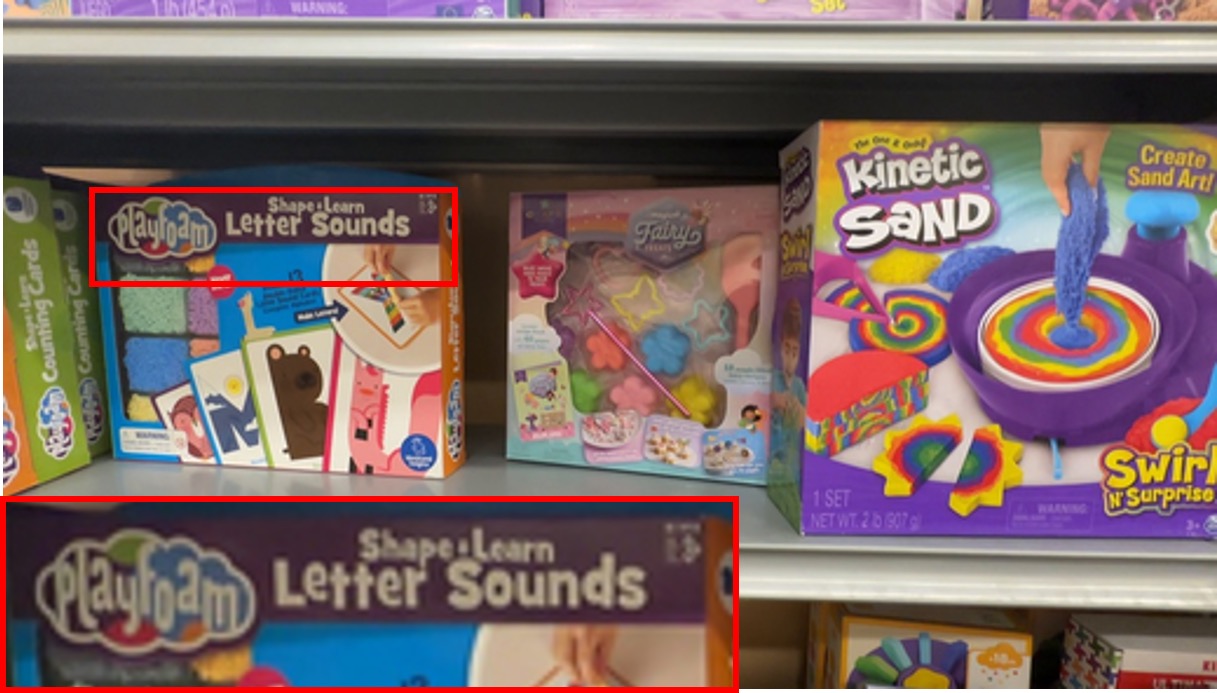}\\
    \includegraphics[width=0.25\textwidth]{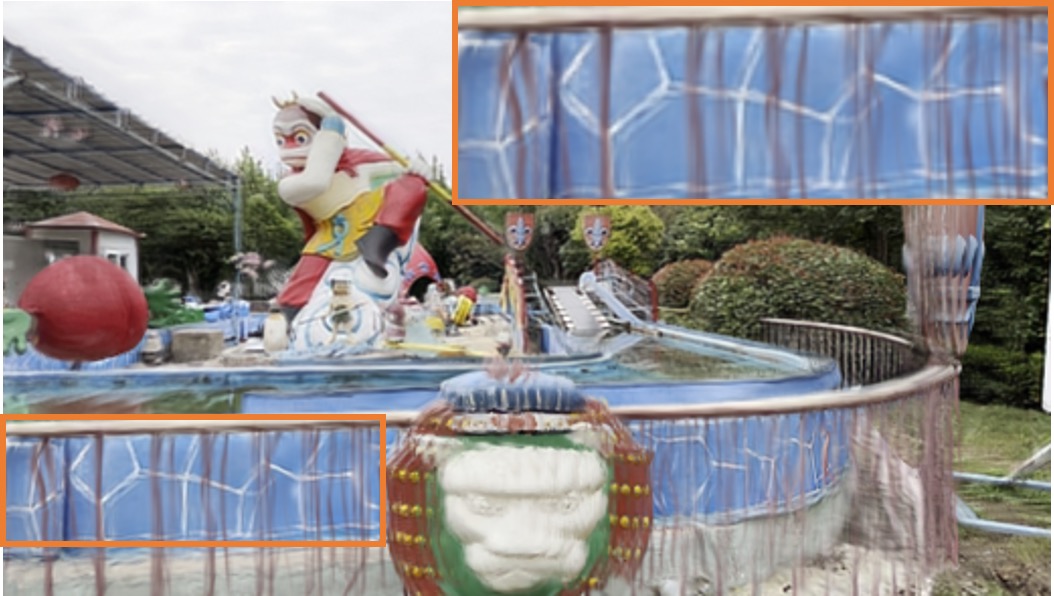} &
    \includegraphics[width=0.25\textwidth]{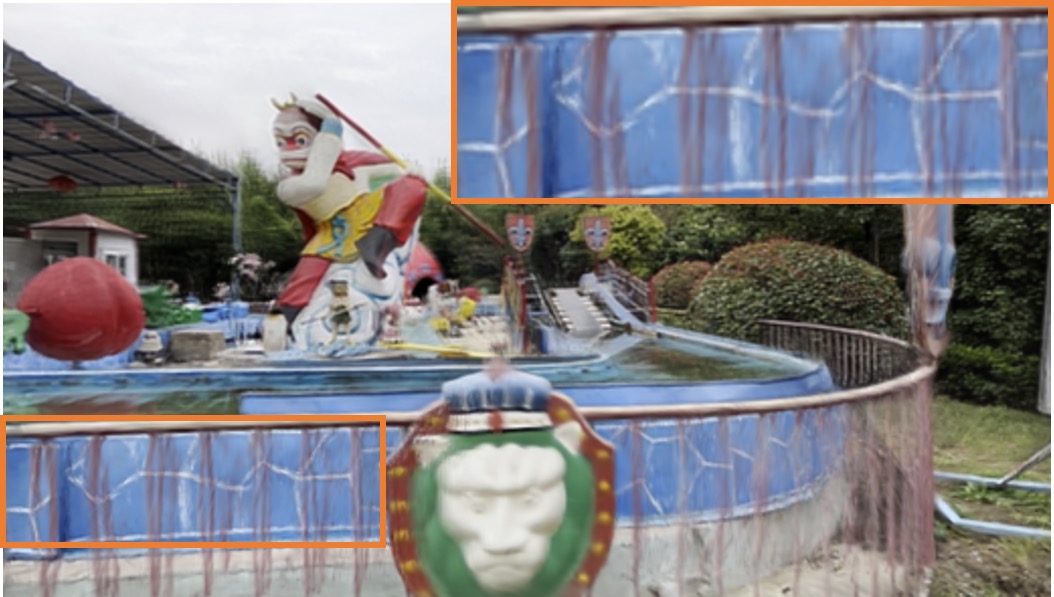} &
    \includegraphics[width=0.25\textwidth]{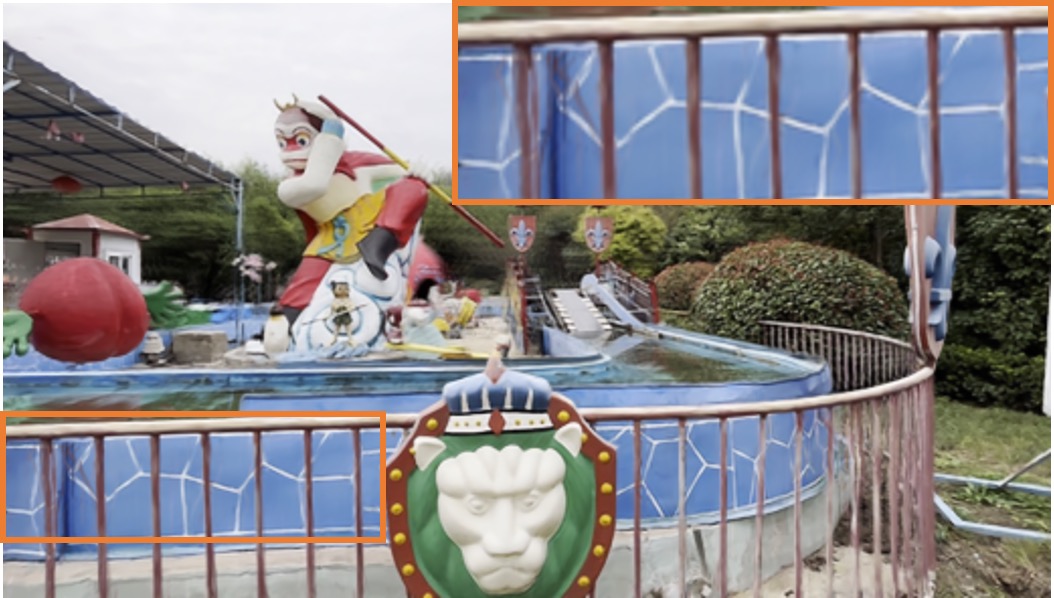} &
    \includegraphics[width=0.25\textwidth]{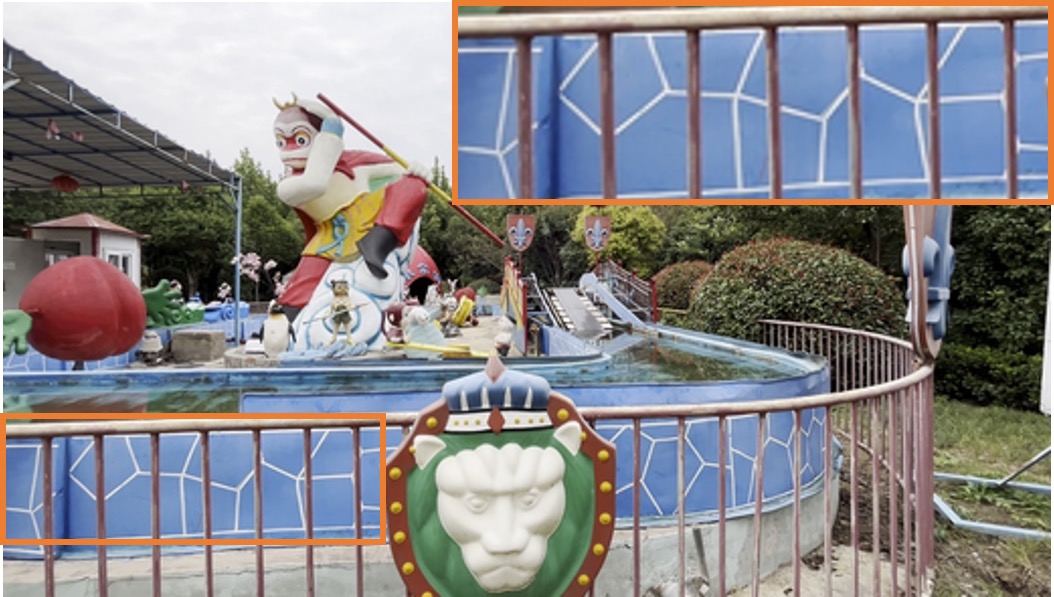}\\
    \includegraphics[width=0.25\textwidth]{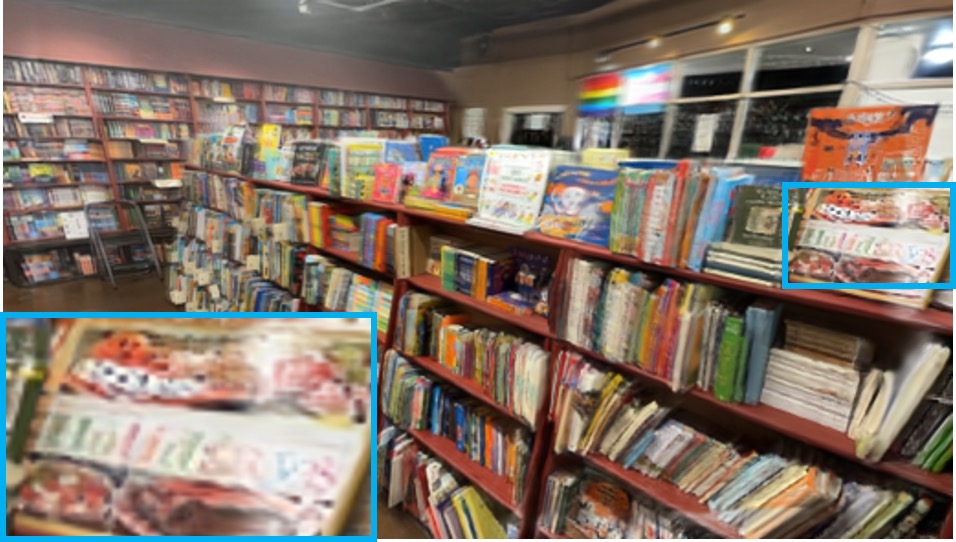} &
    \includegraphics[width=0.25\textwidth]{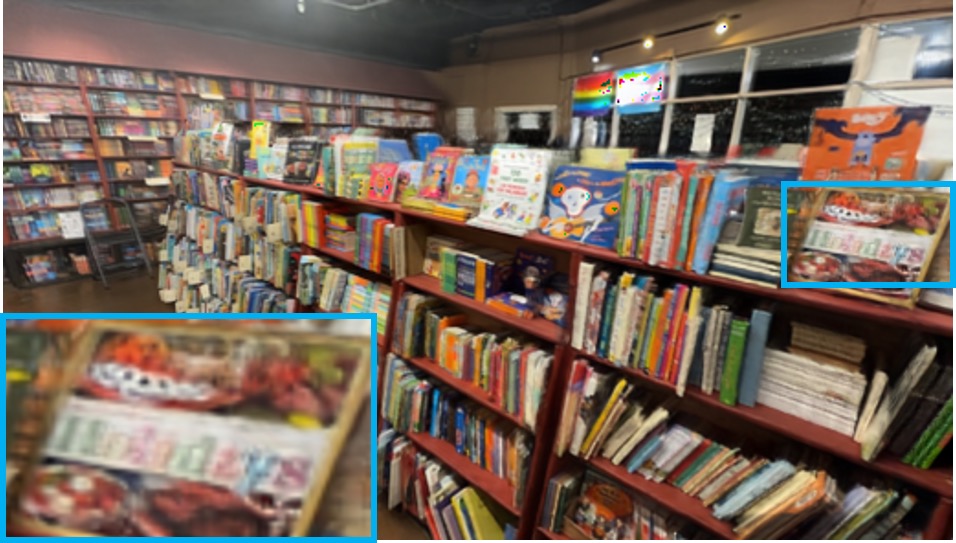} &
    \includegraphics[width=0.25\textwidth]{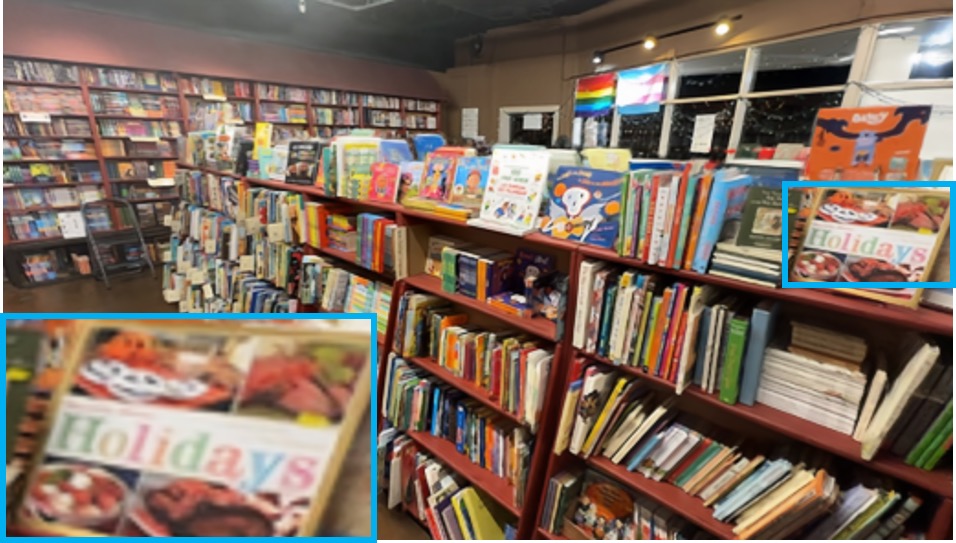} &
    \includegraphics[width=0.25\textwidth]{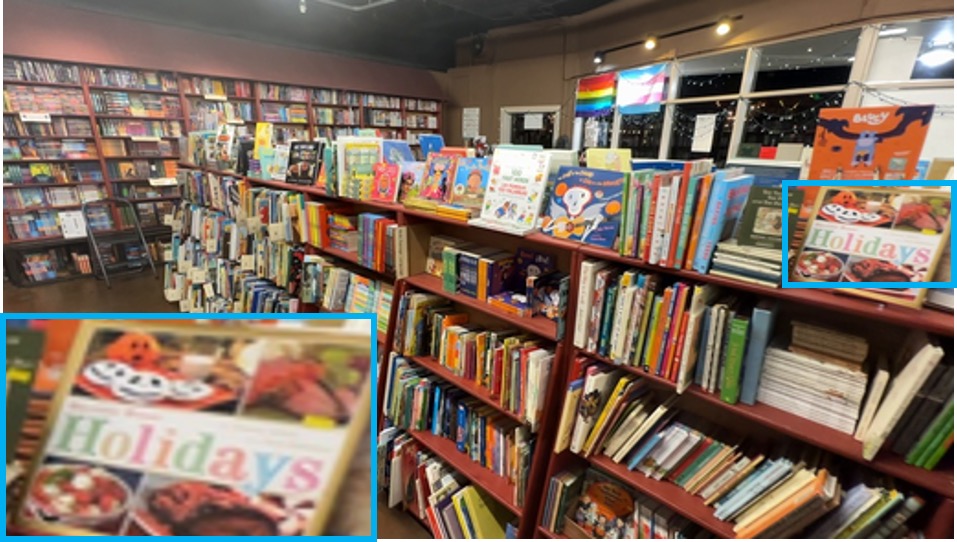}\\
    \multicolumn{1}{c}{(a) AnySplat~\cite{jiang2025anysplat}} &
    \multicolumn{1}{c}{(b) WorldMirror~\cite{liu2025worldmirror}} &
    \multicolumn{1}{c}{(c) \textbf{Ours}} &
    \multicolumn{1}{c}{(d) GT}\\
    \end{tabular}
    }
    \caption{\textbf{Qualitative Comparisons on DL3DV~\cite{ling2024dl3dv}}. We visualize novel view renderings from AnySplat~\cite{jiang2025anysplat}, WorldMirror~\cite{liu2025worldmirror}, and our method. Our method successfully recovers thin structures, such as guardrails, and fine-grained details, such as the text ``Holidays".}
    \label{fig:qualitative_dl3dv}
\end{figure*}

\section{Experiments\label{sec:experiments}}
In this section, we first briefly introduce the datasets used for training, implementation details, and evaluation metrics in~\cref{sec:exp_setup}. We demonstrate the performance of our method by comparing it with various novel view synthesis (NVS) baselines in~\cref{sec:exp_nvs}, and subsequently show in~\cref{sec:exp_pose} that our aligned features are effective for bundle adjustment (BA) and can further improve pose estimation. Finally, we present ablation studies on all the design choices described in~\cref{sec:exp_ablation}.

\begin{table*}[t]
\centering
\resizebox{\textwidth}{!}{
\begin{tabular}{llcccccccccccc}
\toprule
\multirow{2}{*}{Dataset} & \multirow{2}{*}{Method} 
& \multicolumn{3}{c}{Short (6 Frames)} 
& \multicolumn{3}{c}{Medium (12 Frames)} 
& \multicolumn{3}{c}{Long (24 Frames)} 
& \multicolumn{3}{c}{Longer (36 Frames)} \\
\cmidrule(lr){3-5} \cmidrule(lr){6-8} \cmidrule(lr){9-11} \cmidrule(lr){12-14}
 & & PSNR $\uparrow$ & SSIM $\uparrow$ & LPIPS $\downarrow$ 
 & PSNR $\uparrow$ & SSIM $\uparrow$ & LPIPS $\downarrow$ 
 & PSNR $\uparrow$ & SSIM $\uparrow$ & LPIPS $\downarrow$ 
 & PSNR $\uparrow$ & SSIM $\uparrow$ & LPIPS $\downarrow$ \\
\midrule
\multirow{5}{*}{DL3DV~\cite{ling2024dl3dv}}
 & 3DGS~\cite{kerbl20233d} & \textcolor{gray}{25.63} & \textcolor{gray}{0.8376} & \textcolor{gray}{0.1985} & \textcolor{gray}{27.92} & \textcolor{gray}{0.8794} & \textcolor{gray}{0.1678} & \textcolor{gray}{27.97} & \textcolor{gray}{0.8776} & \textcolor{gray}{0.1745} & \textcolor{gray}{28.47} & \textcolor{gray}{0.8875} & \textcolor{gray}{0.1693} \\ \cmidrule(lr){2-14}
 & AnySplat~\cite{jiang2025anysplat} & 18.84 & 0.5665 & 0.2949 & 18.16 & 0.5297 & 0.3323 & 17.41 & 0.4871 & 0.3769 & 17.25 & 0.4727 & 0.3969 \\
 & WorldMirror~\cite{liu2025worldmirror} & \underline{21.76} & \underline{0.7389} & \underline{0.2162} & \underline{20.79} & \underline{0.6884} & \underline{0.2620} & \underline{19.56} & \underline{0.6067} & \underline{0.3243} & \underline{19.24} & \underline{0.6007} & \underline{0.3412} \\
 & \textbf{Ours} & \textbf{24.94} & \textbf{0.8442} & \textbf{0.1566} & \textbf{22.96} & \textbf{0.7849} & \textbf{0.2090} & \textbf{21.58} & \textbf{0.7302} & \textbf{0.2509} & \textbf{19.97} & \textbf{0.6632} & \textbf{0.3119} \\
\midrule
\multirow{5}{*}{RealEstate10K~\cite{zhou2018stereo}}
 & 3DGS~\cite{kerbl20233d} & \textcolor{gray}{28.46} & \textcolor{gray}{0.9034} & \textcolor{gray}{0.1477} & \textcolor{gray}{29.57} & \textcolor{gray}{0.9221} & \textcolor{gray}{0.1275} & \textcolor{gray}{30.72} & \textcolor{gray}{0.9361} & \textcolor{gray}{0.1025} & \textcolor{gray}{28.47} & \textcolor{gray}{0.8875} & \textcolor{gray}{0.1693} \\ \cmidrule(lr){2-14}
 & AnySplat~\cite{jiang2025anysplat} & 23.88 & 0.7949 & 0.1836 & 21.49 & 0.7439 & 0.2352 & 19.99 & 0.6932 & 0.2727 & 21.38 & 0.7344 & 0.2377 \\
 & WorldMirror~\cite{liu2025worldmirror} & \underline{25.54} & \underline{0.8691} & \underline{0.1502} & \underline{23.99} & \underline{0.8464} & \underline{0.1759} & \underline{22.21} & \underline{0.7883} & \underline{0.2103} & \underline{24.48} & \underline{0.8521} & \underline{0.1642} \\
 & \textbf{Ours} & \textbf{28.34} & \textbf{0.9021} & \textbf{0.1206} & \textbf{27.46} & \textbf{0.8974} & \textbf{0.1317} & \textbf{26.27} & \textbf{0.8815} & \textbf{0.1469} & \textbf{25.37} & \textbf{0.8537} & \textbf{0.1602}\\
\bottomrule
\end{tabular}
}
\caption{\textbf{Novel View Synthesis with Varying Sequence Length}. We compare our method against several baselines on the RealEstate10K~\cite{zhou2018stereo} and DL3DV~\cite{ling2024dl3dv} datasets. As the number of input frames increases, the performance of all feed-forward methods degrades, as it becomes more challenging to predict consistent 3D Gaussians from a greater number of views. In contrast, 3DGS~\cite{kerbl20233d} with GT camera parameters, which we include as an upper bound, improves with more views as it can better optimize for consistency.
}
\label{tab:length_realestate_dl3dv}
\end{table*}

\begin{figure*}[t]
    \centering
    \setlength{\tabcolsep}{1pt}
    \resizebox{\textwidth}{!}{
    \begin{tabular}{cccc}
    \includegraphics[width=0.25\textwidth]{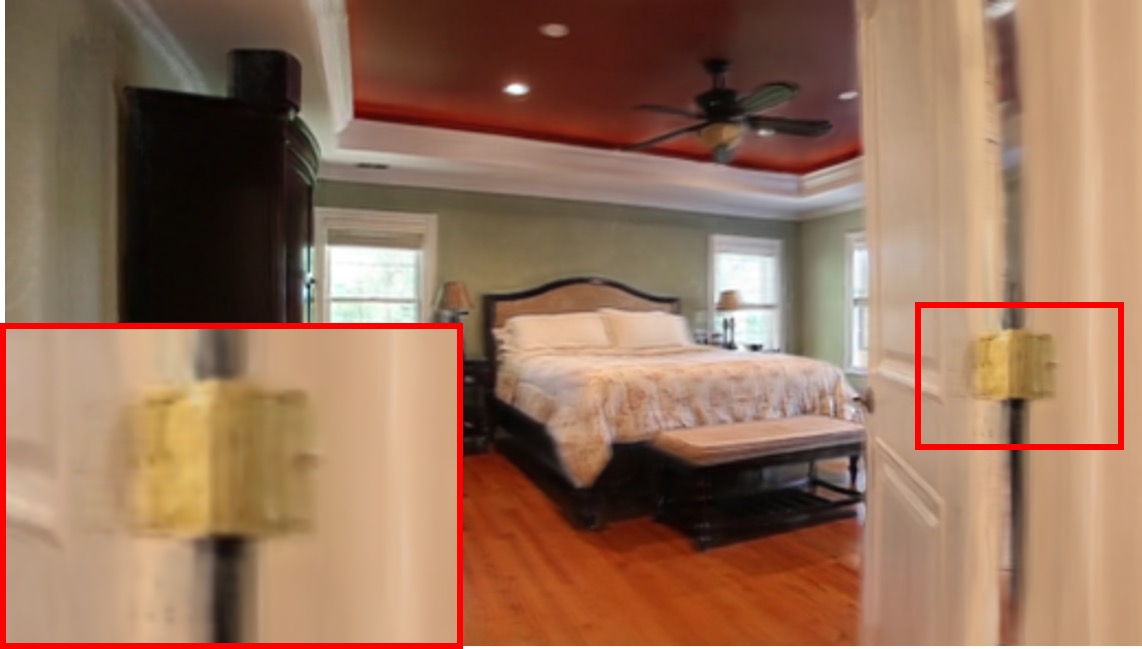} &
    \includegraphics[width=0.25\textwidth]{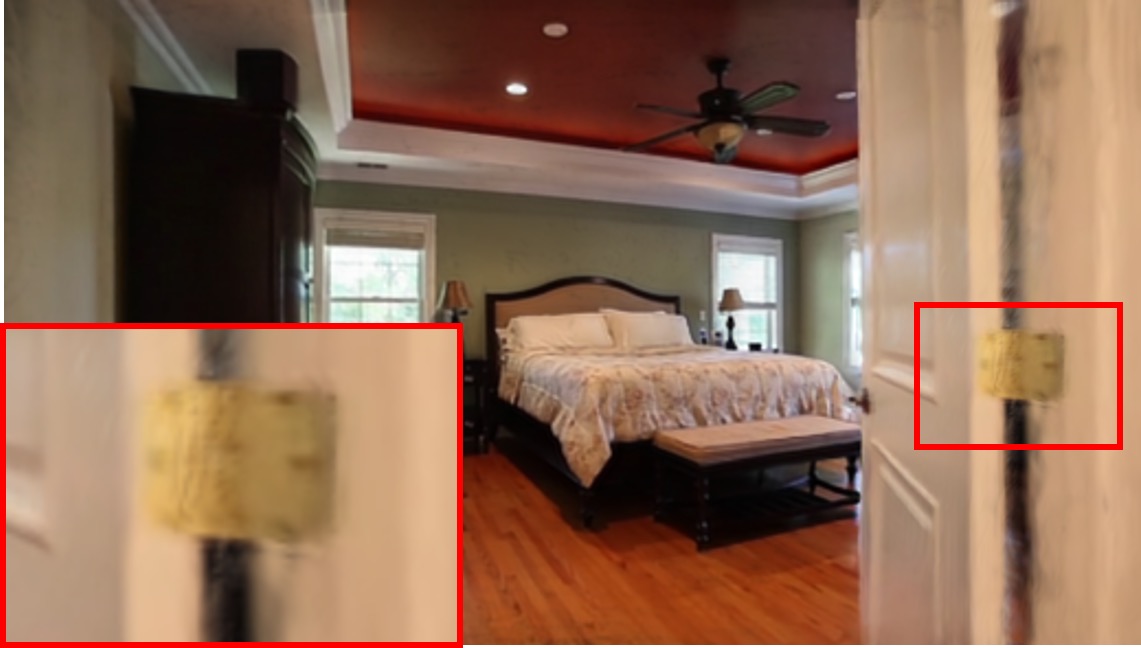} &
    \includegraphics[width=0.25\textwidth]{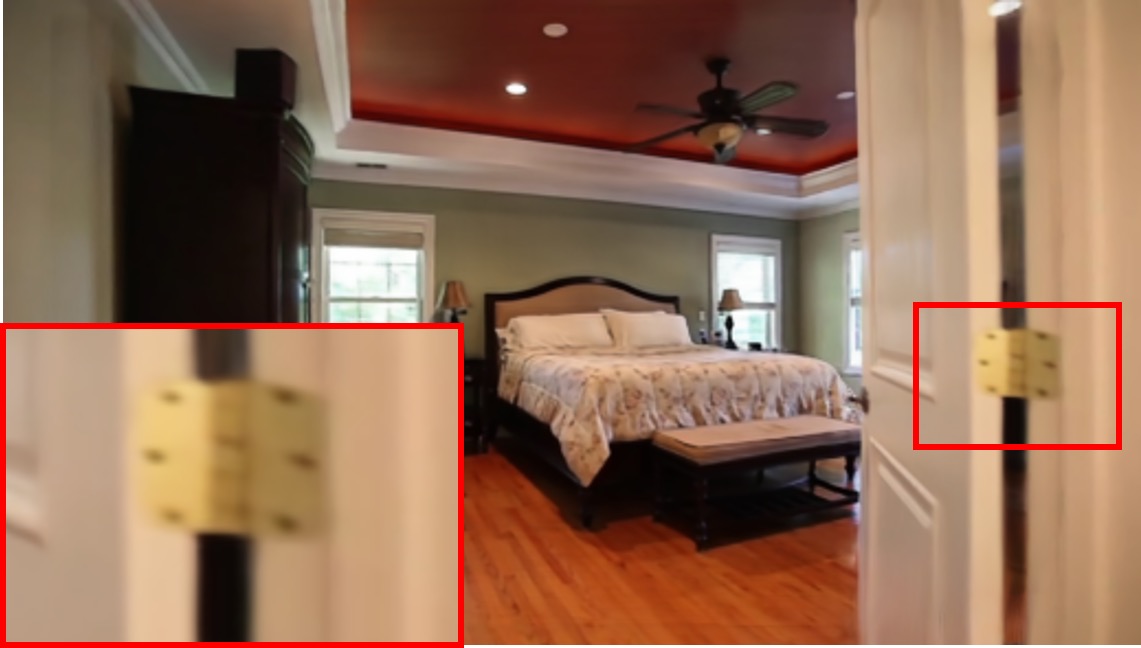} &
    \includegraphics[width=0.25\textwidth]{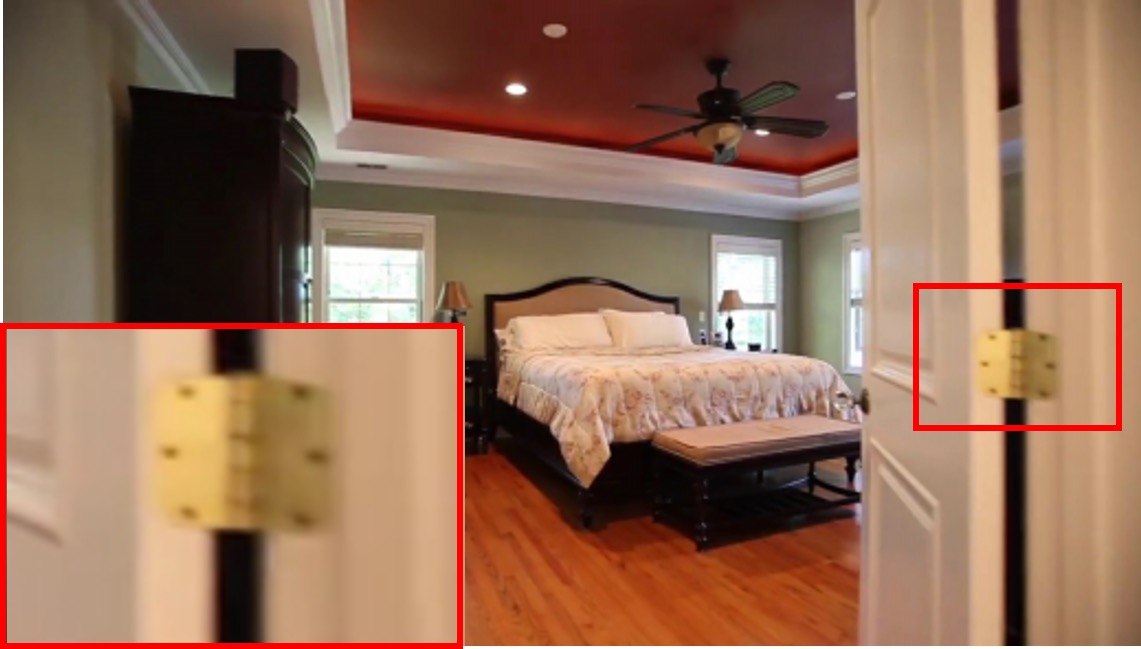}\\
    \includegraphics[width=0.25\textwidth]{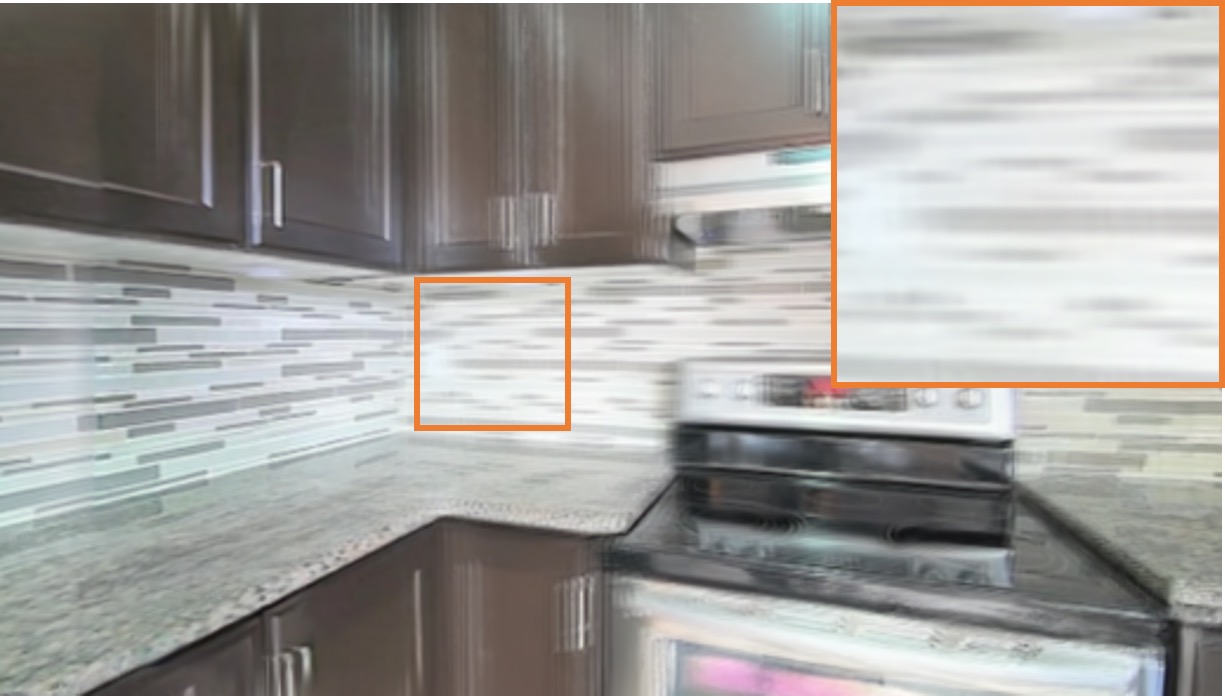} &
    \includegraphics[width=0.25\textwidth]{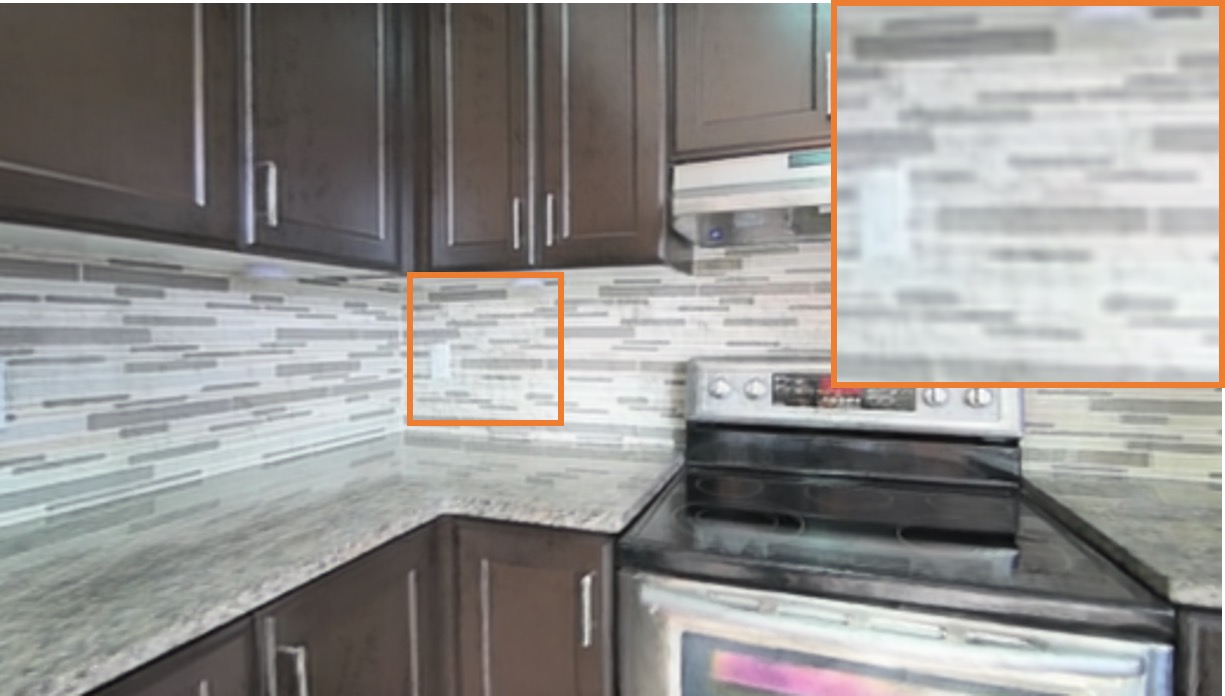} &
    \includegraphics[width=0.25\textwidth]{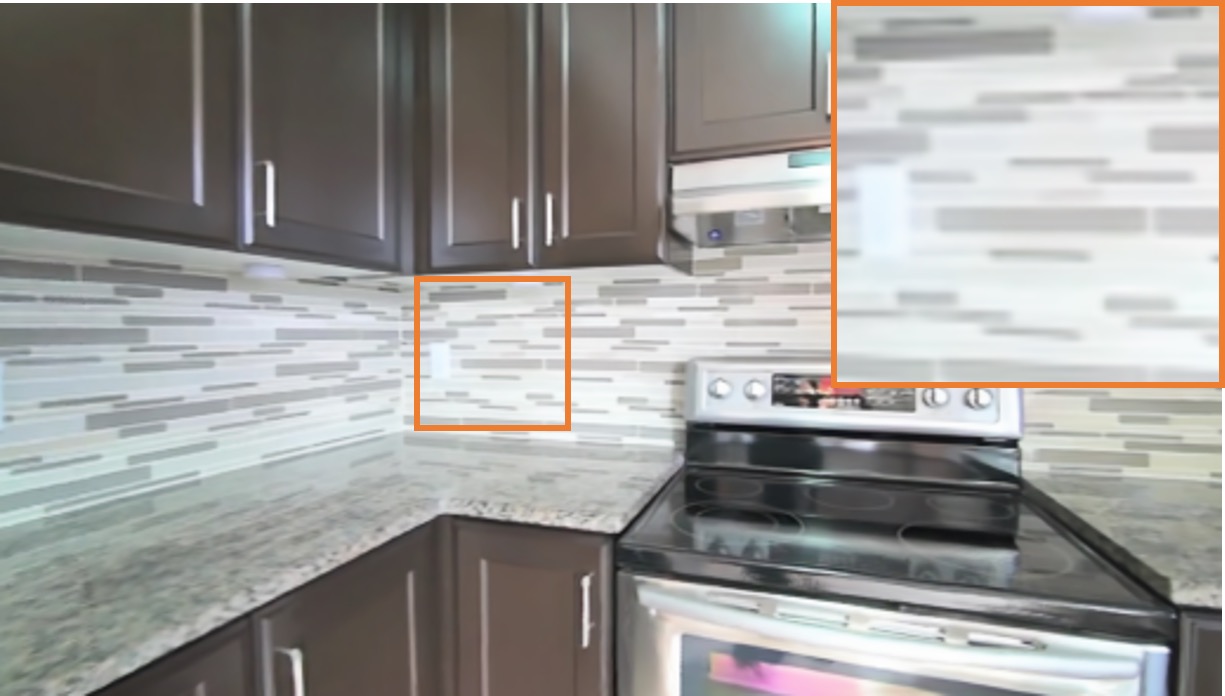} &
    \includegraphics[width=0.25\textwidth]{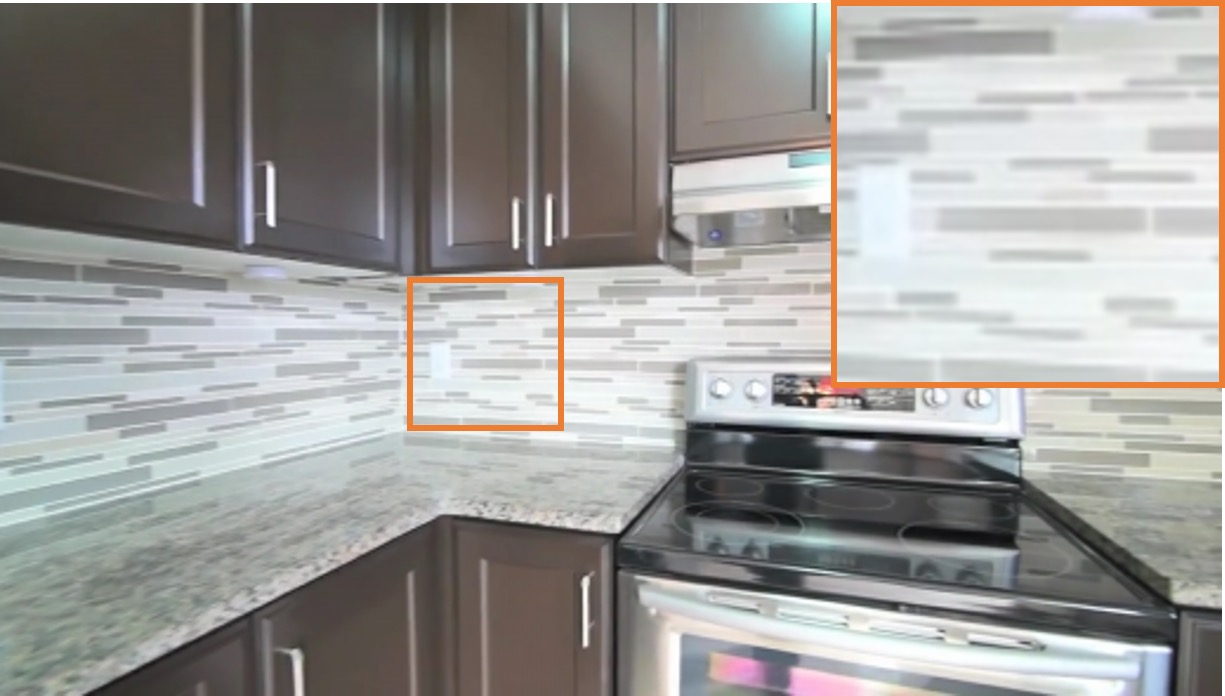}\\
    \multicolumn{1}{c}{(a) AnySplat~\cite{jiang2025anysplat}} &
    \multicolumn{1}{c}{(b) WorldMirror~\cite{liu2025worldmirror}} & \multicolumn{1}{c}{(c) \textbf{Ours}} &
    \multicolumn{1}{c}{(d) GT}\\
    \end{tabular}
    }
    \caption{\textbf{Qualitative Comparisons on RealEstate10K~\cite{zhou2018stereo}}. 
    We visualize novel view renderings from AnySplat~\cite{jiang2025anysplat}, WorldMirror~\cite{liu2025worldmirror}, and our model. Our method more faithfully reconstructs details such as the door hinge and the tiled wall.}
    \label{fig:qualitative_re10k}
\end{figure*}

\subsection{Experimental Setup}
\label{sec:exp_setup}
\myparagraph{Datasets.} We train and test our model on two large-scale public datasets: DL3DV~\cite{ling2024dl3dv} and RealEstate10K~\cite{zhou2018stereo}. Both datasets contain diverse indoor and outdoor real-world scenes at different scales, enabling the model to be robust and generalizable across various scenes. We also include conventional datasets like MipNeRF~\cite{wang2021nerf} and Tanks$\&$Temples~\cite{knapitsch2017tanks} for NVS evaluation.

\myparagraph{Training View Sampling.}
For geometric feature alignment, we sample 11 frames with different strides. The middle frame serves as the source frame, and all others are used as targets. This stage is trained exclusively on DL3DV. For the Gaussian U-Net decoder, we combine both DL3DV and RealEstate10K. We sample 6 source frames at various strides, and additionally, 5 frames between the source frames are used as the target views. More details are in the supplementary.

\myparagraph{Implementation Details.} We train the geometric features using a DPT~\cite{ranftl2021vision} decoder with the AdamW optimizer for 150K iterations. We use a cosine learning rate scheduler with a peak learning rate of $1\times10^{-4}$ and a warm-up phase of 1K steps. 
We randomly sample 4,096 query points on the source frame for training. The geometric feature alignment step takes roughly 2 days on 128 NVIDIA H100 GPUs. Next, to train the Gaussian head, we use the same optimizer and number of iterations but with a higher learning rate of $2\times10^{-4}$. The input and target rendering resolution is fixed to $294 \times 518$. This stage uses 128 NVIDIA H100 GPUs and takes about 1.5 days. The visibility threshold $\alpha$ is set to 0.05, and the softmax temperature $\tau$ is 100. The entire project is implemented in JAX~\cite{jax2018github}. More details are in the supplementary.

\begin{table*}[t]
\centering
\resizebox{\textwidth}{!}{
\begin{tabular}{llcccccccccccc}
\toprule
\multirow{2}{*}{Dataset} & \multirow{2}{*}{Method} 
& \multicolumn{3}{c}{Small (Sparse)} 
& \multicolumn{3}{c}{Medium (Regular)} 
& \multicolumn{3}{c}{Large (Dense)} 
& \multicolumn{3}{c}{Average} \\
\cmidrule(lr){3-5} \cmidrule(lr){6-8} \cmidrule(lr){9-11} \cmidrule(lr){12-14}
 &  & PSNR$\uparrow$ & SSIM$\uparrow$ & LPIPS$\downarrow$
 & PSNR$\uparrow$ & SSIM$\uparrow$ & LPIPS$\downarrow$
 & PSNR$\uparrow$ & SSIM$\uparrow$ & LPIPS$\downarrow$
 & PSNR$\uparrow$ & SSIM$\uparrow$ & LPIPS$\downarrow$ \\
\midrule
\multirow{5}{*}{DL3DV~\cite{ling2024dl3dv}}
 & 3DGS~\cite{kerbl20233d}  & \textcolor{gray}{19.58} & \textcolor{gray}{0.6628} & \textcolor{gray}{0.3517} & \textcolor{gray}{25.55} & \textcolor{gray}{0.8365} & \textcolor{gray}{0.2000} & \textcolor{gray}{25.63} & \textcolor{gray}{0.8376} & \textcolor{gray}{0.1985} & \textcolor{gray}{23.59} & \textcolor{gray}{0.7790} & \textcolor{gray}{0.2501} \\\cmidrule(lr){2-14}
 & AnySplat~\cite{jiang2025anysplat}  & 11.97 & 0.3251 & 0.5173 & 14.01 & 0.3845 & 0.4447 & 18.84 & 0.5665 & 0.2949 & 14.94 & 0.4254 & 0.4189 \\
 & Flare~\cite{zhang2025flare}  & 13.86 & 0.3223 & 0.5145 & 15.26 & 0.3715 & 0.4281 & 18.93 & 0.5364 & 0.2429 & 16.02 & 0.4101 & 0.3952 \\
 & WorldMirror~\cite{liu2025worldmirror} & \underline{14.98} & \underline{0.4641} & \underline{0.4363} & \underline{17.03} & \underline{0.5500} & \underline{0.3571} & \underline{21.76} & \underline{0.7389} & \underline{0.2162} & \underline{17.92} & \underline{0.5843} & \underline{0.3365} \\
 & \textbf{Ours} 
  & \textbf{19.46} & \textbf{0.7288} & \textbf{0.2589}
  & \textbf{22.52} & \textbf{0.8057} & \textbf{0.1903}
  & \textbf{24.94} & \textbf{0.8442} & \textbf{0.1566}
  & \textbf{22.30} & \textbf{0.7929} & \textbf{0.2019} \\
\midrule
\multirow{5}{*}{RealEstate10K~\cite{zhou2018stereo}}
 & 3DGS~\cite{kerbl20233d}  & \textcolor{gray}{21.79} & \textcolor{gray}{0.7755} & \textcolor{gray}{0.2695}  & \textcolor{gray}{24.40} & \textcolor{gray}{0.8405} & \textcolor{gray}{0.2160}  & \textcolor{gray}{28.46} & \textcolor{gray}{0.9034} & \textcolor{gray}{0.1477} & \textcolor{gray}{24.88} & \textcolor{gray}{0.8398} & \textcolor{gray}{0.2111} \\\cmidrule(lr){2-14}
 & AnySplat~\cite{jiang2025anysplat}  & 18.02 & 0.6427 & 0.2993 & 20.09 & 0.6993 & 0.2554 & 23.88 & 0.7949 & 0.1836 & 20.66 & 0.7123 & 0.2461 \\
 & Flare~\cite{zhang2025flare}  & 20.12 & 0.6423 & \underline{0.2092} & 21.66 & 0.6950 & \underline{0.1624} & 24.25 & 0.7767 & \textbf{0.1089} & 22.01 & 0.7047 & \underline{0.1602} \\
 & WorldMirror~\cite{liu2025worldmirror} & \textbf{21.31} & \underline{0.8005} & 0.2163 & \underline{22.65} & \underline{0.8283} & 0.1924 & \underline{25.54} & \underline{0.8691} & 0.1502 & \underline{23.17} & \underline{0.8326} & 0.1863 \\
 & \textbf{Ours} 
  & \underline{21.13} & \textbf{0.8575} & \textbf{0.1609}
  & \textbf{23.95} & \textbf{0.8835} & \textbf{0.1390}
  & \textbf{28.34} & \textbf{0.9021} & \underline{0.1206}
  & \textbf{24.47} & \textbf{0.8810} & \textbf{0.1402} \\
\bottomrule
\end{tabular}}
\caption{\textbf{Novel View Synthesis with Varying Overlap}. We vary the degree of overlap between the input images by changing the sampling stride. With more overlap, all methods demonstrate improved rendering performance, with our method achieving the best performance overall compared to other feed-forward baselines. Notably, our method can achieve performance on par with 3DGS~\cite{kerbl20233d}, despite 3DGS using ground-truth poses and SfM point initialization.
}
\label{tab:stride_realestate_dl3dv}
\end{table*}

\begin{table}[t]
\centering
\resizebox{\linewidth}{!}{
\begin{tabular}{llccc}
\toprule
\multicolumn{2}{c}{Method} & PSNR $\uparrow$ & SSIM $\uparrow$ & LPIPS $\downarrow$ \\
\midrule
\multirow{11}{*}{Pose-Required} 
& pixelNeRF~\cite{yu2021pixelnerf}   & 20.43 & 0.589 & 0.550 \\
& GPNR~\cite{suhail2022generalizable}           & 24.11 & 0.793 & 0.255 \\
& Du \textit{et al.}~\cite{du2023learning}   & 24.78 & 0.820 & 0.213 \\
& pixelSplat~\cite{charatan2024pixelsplat}      & 26.09 & 0.863 & 0.136 \\
& MVSplat~\cite{chen2024mvsplat}      & 26.39 & 0.869 & 0.128 \\
& DepthSplat~\cite{xu2025depthsplat} & 27.47 & 0.889 & 0.114 \\
& GS-LRM~\cite{zhang2024gs}      & 28.10 & 0.892 & 0.114 \\
& Long-LRM~\cite{ziwen2024long} & 28.54 & 0.895 & 0.109 \\
& LVSM (enc-dec)~\cite{jin2024lvsm} & 28.58 & 0.893 & 0.114 \\
& LVSM (dec-only)~\cite{jin2024lvsm} & \underline{29.67} & 0.906 & \underline{0.098} \\
& ReSplat~\cite{xu2025resplat} & \textbf{29.72} & \underline{0.911} & 0.100 \\
\midrule
\multirow{3}{*}{Pose-Free} 
& NoPoSplat~\cite{ye2024no} & 26.82 & 0.880 & 0.125 \\
& Flare~\cite{zhang2025flare} & 26.91 & 0.873 & 0.127 \\
& \textbf{Ours}  & 29.01 & \textbf{0.942} & \textbf{0.053} \\
\bottomrule
\end{tabular}
}
\caption{\textbf{Quantitative Comparison with the Two-View Convention}. We follow the two-view convention previously used by PixelSplat~\cite{charatan2024pixelsplat} on RealEstate10K~\cite{zhou2018stereo}. Except for Flare~\cite{zhang2025flare}, NoPoSplat~\cite{ye2024no}, and our method, all other methods require calibrated images as input. Our method remains competitive even against those that require ground-truth camera parameters. 
}
\label{tab:re10k_two_views}
\end{table}

\subsection{NVS with Feed-forward Gaussians}
\label{sec:exp_nvs}
Using our lightweight Gaussian head on top of geometrically aligned features, we outperform relevant baselines by a large margin in rendering quality.
Despite using uncalibrated inputs, our method achieves similar or superior performance when compared to other methods that assume known poses, including pixelNeRF~\cite{yu2021pixelnerf}, GPNR~\cite{suhail2022generalizable}, Du et al.~\cite{du2023learning}, PixelSplat~\cite{charatan2024pixelsplat}, MVSplat~\cite{chen2024mvsplat}, DepthSplat~\cite{xu2025depthsplat}, GS-LRM~\cite{zhang2024gs}, Long-LRM~\cite{ziwen2024long}, LVSM~\cite{jin2024lvsm}, and ReSplat~\cite{xu2025resplat}. At the same time, we outperform existing pose-free feed-forward Gaussian methods such as NoPoSplat~\cite{ye2024no}, Flare~\cite{zhang2025flare}, AnySplat~\cite{jiang2025anysplat}, and WorldMirror~\cite{liu2025worldmirror} by a significant margin. We also include per-scene optimized 3DGS~\cite{kerbl20233d} with ground-truth camera parameters as the strongest baseline for comparison.

For comprehensive evaluation, we evaluate performance on:  1) Varying number of input views; 2) Varying degrees of overlap among input views, by changing the sampling stride of the images; 3) Two-view evaluation split from PixelSplat~\cite{charatan2024pixelsplat} on RealEstate10K~\cite{zhou2018stereo}.
All NVS evaluations are conducted on hold-out scenes from both RealEstate10K~\cite{zhou2018stereo} and DL3DV~\cite{ling2024dl3dv}.

\myparagraph{Comparisons over Varying Sequence Lengths.}
We use a fixed minimal sampling stride (5 for RealEstate10K, 2 for DL3DV) with varying numbers of input images (6, 12, 24, and 36), corresponding to longer sequences. This setup tests the robustness of NVS models to different trajectory lengths while maintaining a relatively consistent coverage density. As detailed in~\cref{tab:length_realestate_dl3dv}, our method consistently outperforms all baselines across all settings, and is even comparable to the per-scene optimization-based 3DGS with GT camera parameters for shorter sequences.
We do not report the performance of Flare~\cite{zhang2025flare} in~\cref{tab:length_realestate_dl3dv} due to its limitations on the number of inputs.

\myparagraph{Comparisons over Varying Degrees of Overlap.}
We use a fixed number of input images (6 frames) with varying sampling strides, where larger strides correspond to smaller overlap (\textit{i.e.,} sparse-view NVS). We test strides of 5, 10, and 15 on RealEstate10K~\cite{zhou2018stereo}, and 2, 4, and 8 on DL3DV~\cite{ling2024dl3dv}. As shown in~\cref{tab:stride_realestate_dl3dv}, 
our method consistently outperforms all pose-free baselines, with performance approaching or exceeding 3DGS~\cite{kerbl20233d} in challenging sparse-view settings.
This demonstrates the effectiveness of our feature alignment objectives in creating a robust and generalized 3D scene representation, even when geometric input is sparse.

\myparagraph{Qualitative Comparison.} 
Our NVS results shown in~\cref{fig:qualitative_dl3dv} and~\cref{fig:qualitative_re10k}, using six input frames with the large overlap setting,
are noticeably sharper and recover higher-frequency details, like text and thin structures, than the baseline methods, further validating the effectiveness of our method.

\myparagraph{Two-View Evaluation.} 
We compare our method against a range of pose-required two-view and sparse-view reconstruction methods on 7k pairs from RealEstate10K~\cite{zhou2018stereo} in~\cref{tab:re10k_two_views}.
Our method achieves the best results for SSIM~\cite{wang2004image} and LPIPS~\cite{zhang2018unreasonable} among all compared methods (both pose-required and pose-free baselines). We hypothesize that the slightly lower PSNR for our model can be attributed to exposure differences in the two inputs and the fact that our model is trained for multi-view NVS, so its robustness is slightly reduced when inputs are limited to only two views.

\myparagraph{Evaluation on the RayZer~\cite{jiang2025rayzer} Split.} We evaluate our model using the random sample split used in RayZer, which consists of 16 input images and 8 target images rendered at 256$\times$256 resolution. RayZer uses a neural renderer, requiring a transformer decoder to be run for a novel view, whereas our model produces a 3D representation that can be directly rendered to novel views with Gaussian Splatting. We achieve competitive results compared to RayZer, outperforming on SSIM and LPIPS~\cref{tab:rayzer}. We note that Rayzer reports slightly different metrics for two different internal and external implementations and we compare with both implementations.

\begin{table*}[t]
\centering
\resizebox{0.7\linewidth}{!}{
\renewcommand{\arraystretch}{1.2} %
\begin{tabular}{cc c c c c}
\toprule
\multirow{2}{*}{Method} & Training & Inference w. & \multicolumn{3}{c}{Random Sample} \\
 & Supervision & COLMAP Cam. & PSNR$_{\uparrow}$ & SSIM$_{\uparrow}$ & LPIPS$_{\downarrow}$ \\ 
\midrule
GS-LRM~\cite{zhang2024gs} & 2D + Camera & \cmark &  23.02 &  0.705 &  0.266 \\
LVSM~\cite{jin2024lvsm} & 2D + Camera & \cmark &  23.10 &  0.703 &  0.257 \\ 
RayZer~\cite{jiang2025rayzer} &  2D &  \xmark &  {23.72} &  {0.733} &  {0.222} \\ 
RayZer$^*$~\cite{jiang2025rayzer} &  2D &  \xmark &  \textbf{25.47} &  \underline{0.795} &  \underline{0.181} \\ 
\textbf{Ours} &  2D + VGGT~\cite{wang2025vggt} &  \xmark &  \underline{23.75} &  \textbf{0.850} &  \textbf{0.129} \\
\bottomrule
\end{tabular}
}
\vspace{-2mm}
\caption{\textbf{Quantitative Comparisons in RayZer~\cite{jiang2025rayzer} Setup}. We evaluate renderings from our model using the evaluation index proposed in RayZer, consisting of 16 inputs and 8 targets at 256$\times$256 resolution. We attain competitive performance to RayZer, outperforming on SSIM and LPIPS, and our scene representation can be directed rasterized to novel views without the need for an additional network pass. Rayzer$^*$ is the external re-implementation.}
\label{tab:rayzer}
\end{table*}

\begin{table*}[t]
  \centering
  \resizebox{1.7\columnwidth}{!}{%
  \begin{tabular}{@{}lccccccc@{}}
    \toprule
    \multirow{2}{*}{{Dataset}} & \multirow{2}{*}{{Bundle Adjustment}} & \multicolumn{3}{c}{{Novel View Synthesis}} & \multicolumn{3}{c}{{Pose Estimation}} \\
    \cmidrule(lr){3-5} \cmidrule(lr){6-8}
    & & PSNR$\uparrow$ & SSIM$\uparrow$ & LPIPS$\downarrow$ & AUC@3 & AUC@5 & AUC@15 \\
    \midrule
    \multirow{2}{*}{DL3DV~\cite{ling2024dl3dv}} & \xmark & {24.94} & {0.844} & {0.156} & 0.813 & 0.876 & 0.955 \\
    & \cmark & {25.13} & {0.850} & {0.152} & 0.870 & 0.916 & 0.970 \\
    \midrule
    \multirow{2}{*}{MipNeRF360~\cite{barron2022mip}} & \xmark & 23.27 & 0.730 & 0.212 & 0.261 & 0.394 & 0.711\\
    & \cmark & 23.46 & 0.741 & 0.204 & 0.312 & 0.454 & 0.775 \\
    \midrule
    \multirow{2}{*}{Tanks$\&$Temples~\cite{knapitsch2017tanks}} & \xmark & 19.41 & 0.746 & 0.200 & 0.818 & 0.886 & 0.961 \\
    & \cmark & 19.54 & 0.755 & 0.198 & 0.955 & 0.973 & 0.991 \\
    \bottomrule
  \end{tabular}%
  } %
  \caption{\textbf{BA for Joint Pose Estimation and NVS.}
  Using the aligned features, we refine the camera poses using BA and further adjust the predicted Gaussian positions to be consistent with the BA output. This self-refinement operation yields further improvements over the initial NVS without BA, which we demonstrate on DL3DV and apply zero-shot to MipNeRF360 and Tanks\&Temples.
  }
  \label{tab:ba_nvs_pose}
\end{table*}

\subsection{BA for Pose Estimation and NVS}
\label{sec:exp_pose}
We demonstrate that our geometrically aligned features enable a highly effective and efficient BA stage, refining both camera poses and NVS quality.
Unlike VGGT~\cite{wang2025vggt}, which relied on the computationally heavy CoTracker~\cite{karaev2024cotracker} for correspondence -- limiting sequence length due to memory constraints -- our approach directly utilizes our aligned features for fast, pair-wise matching via simple dot products.

We evaluate pose refinement by running BA on VGGT's initial predictions using correspondences from both Co-Tracker and our method (see~\cref{tab:quantitative_ba}).
Our feature-based matching yields superior pose accuracy across varying input sizes.
Crucially, our method scales robustly to larger input sets (\textit{e.g.,} $>40$ images) where Co-Tracker fails due to memory exhaustion. %

We further validate that this improvement in pose jointly improves NVS.
We take these refined camera poses and sparse Gaussian locations, move all predicted Gaussians with the affine depth correction following~\cref{sec:dense_ba}, and obtain further improvements in NVS, as shown in~\cref{tab:ba_nvs_pose} on DL3DV~\cite{ling2024dl3dv}, MipNeRF360~\cite{barron2022mip}, and Tanks$\&$Temples~\cite{knapitsch2017tanks}.

\begin{table}[t]
\centering
\resizebox{\columnwidth}{!}{
\begin{tabular}{lcccc}
\toprule
Method & AUC@3 & AUC@5 & AUC@15 & AUC@30 \\
\midrule
\multicolumn{5}{c}{\textbf{10 frames}} \\
\midrule
VGGT w/o BA~\cite{wang2025vggt}       & 0.7900 & 0.8600 & 0.9470 & 0.9680 \\
VGGT w/ BA~\cite{karaev2024cotracker}   & 0.8350 & 0.8840 & 0.9277 & 0.9734 \\
Ours    & \textbf{0.8670} & \textbf{0.9142} & \textbf{0.9690} & \textbf{0.9842} \\
\midrule
\multicolumn{5}{c}{\textbf{40 frames}} \\
\midrule
VGGT w/o BA~\cite{wang2025vggt}       & 0.8583 & 0.9014 & 0.9602 & 0.9787 \\
VGGT w/ BA~\cite{karaev2024cotracker}   & 0.8720 & 0.9137 & 0.9588 & 0.9824 \\
Ours    & \textbf{0.8819} & \textbf{0.9235} & \textbf{0.9713} & \textbf{0.9846} \\
\midrule
\multicolumn{5}{c}{\textbf{100 frames}} \\
\midrule
VGGT w/o BA~\cite{wang2025vggt}       & 0.8447 & 0.8977 & 0.9617 & 0.9796 \\
VGGT w/ BA~\cite{karaev2024cotracker}   & \multicolumn{4}{c}{Out-of-Memory} \\
Ours    & \textbf{0.8762} & \textbf{0.9192} & \textbf{0.9699} & \textbf{0.9839} \\
\bottomrule
\end{tabular}
}
\caption{\textbf{Pose Estimation Evaluation}. VGGT achieves reasonable performance, but the estimated poses can be further improved via BA. We report results for different numbers of input images and compare our BA results against Co-Tracker~\cite{karaev2024cotracker}. Our method consistently improves the predictions, even when Co-Tracker~\cite{karaev2024cotracker} fails due to out-of-memory.}
\label{tab:quantitative_ba}
\end{table}

\begin{table*}[t]
\centering
\resizebox{1.76\columnwidth}{!}{
\begin{tabular}{cccccccc}
\toprule
Feature Alignment & Density SHs & RGB SHs & Bundle Adjustment & Depth Shift & PSNR$\uparrow$ & SSIM$\uparrow$ & LPIPS$\downarrow$ \\
\midrule
\xmark & \xmark & \xmark & \xmark & \xmark & 22.53 & 0.759 & 0.240 \\
\cmark & \xmark & \xmark & \xmark & \xmark & 23.29 & 0.792 & 0.210 \\
\cmark & \xmark & \cmark & \xmark & \xmark & 23.70 & 0.801 & 0.207 \\
\cmark & \cmark & \xmark & \xmark & \xmark & 23.55 & 0.798 & 0.205 \\
\cmark & \cmark & \cmark & \xmark & \xmark & 24.67 & 0.835 & 0.169 \\
\cmark & \cmark & \cmark & \cmark & \xmark & 24.61 & 0.833 & 0.164 \\
\cmark & \cmark & \cmark & \cmark & \cmark & \textbf{24.88} & \textbf{0.844} & \textbf{0.157} \\
\bottomrule
\end{tabular}
}
\caption{\textbf{Ablation Studies on DL3DV~\cite{ling2024dl3dv}.} We quantify the benefits of geometric feature alignment, SH prediction, and depth correction after BA. For ablations, we use a smaller batch size of 32 for training and 6 inputs with a stride of 2 for evaluation.
}
\label{tab:ablation_small_table}
\end{table*}

\begin{figure*}[t]
    \centering
    \setlength{\tabcolsep}{1pt}
    \resizebox{\textwidth}{!}{
    \begin{tabular}{cccccc}
    \includegraphics[width=0.333\textwidth]{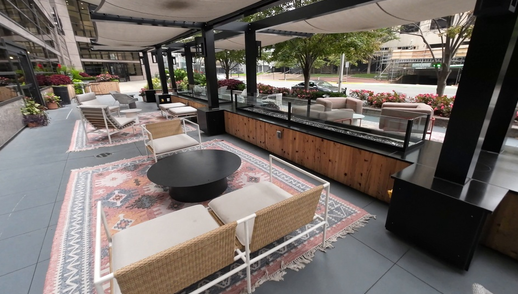} &
    \includegraphics[width=0.333\textwidth]{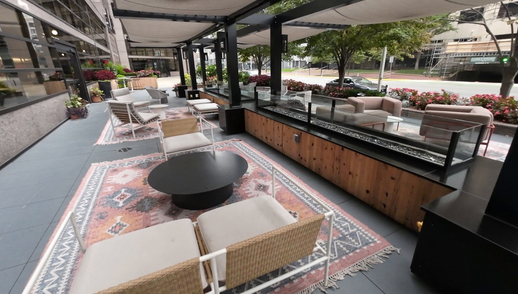} &
    \includegraphics[width=0.333\textwidth]{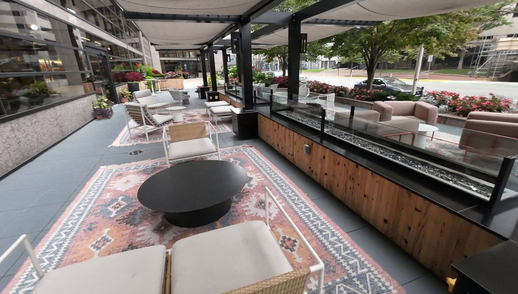} &
    \includegraphics[width=0.333\textwidth]{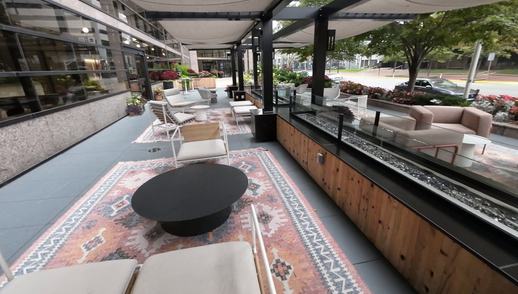} &
    \includegraphics[width=0.333\textwidth]{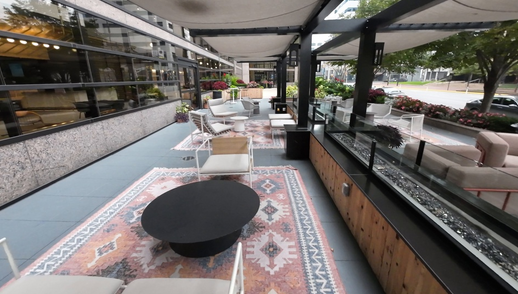} &
    \includegraphics[width=0.333\textwidth]{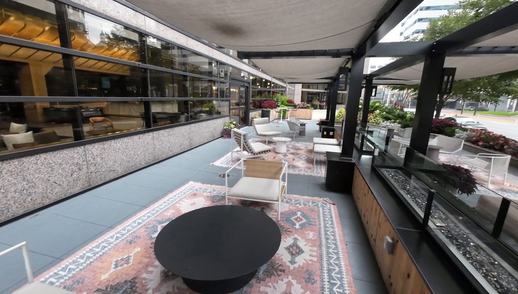} \\
    \includegraphics[width=0.333\textwidth]{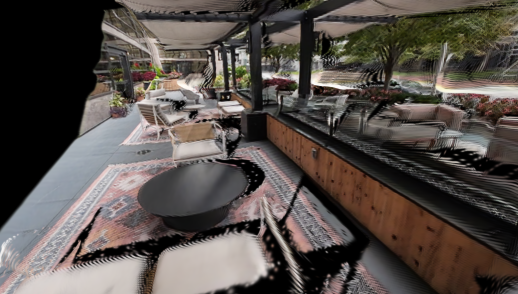} &
    \includegraphics[width=0.333\textwidth]{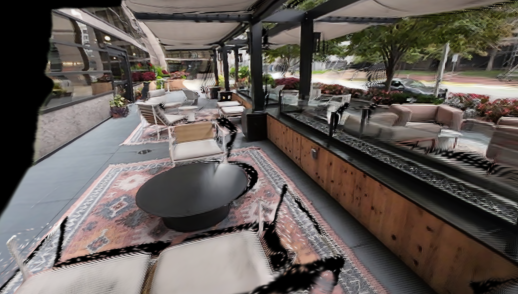} &
    \includegraphics[width=0.333\textwidth]{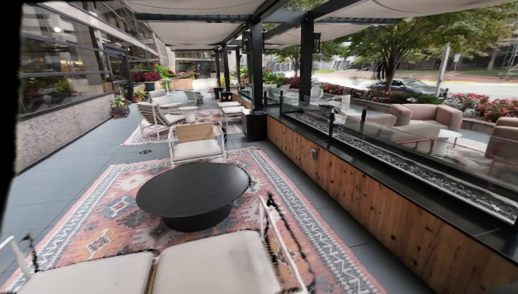} &
    \includegraphics[width=0.333\textwidth]{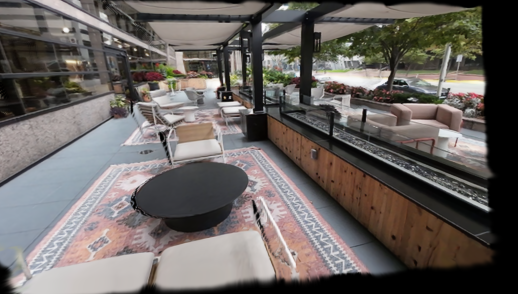} &
    \includegraphics[width=0.333\textwidth]{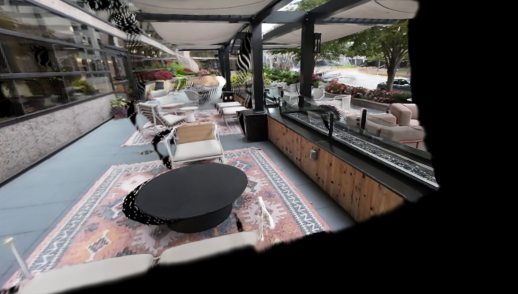} &
    \includegraphics[width=0.333\textwidth]{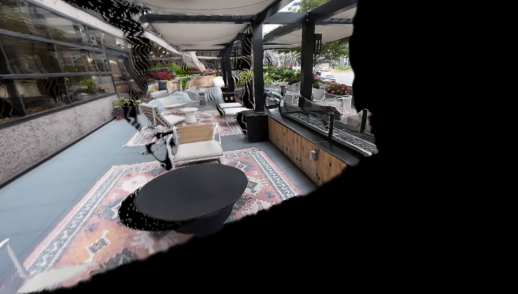} \\
    
    \includegraphics[width=0.333\textwidth]{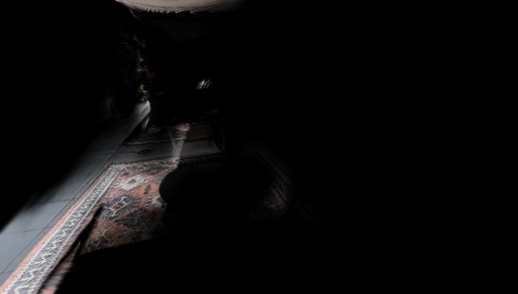} &
    \includegraphics[width=0.333\textwidth]{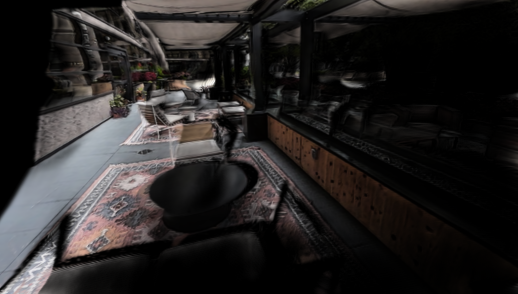} &
    \includegraphics[width=0.333\textwidth]{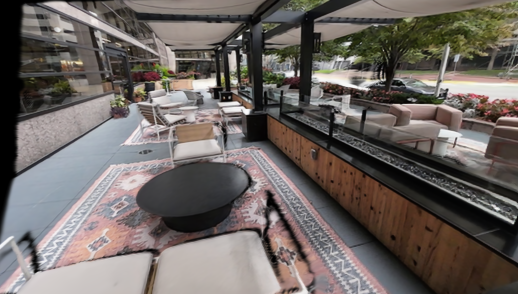} &
    \includegraphics[width=0.333\textwidth]{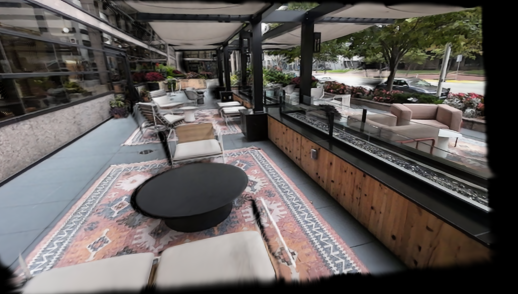} &
    \includegraphics[width=0.333\textwidth]{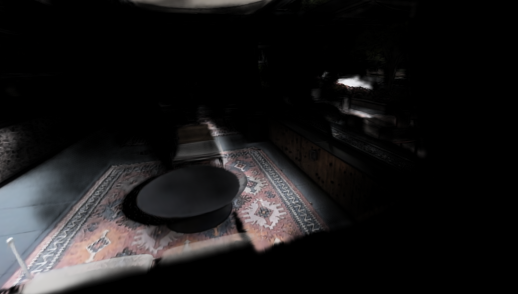} &
    \includegraphics[width=0.333\textwidth]{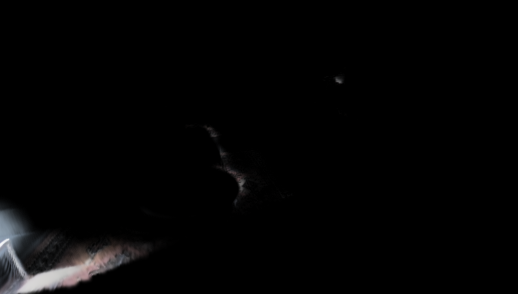} \\
    \end{tabular}
    }
    \caption{\textbf{View-dependent Density.} 
    Given six input views (top row), we render the Gaussians from each individual input to a target camera at the midpoint of the two center views. This process is shown for models trained without (middle row) and with (bottom row) view-dependent density. The view-dependent density serves as a confidence score that learns to downweight input views that are farther from the target view.
    }
    \label{fig:view_dependent}
\end{figure*}

\subsection{Ablation Study}
\label{sec:exp_ablation}
\myparagraph{Geometric Feature Alignment.} We observe that our geometrically aligned features prove crucial for obtaining a high-quality Gaussian head. To ablate this, we directly adopt features from the tracking head of VGGT, with an identical training schedule. 
The first two rows of~\cref{tab:ablation_small_table} show that our features, despite also being trained independently from the Gaussian head, offer a strong improvement in NVS metrics. 
The importance of feature alignment is also corroborated in our comparison to WorldMirror~\cite{liu2025worldmirror}, which directly passes feature tokens to the Gaussian head.

\myparagraph{Spherical Harmonics (SH) Prediction.} We ablate the importance of predicting SHs for both RGB color and density in~\cref{tab:ablation_small_table}. Unlike the original 3DGS~\cite{kerbl20233d}, which assigns only a scalar density to each Gaussian, density SHs play a crucial role for NVS in helping to combat noisy estimations in pose and depth, as shown in rows 4 and 5 of~\cref{tab:ablation_small_table}. We provide visualizations in~\cref{fig:view_dependent} for better understanding. The first row shows six input images, with the target camera placed at their midpoint. For each input view, we obtain a set of Gaussians and rasterize them into the target camera view. We show the rasterized results without (second row) and with (last row) view-dependent density in~\cref{fig:view_dependent}. Our model learns to use Gaussians from the closest input frames, while Gaussians from inputs farther from the target are assigned nearly zero density. We interpret this behavior as a learned confidence score over density, and find that it greatly improves rendering quality. %

\myparagraph{Depth Shift with BA.} Although subsequent BA using our aligned features improves poses, we find that directly substituting the new poses actually hinders NVS, as shown in the penultimate row of~\cref{tab:ablation_small_table}. 
This is because updating the camera parameters after refinement requires simultaneously moving all Gaussians to their correct locations and adjusting their scales accordingly. 
Our observations, explained in~\cref{sec:dense_ba} and~\cref{fig:ablation_depth_shift}, suggest that this Gaussian transformation can be decomposed into a camera change and an affine transformation along the z-axis. By incorporating this adjustment, we achieve better NVS results, shown in the last row of~\cref{tab:ablation_small_table}.

\vspace{-1mm}
\section{Discussions}
\label{sec:conclusion}

\myparagraph{Limitations.} Our method currently faces three primary limitations. First, reliance on the VGGT backbone's normalized scale can lead to depth inaccuracies in distant regions, such as the sky. Second, exposure discrepancies between input images and ground-truth targets may negatively impact evaluation metrics. Finally, the model is currently restricted to static scenes, which can result in feature mismatching when applied to dynamic environments.

\myparagraph{Conclusion.} We introduce \modelname, a self-improving approach for novel view synthesis from unposed and uncalibrated images.
Our key insight is that while modern visual foundation models provide a strong prior for a variety of 3D reasoning tasks, their feature spaces may not be explicitly 3D aligned.
Using VGGT as a VFM backbone, we train a feature adapter by constructing a self-supervised reprojection task, enforcing feature consistency between 2D projections that refer to a common 3D point.
This geometric alignment not only improves downstream NVS performance but also yields robust correspondences for refining camera parameters via bundle adjustment. 
We can then feed the refined cameras back into the NVS model, and obtain further improvements in rendering quality. 
Using a self-supervised loop that aligns VFM features and refines poses, our model achieves state-of-the-art performance, demonstrating a powerful new direction for robust, unposed novel view synthesis.

\section*{Acknowledgment}
We would like to thank Xichen Pan and Noah Snavely for insightful discussions during the project, and Clément Godard, Michael Broxton, and Maggie Oh for help with compute support. Additionally, we thank Stephen Lombardi, Ryan Overbeck, and Jason Lawrence for helpful suggestions and feedback.

\clearpage
\maketitlesupplementary
\setcounter{section}{0}

{
\renewcommand{\cftsecafterpnum}{\vskip15pt}
\setlength{\cftsecindent}{1em}
\setlength{\cftbeforesecskip}{0.5em}

\setcounter{tocdepth}{-1}

\begingroup
\renewcommand{\contentsname}{Content}
\endgroup

\addtocontents{toc}{\protect\setcounter{tocdepth}{1}}
}

\noindent In our supplementary material we provide additional details on training and evaluation strategies in~\cref{sec:supp_method}. We follow with experiments on alternative loss objectives, more evaluations, and visualizations in~\cref{sec:supp_experiment}, and a discussion on limitations in~\cref{sec:supp_limitations}. We additionally include a webpage with animated video results and visualizations.

\section{Supplementary Methods}\label{sec:supp_method}

\subsection{Additional Implementation Details}

\myparagraph{Feature Correspondence Training.} For pseudo-ground-truth supervision, we obtain 2D query points from a source frame, and use depth and camera parameters to project the query point onto corresponding target points in the target frames. During training, the query points are selected uniformly at random from the source frames, and we use a batch size of 4096 query points per frame. We select the middle frame in the sequence as the source frame and project all other frames as target frames. During inference, we select 2048 points per source frame, and we use every fifth frame in the sequence as source frames. We use a radius of five frames around each source frame as the target frames for feature matching. To filter out unreliable matches, we adopt three criteria to remove matches: (1) if the query point falls outside of the target field-of-view using predicted depth and relative camera pose, (2) if the query point is projected behind the target camera (with negative depth), and (3) if the confidence of either point is less than 1.2 using the VGGT confidence maps.

\myparagraph{Gaussian Head Training.} For Gaussian head training, we sample a random stride between 2 and 6 for DL3DV and between 5 and 15 for RealEstate10K, then take a sequence of 11 frames -- 6 inputs and 5 targets in between them. In addition, as the view-dependent density term results in low-confidence Gaussians becoming transparent, we prune these Gaussians for rendering efficiency. For a given render camera, we sort the Gaussians by predicted opacity, and remove 30\% of the Gaussians with lowest opacity. This reduces the number of primitives that need to be rasterized, thus improving memory efficiency during training.

\subsection{Baseline Evaluations}

\myparagraph{Two-View RealEstate10K.} We evaluate on the two-view split proposed by PixelSplat~\cite{charatan2024pixelsplat}, which consists of three target images randomly sampled between two context images at 256x256 resolution. We adopt the metrics from ReSplat~\cite{xu2025resplat} for evaluation on methods that require posed inputs, and we similarly evaluate NoPoSplat~\cite{ye2024no}, Flare~\cite{zhang2025flare}, and our model on the same images. For NoPoSplat and Flare, we perform post-hoc camera optimization to maximize the similarity of the rendered image to the target image using the ground-truth cameras as initialization, following the official codebases for each method. We do not perform any per-scene optimization with our method.

\myparagraph{Multi-view Evaluation.} We use the official codebases and pretrained checkpoints for Anysplat~\cite{jiang2025anysplat} and WorldMirror~\cite{liu2025worldmirror}. We evaluate these baselines and our model at 294x518 resolution on the same images. We first provide inputs and targets together to obtain poses for all images, then we run the Gaussian U-Net only on the input images and render the input-aligned primitives to the target images. For Flare~\cite{zhang2025flare}, we find that with the released checkpoint we obtain better performance using 256x256 resolution and three images rather than the full image set, so we evaluate using a sliding window of one target and the two adjacent input frames using PnP to produce the poses; both the reduced resolution and sliding evaluation are advantageous to the Flare baseline relative to our evaluation of our model and the remaining baseline methods.

\myparagraph{Varying Sequence Length and Varying Stride.} In our main paper Tab.~1, %
we investigate the impact of changing the number of inputs, whereas in main paper Tab.~2 ,%
we evaluate the impact of changing the overlap between views.
We note that these are two separate challenges for our method. We control overlap by changing the sampling stride of the sequence. Decreasing the overlap between views (\textit{i.e.,} increase the camera baseline) makes reasoning about 3D structure more difficult. Therefore reconstruction better for large overlap than small overlap. On the other hand, we separately hold the sampling stride fixed and change the number of inputs. As our Gaussian primitives are pixel-aligned, having more inputs becomes harder because the model must produce consistent geometry for pixels among all views to avoid duplication or occlusion artifacts. Thus, we find that reconstruction metrics are better for short sequences compared to longer ones.

\section{Supplementary Experiments}\label{sec:supp_experiment}

\subsection{Alternative Training Objectives}

\begin{figure*}[t]
    \centering
    \setlength{\tabcolsep}{1pt}
    \resizebox{\textwidth}{!}{
    \begin{tabular}{cccccc}
    \includegraphics[width=0.333\textwidth]{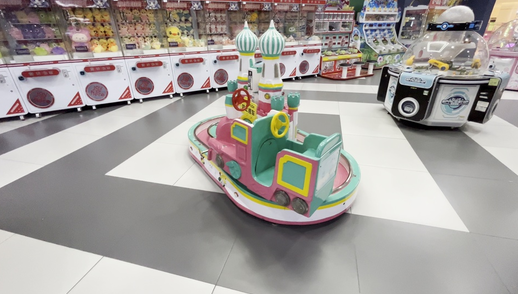} &
    \includegraphics[width=0.333\textwidth]{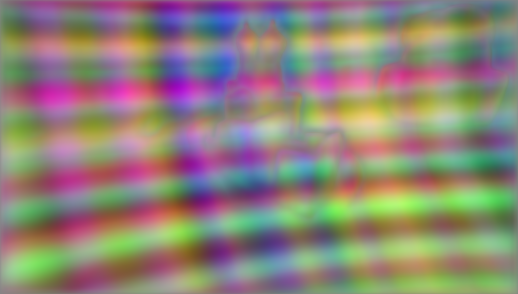} &
    \includegraphics[width=0.333\textwidth]{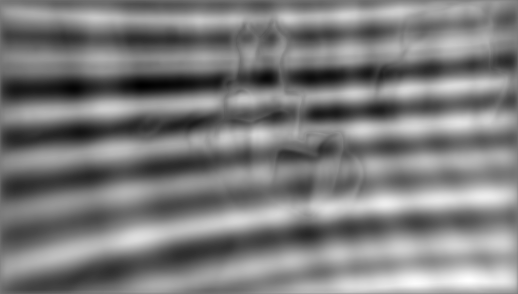} &
    \includegraphics[width=0.333\textwidth]{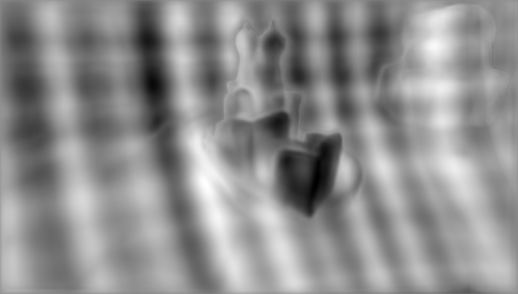} &
    \includegraphics[width=0.333\textwidth]{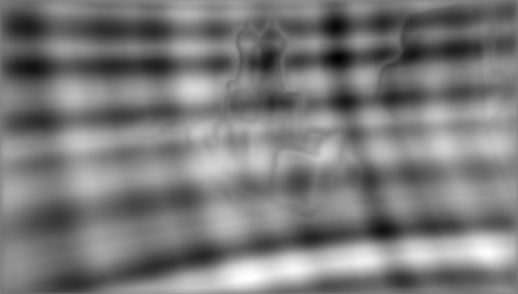} &
    \includegraphics[width=0.333\textwidth]{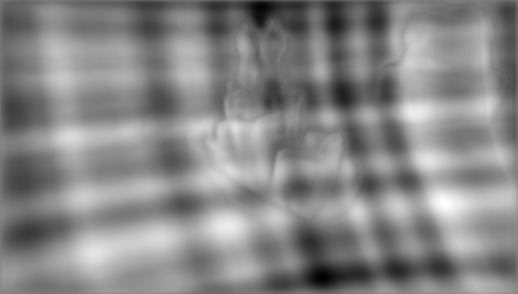} \\
    
    \includegraphics[width=0.333\textwidth]{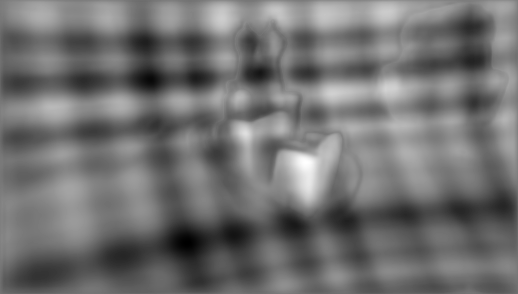} &
    \includegraphics[width=0.333\textwidth]{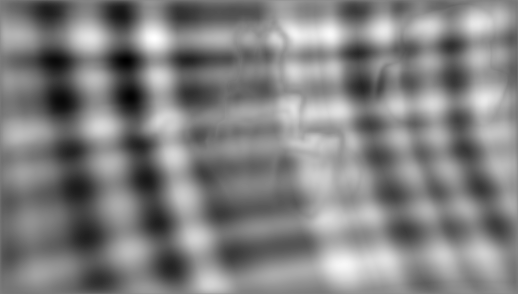} &
    \includegraphics[width=0.333\textwidth]{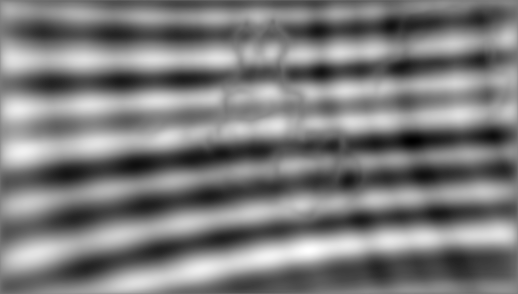} &
    \includegraphics[width=0.333\textwidth]{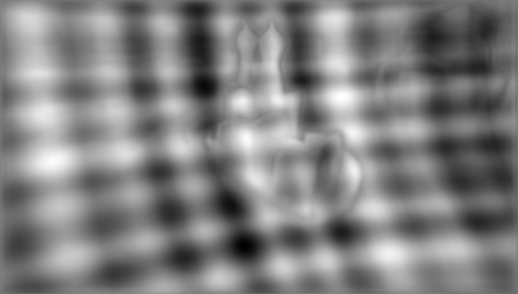} &
    \includegraphics[width=0.333\textwidth]{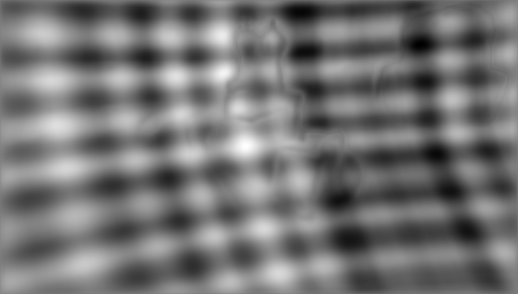} &
    \includegraphics[width=0.333\textwidth]{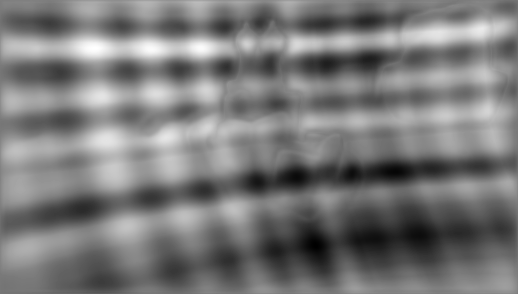} \\
    
    \includegraphics[width=0.333\textwidth]{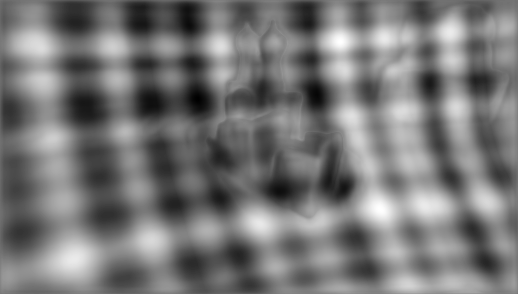} &
    \includegraphics[width=0.333\textwidth]{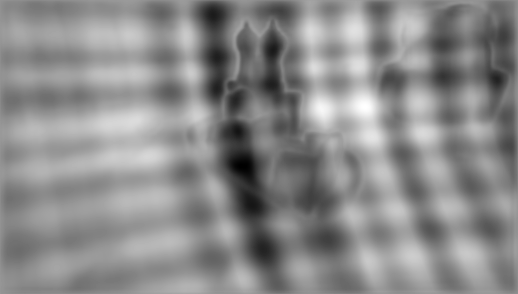} &
    \includegraphics[width=0.333\textwidth]{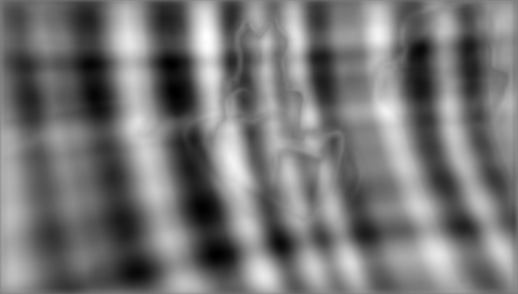} &
    \includegraphics[width=0.333\textwidth]{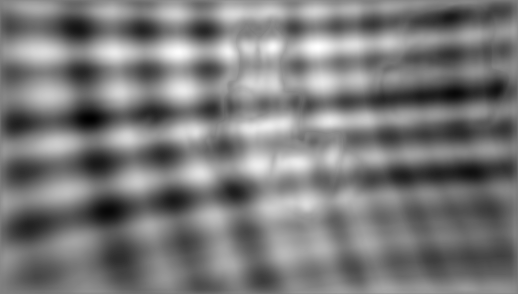} &
    \includegraphics[width=0.333\textwidth]{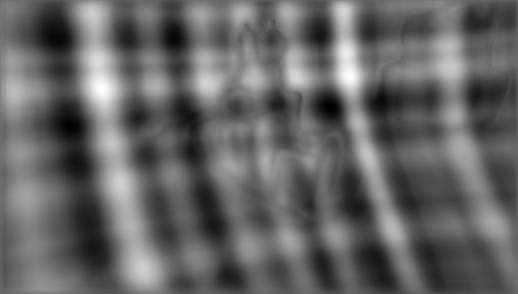} &
    \includegraphics[width=0.333\textwidth]{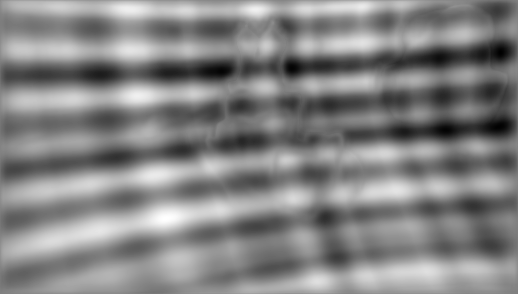} \\

    \includegraphics[width=0.333\textwidth]{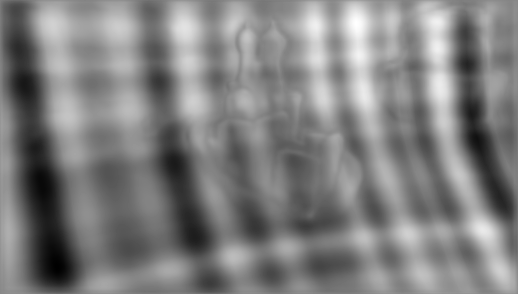} &
    \includegraphics[width=0.333\textwidth]{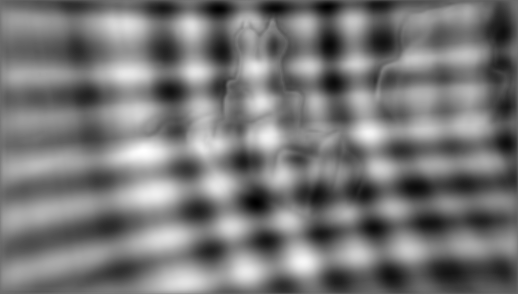} &
    \includegraphics[width=0.333\textwidth]{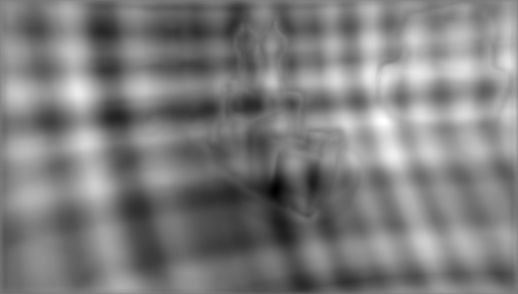} &
    \includegraphics[width=0.333\textwidth]{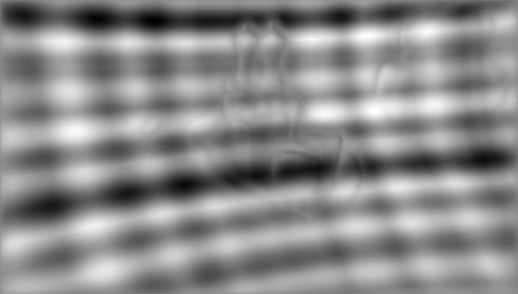} &
    \includegraphics[width=0.333\textwidth]{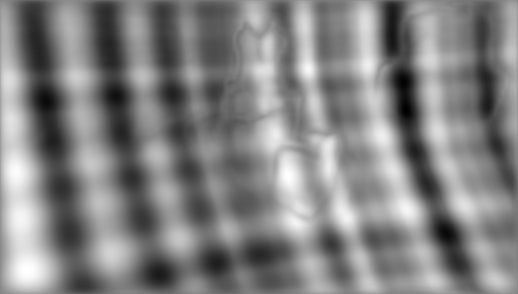} &
    \includegraphics[width=0.333\textwidth]{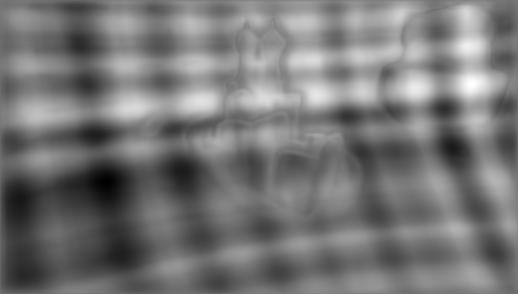} \\
    \end{tabular}
    }
    \caption{\textbf{Feature Visualization.} For the top-left image, we produce a 24-dimensional feature map to compute corresponding points. First, we average across three groups of eight channels to produce a color visualization. While we observe a structured spatial pattern in this image, visualizing the channels independently (showing 22 channels in gray) reveals that each channel captures a slightly different pattern.
    }
    \label{fig:featmap_channel}
\end{figure*}

\begin{figure}[t]
    \centering
    \includegraphics[width=\columnwidth]{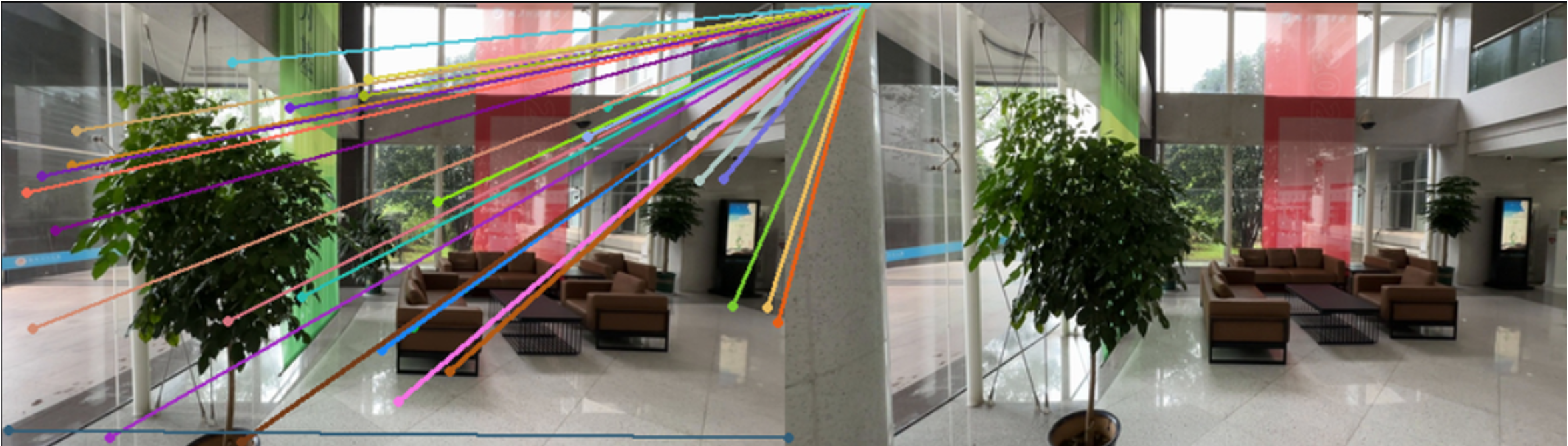}
    \caption{\textbf{Contrastive Loss Experiment.}  Using a CLIP-style contrastive training objective encourages features of correct 2D-2D matching points to be more similar than those of incorrect matches. In our experiments, we found that this objective simply resulted in the features converging to the same value, so that all queries match to the same target point. Thus we adopted the alternative strategy that encourages features of the correct match to be more similar than all other pixels in the target image. 
    }
    \label{fig:contrastive}
\end{figure}

\myparagraph{Contrastive Training Objective.} The goal of our feature alignment is to train the feature adapter such that features corresponding to spatially proximal 3D locations exhibit high similarity.  One straightforward approach is to follow a CLIP-style contrastive learning scheme~\cite{radford2021learning}, as used in~\cite{mast3r_eccv24}. Given a set of 2D-2D correspondences, the training objective encourages the correct match to have more similar features than the incorrect matches. However, in our experiments we find that this objective causes our features to collapse to the same value, as shown in~\cref{fig:contrastive}. We hypothesize that this may be due to insufficient supervision, where the contrastive loss only supervises over matched points, whereas our final objective supervises over all pixels in the target frame.

\myparagraph{Nearest Neighbor as Pseudo-Ground-Truth.} For our pseudo-ground-truth we use the predicted depth from the source frame $\mathbf{D}_s$ and the relative camera pose from source to target frame $\mathbf{R}_{\tgt \leftarrow \src}$, $\mathbf{t}_{\tgt \leftarrow \src}$ to determine 2D-2D matches for the query pixels in the target frames. An alternative strategy is to compare the 3D points produced from source and target frames and assign correspondences when 2D points coincide in 3D. For this, we compute the 3D target-frame coordinates of the query point $\mathbf{P}_t^n$ following Eq.~5 in the main text,%
 and also unproject all points in the target frame using $\mathbf{D}_t\pi^{-1}_{\mathbf{K}}\mathbf{p}_t$. We assign the correpondence in the target frame as the pixel that is closest in 3D space to $\mathbf{P}_t^n$. However, we observed that this pseudo-ground-truth is prone to noise, and produces matches that jitter as the target frame moves. A visualization of this effect is presented in the pseudo-GT videos on the last section of our project page. The pseudo-GT matching points derived from projection remain consistent even across long sequences, whereas points matched using K-Nearest Neighbors appear noisy. Even with a large K (\textit{e.g.,} 10), significant jitter is observed. This pseudo-GT negatively impacts the training of our feature adapter.

\subsection{Additional Evaluations and Visualizations}

 \begin{table}[t]
\centering
\resizebox{0.8\columnwidth}{!}{
\begin{tabular}{ccccc}
\toprule
Stride & BA & AUC@3 & AUC@5 & AUC@15 \\
\midrule
\multirow{2}{*}{2} & \xmark & 0.8130 & 0.8767 & 0.9550 \\
 & \cmark & 0.8669 & 0.9140 & 0.9690 \\
\midrule
\multirow{2}{*}{4} & \xmark & 0.8439 & 0.8973 & 0.9623 \\
 & \cmark & 0.8672 & 0.9105 & 0.9657 \\
\midrule
\multirow{2}{*}{8} & \xmark & 0.8443 & 0.9008 & 0.9648 \\
 & \cmark & 0.8501 & 0.9049 & 0.9662 \\
\bottomrule
\end{tabular}
}
\caption{\textbf{Bundle Adjustment with Different Strides}. We increasing the sampling stride of the input sequence, where larger stride corresponds to larger camera baseline. As the stride increases, the initial pose estimate from VGGT improves, but a subsequent bundle adjustment using our features offers further improvements.}
\label{tab:ba_stride}
\end{table}
\myparagraph{Bundle Adjustment with Different Sampling Strides.} %
In Tab.~5 of the main text, we evaluated our feature correspondences for improving poses via bundle adjustment on sequences of different lengths with a fixed stride of 2. Here, we additionally investigate changing the sampling stride between 2, 4, and 8, holding the sequence length fixed to 10 images. This effectively increases the baseline between adjacent images. We find that our correspondences consistently improve the pose estimation after bundle adjustment under these various settings in ~\cref{tab:ba_stride}.

\myparagraph{Visualization of Geometrically Aligned Features.} We visualize the per-channel values of our trained features in~\cref{fig:featmap_channel}. We observe that the channels capture both image structure and learned spatial encoding values. Although we observe a repeating pattern, when we plot individual channel values we find that the learned patten each channel is slightly different.

\begin{figure*}[t]
    \centering
    \setlength{\tabcolsep}{1pt}
    \resizebox{\textwidth}{!}{
    \begin{tabular}{cc}
    \includegraphics[width=0.5\textwidth]{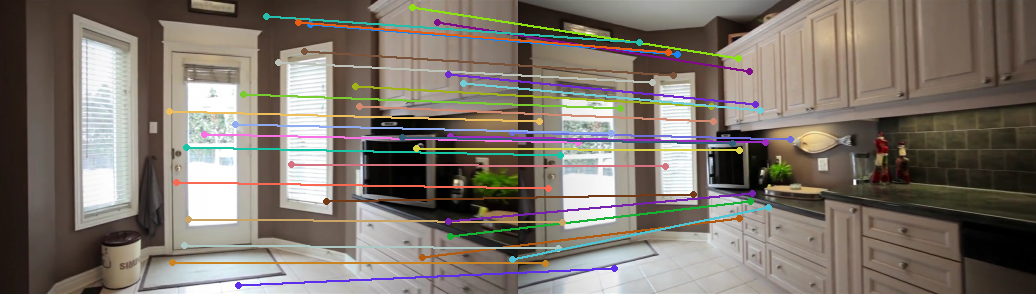} &
    \includegraphics[width=0.5\textwidth]{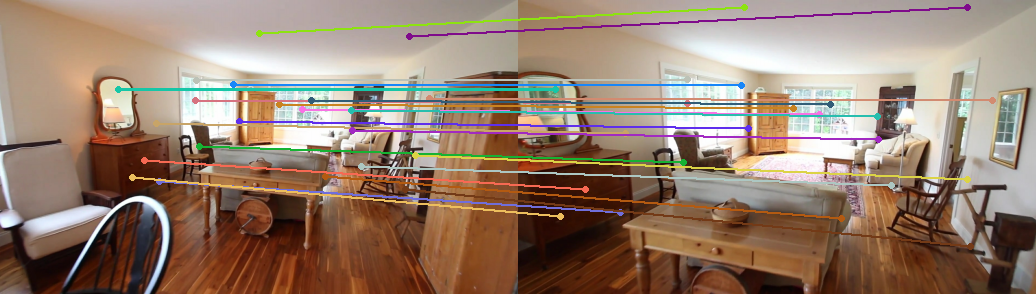} \\
    \includegraphics[width=0.5\textwidth]{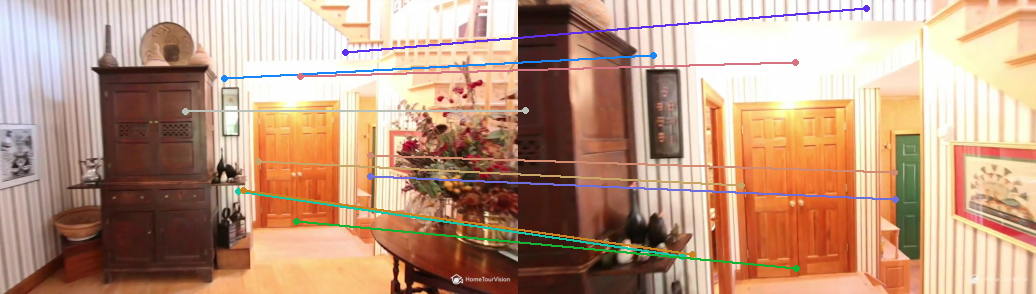} &
    \includegraphics[width=0.5\textwidth]{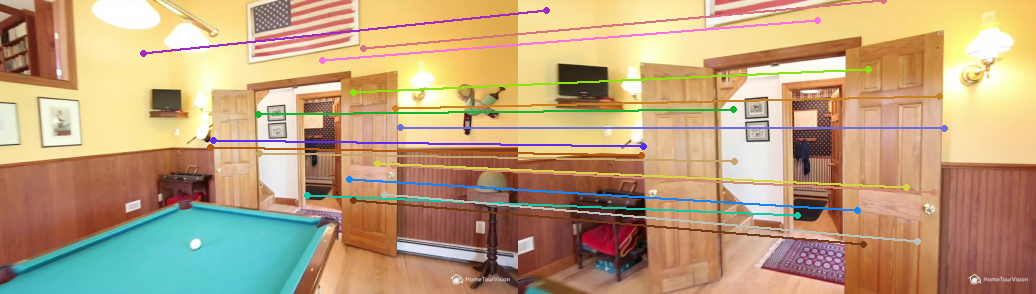} \\
    \includegraphics[width=0.5\textwidth]{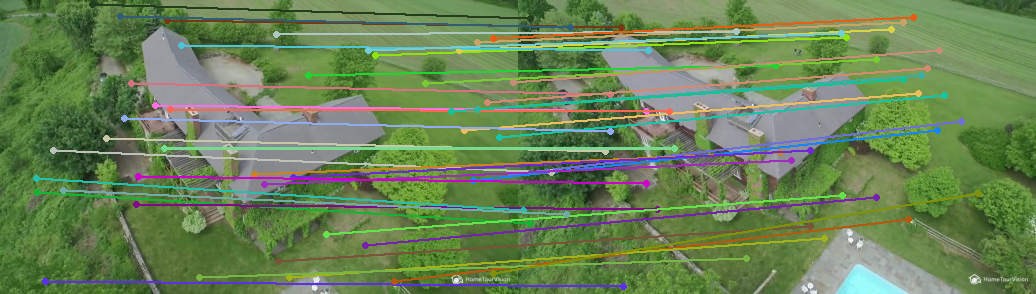} &
    \includegraphics[width=0.5\textwidth]{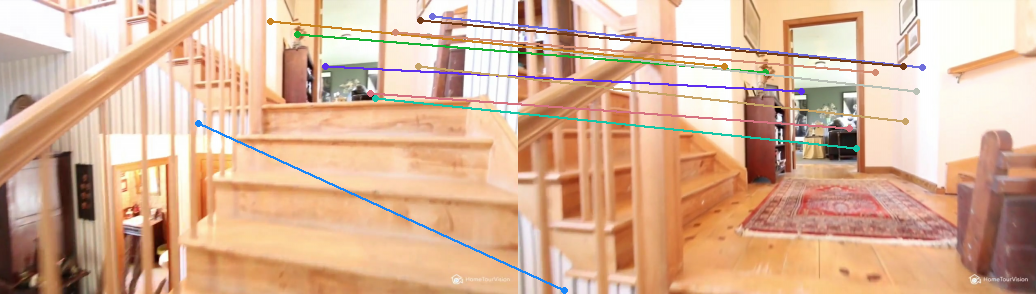} \\
    \includegraphics[width=0.5\textwidth]{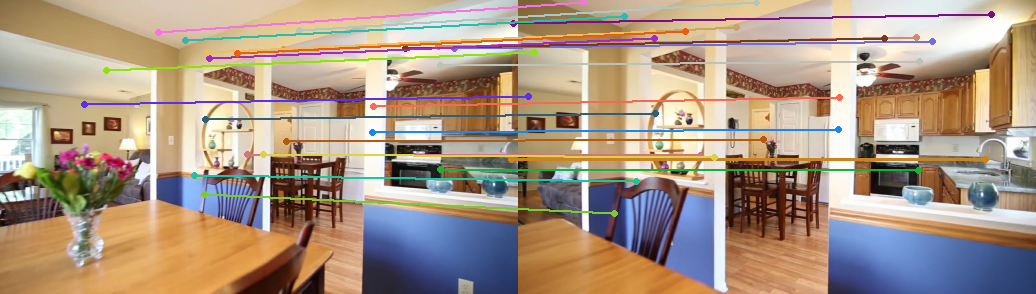} &
    \includegraphics[width=0.5\textwidth]{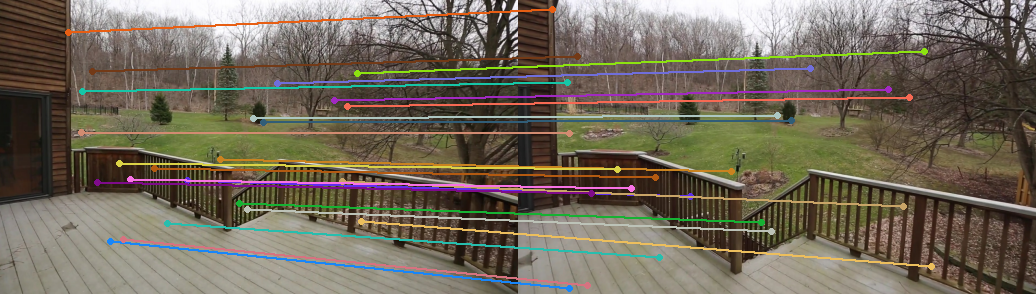} \\
    \includegraphics[width=0.5\textwidth]{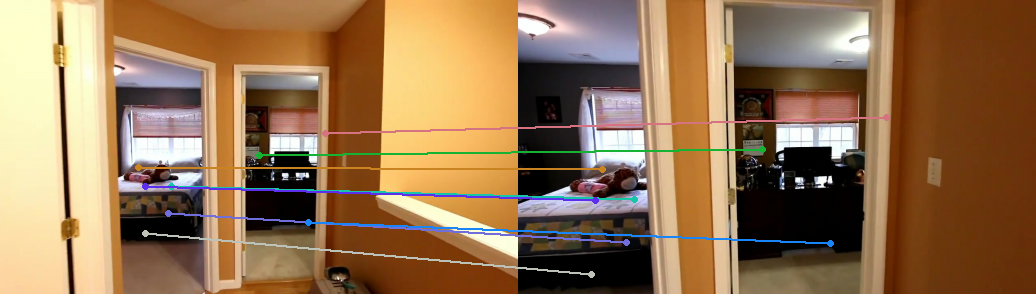} &
    \includegraphics[width=0.5\textwidth]{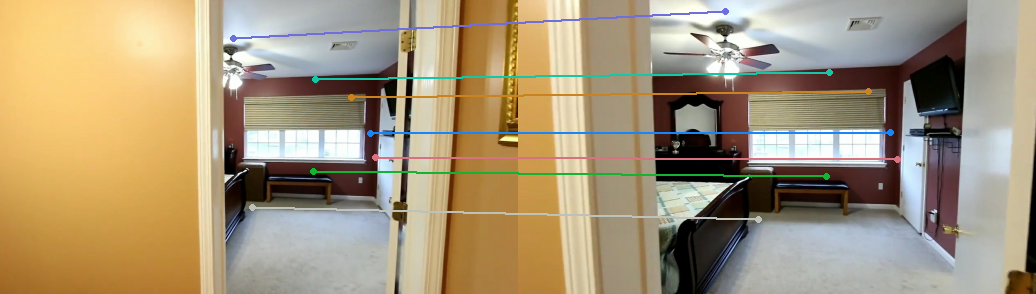}
    \end{tabular}
    }
    \caption{\textbf{Visualization of Matching on Re10K}. We similarly point query points (left image) and their matching pixels (right image) for the RealEstate10K dataset.}
    \label{fig:matching_re10k}
\end{figure*}

\begin{figure*}[t]
    \centering
    \setlength{\tabcolsep}{1pt}
    \resizebox{\textwidth}{!}{
    \begin{tabular}{cc}
    \includegraphics[width=0.5\textwidth]{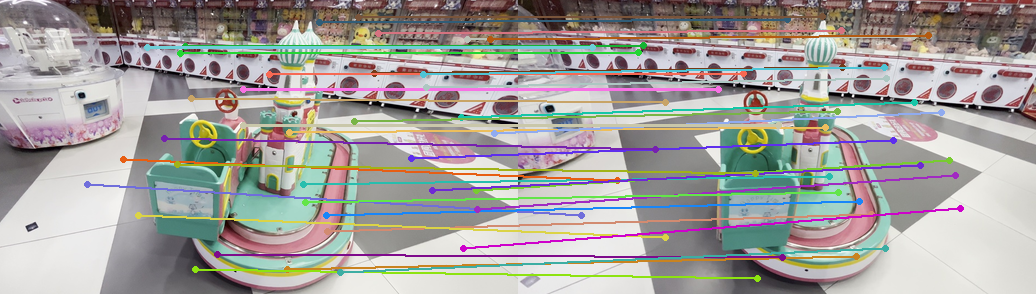} &
    \includegraphics[width=0.5\textwidth]{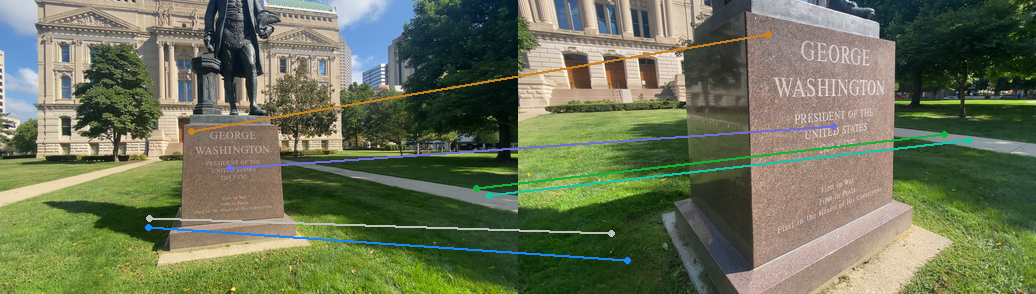} \\
    \includegraphics[width=0.5\textwidth]{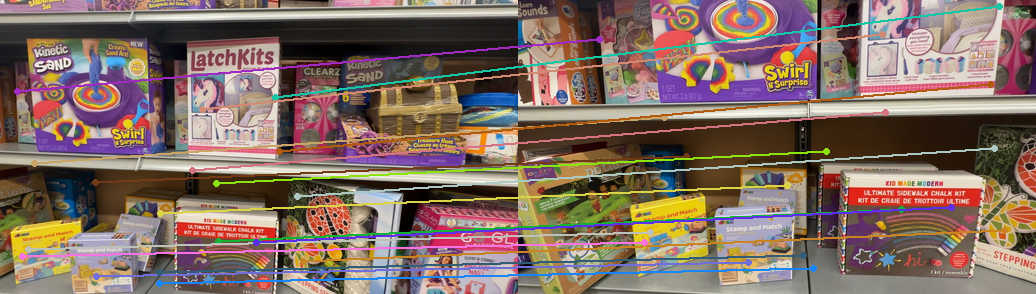} &
    \includegraphics[width=0.5\textwidth]{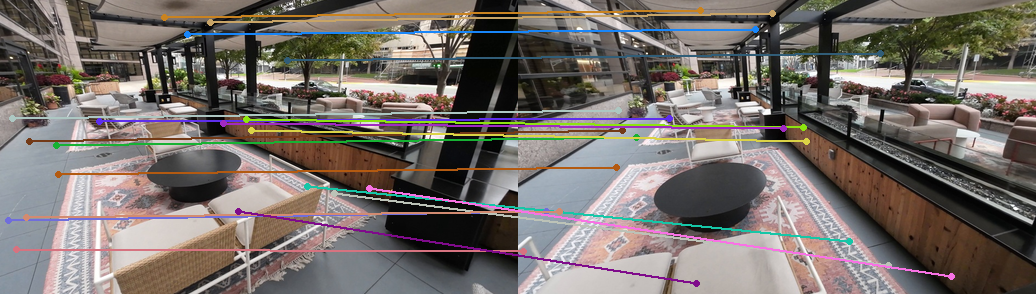} \\
    \includegraphics[width=0.5\textwidth]{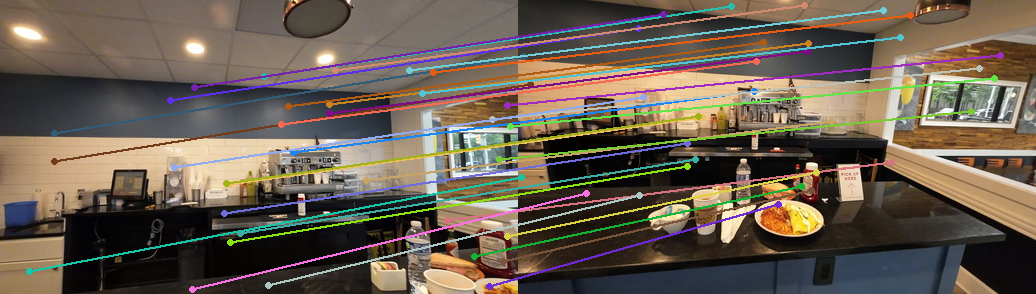} &
    \includegraphics[width=0.5\textwidth]{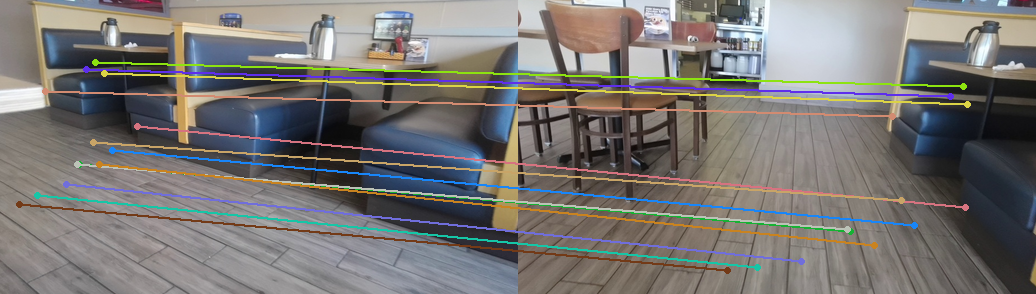} \\
    \includegraphics[width=0.5\textwidth]{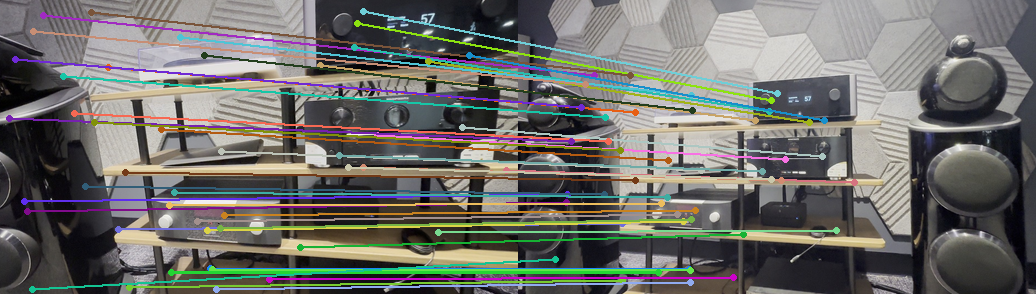} &
    \includegraphics[width=0.5\textwidth]{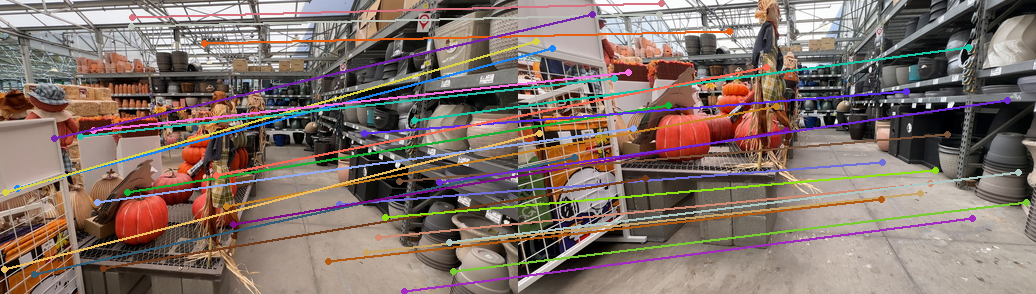} \\
    \includegraphics[width=0.5\textwidth]{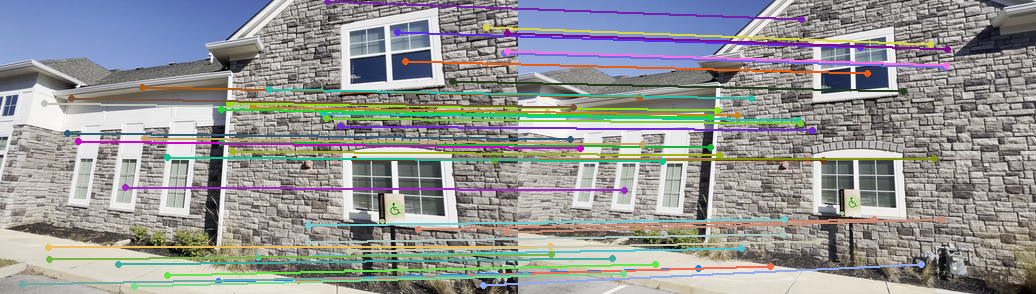} &
    \includegraphics[width=0.5\textwidth]{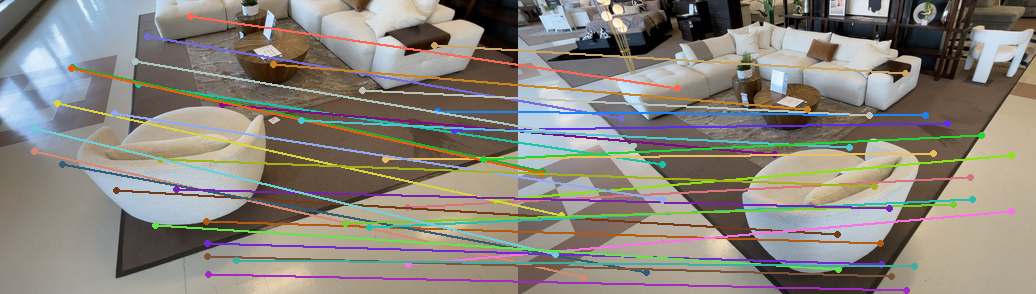} \\
    \includegraphics[width=0.5\textwidth]{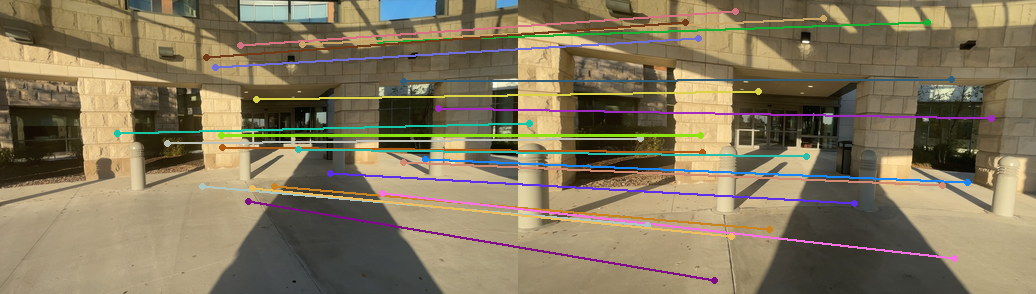} &
    \includegraphics[width=0.5\textwidth]{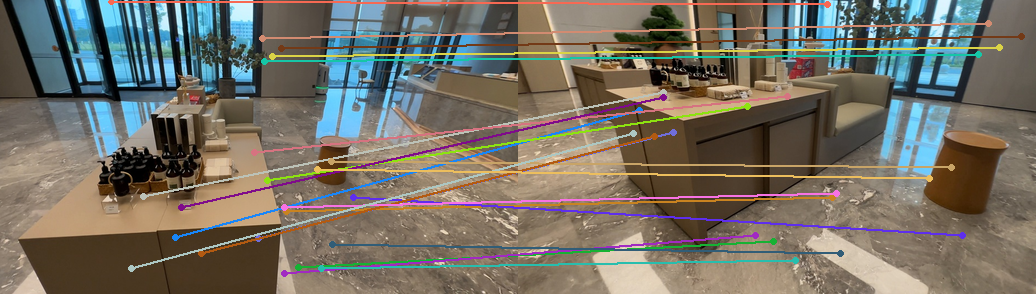} \\
    \includegraphics[width=0.5\textwidth]{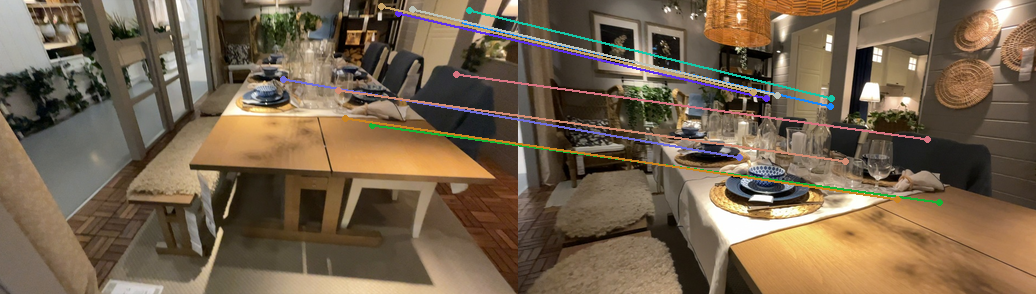} &
    \includegraphics[width=0.5\textwidth]{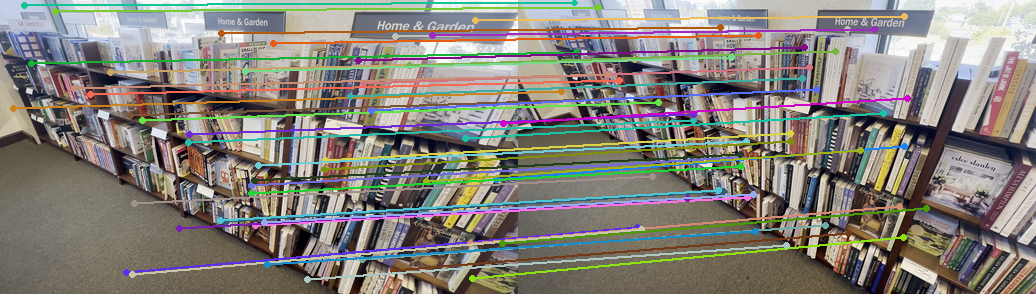} \\
    \includegraphics[width=0.5\textwidth]{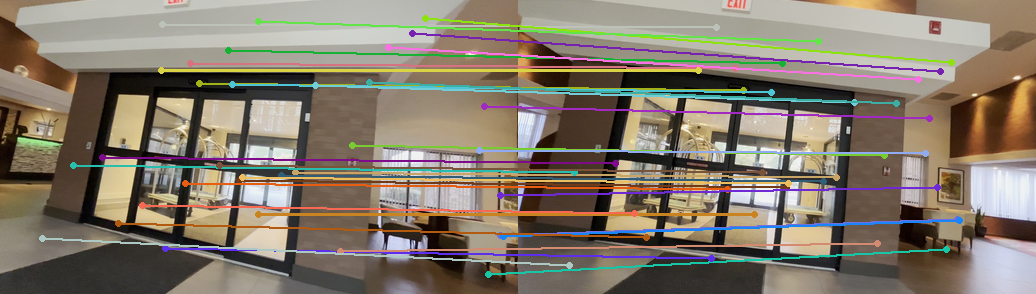} &
    \includegraphics[width=0.5\textwidth]{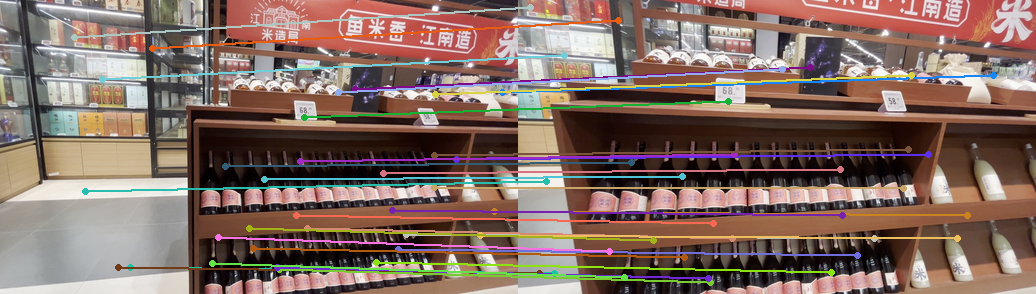}
    \end{tabular}
    }
    \caption{\textbf{Visualization of Matching on DL3DV}. Given query points in the left image, we visualize the matched pixels in the right image. Our matching is robust to large changes in camera viewpoint.}
    \label{fig:matching_dl3dv}
\end{figure*}

\myparagraph{Visualization of Feature Matching Results.}  We plot our computed correspondences from the aligned features in~\cref{fig:matching_dl3dv} and ~\cref{fig:matching_re10k}. Our feature matching successfully handles large changes in camera viewpoint.

\begin{figure*}[t]
    \centering
    \setlength{\tabcolsep}{1pt}
    \resizebox{\textwidth}{!}{
    \begin{tabular}{cccc}
    \includegraphics[width=0.25\textwidth]{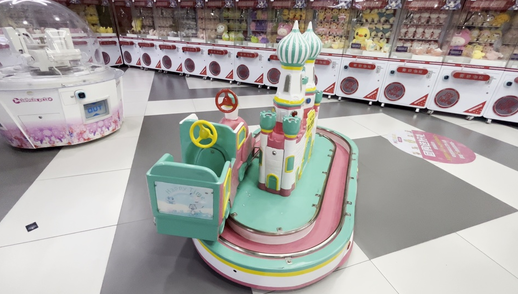} &
    \includegraphics[width=0.25\textwidth]{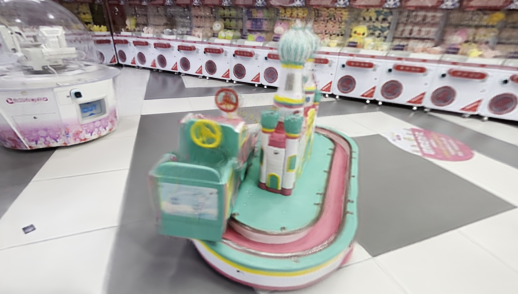} &
    \includegraphics[width=0.25\textwidth]{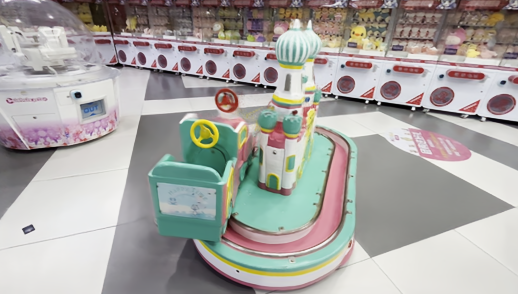} &
    \includegraphics[width=0.25\textwidth]{images/supp_qualitative_dl3dv/0a1b7c20a92c43c6b8954b1ac909fb2f0fa8b2997b80604bc8bbec80a1cb2da3_gt.png}\\
    \includegraphics[width=0.25\textwidth]{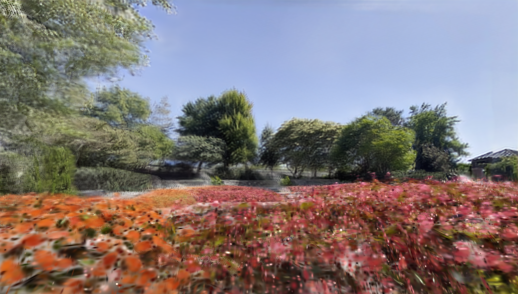} &
    \includegraphics[width=0.25\textwidth]{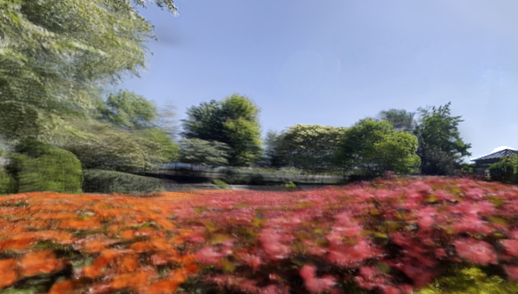} &
    \includegraphics[width=0.25\textwidth]{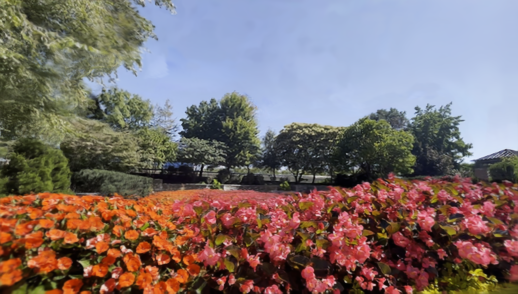} &
    \includegraphics[width=0.25\textwidth]{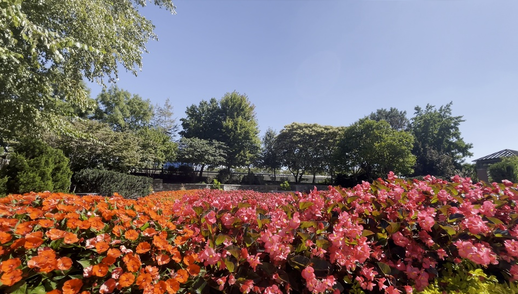}\\
    \includegraphics[width=0.25\textwidth]{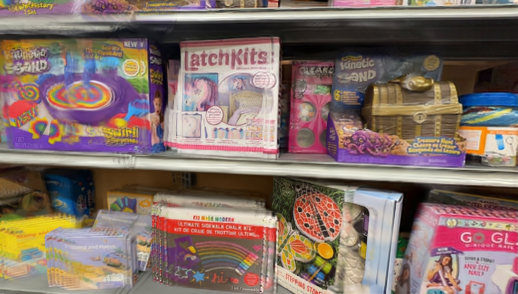} &
    \includegraphics[width=0.25\textwidth]{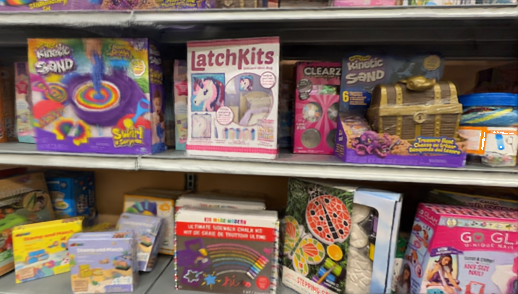} &
    \includegraphics[width=0.25\textwidth]{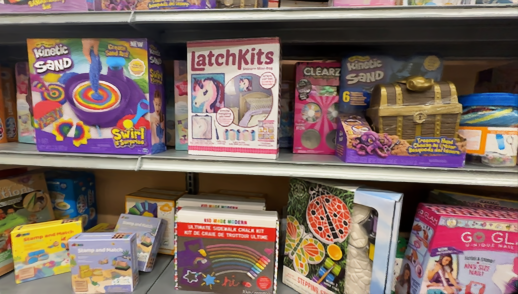} &
    \includegraphics[width=0.25\textwidth]{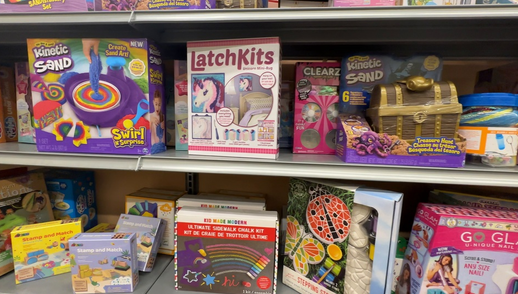}\\
    \includegraphics[width=0.25\textwidth]{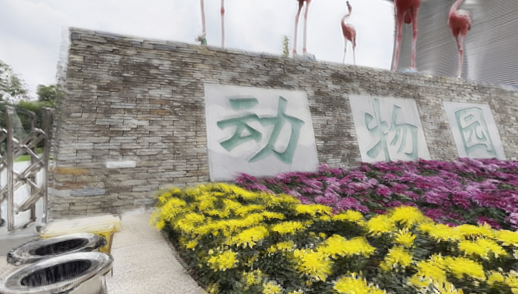} &
    \includegraphics[width=0.25\textwidth]{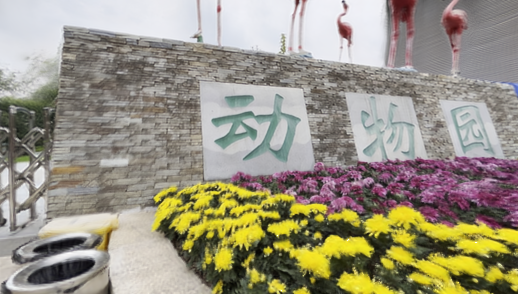} &
    \includegraphics[width=0.25\textwidth]{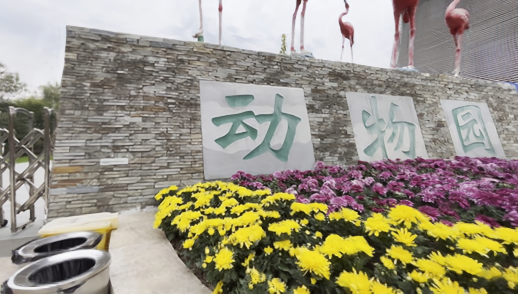} &
    \includegraphics[width=0.25\textwidth]{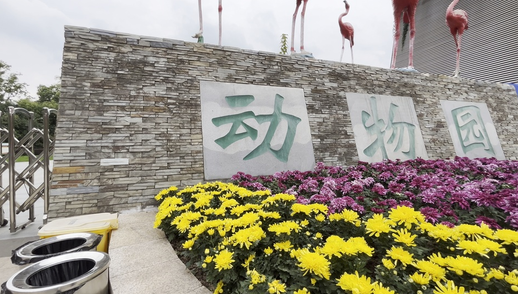}\\
    \includegraphics[width=0.25\textwidth]{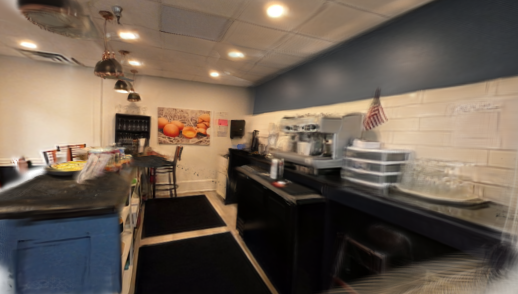} &
    \includegraphics[width=0.25\textwidth]{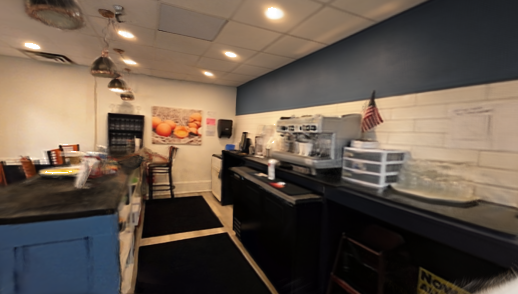} &
    \includegraphics[width=0.25\textwidth]{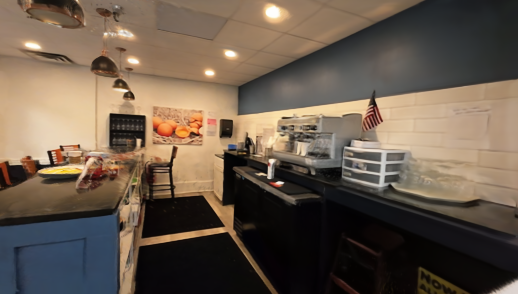} &
    \includegraphics[width=0.25\textwidth]{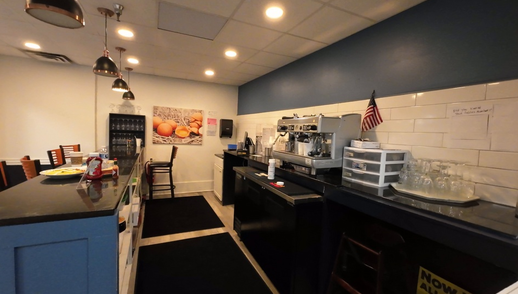}\\
    \includegraphics[width=0.25\textwidth]{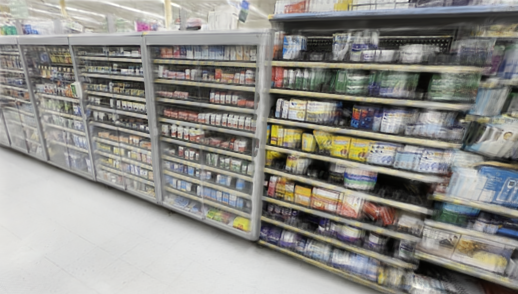} &
    \includegraphics[width=0.25\textwidth]{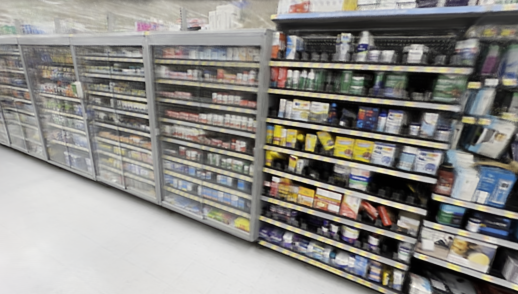} &
    \includegraphics[width=0.25\textwidth]{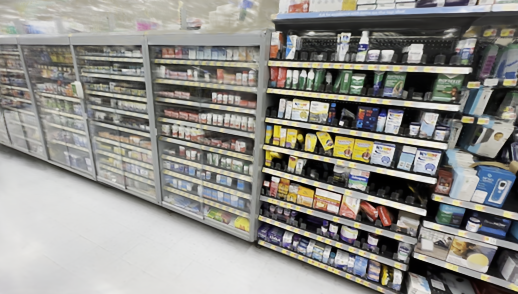} &
    \includegraphics[width=0.25\textwidth]{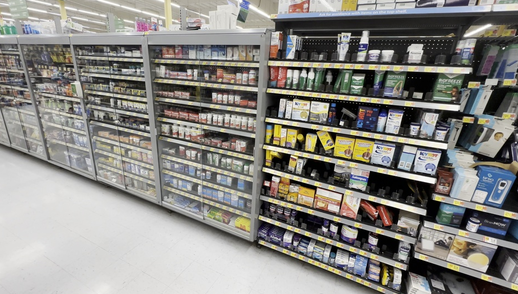}\\
    \includegraphics[width=0.25\textwidth]{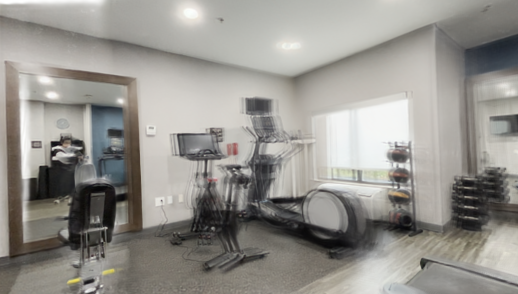} &
    \includegraphics[width=0.25\textwidth]{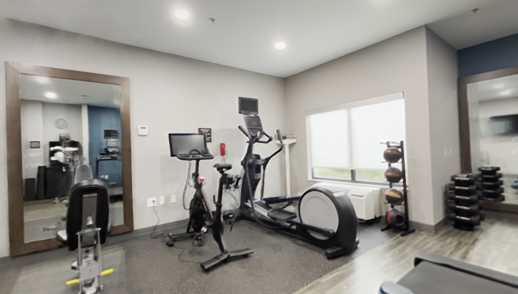} &
    \includegraphics[width=0.25\textwidth]{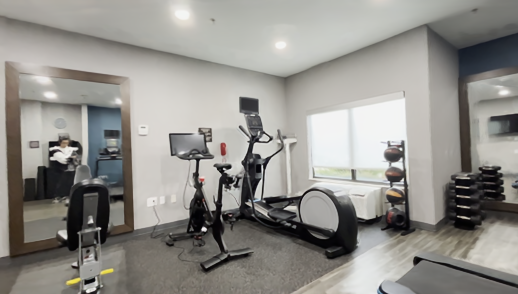} &
    \includegraphics[width=0.25\textwidth]{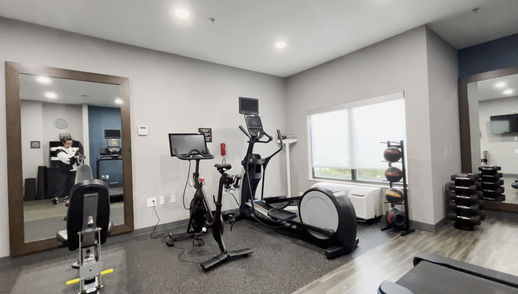}\\
    \includegraphics[width=0.25\textwidth]{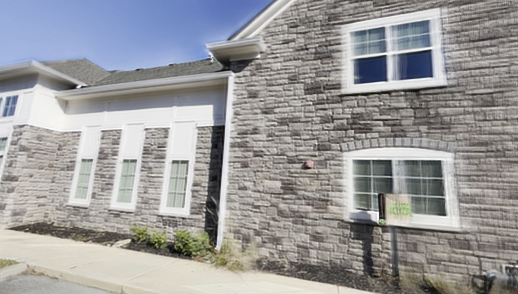} &
    \includegraphics[width=0.25\textwidth]{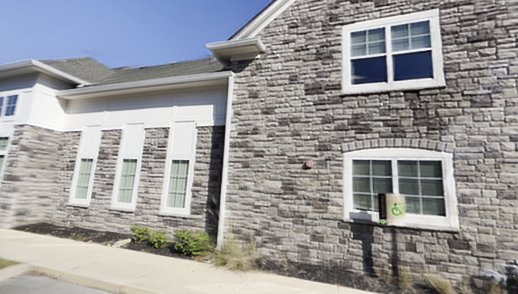} &
    \includegraphics[width=0.25\textwidth]{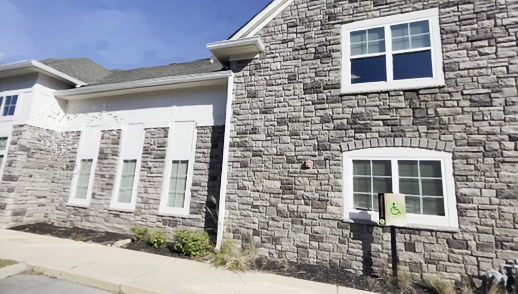} &
    \includegraphics[width=0.25\textwidth]{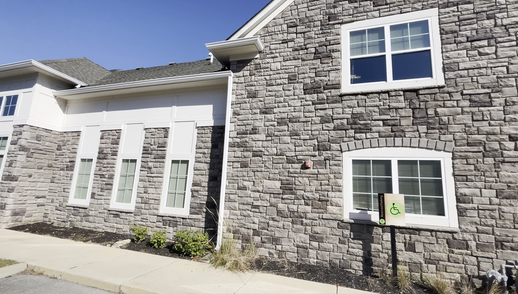}\\
    \multicolumn{1}{c}{(a) AnySplat} &
    \multicolumn{1}{c}{(b) WorldMirror} &
    \multicolumn{1}{c}{(c) \textbf{Ours}} &
    \multicolumn{1}{c}{(d) GT}\\
    \end{tabular}
    }
    \caption{\textbf{Qualitative Comparisons on DL3DV}. We visualize additional examples of renderings from AnySplat~\cite{jiang2025anysplat}, WorldMirror~\cite{liu2025worldmirror}, and our method on DL3DV.}
    \label{fig:supp_qualitative_dl3dv}
\end{figure*}

\begin{figure*}[t]
    \centering
    \setlength{\tabcolsep}{1pt}
    \resizebox{\textwidth}{!}{
    \begin{tabular}{cccc}
    \includegraphics[width=0.25\textwidth]{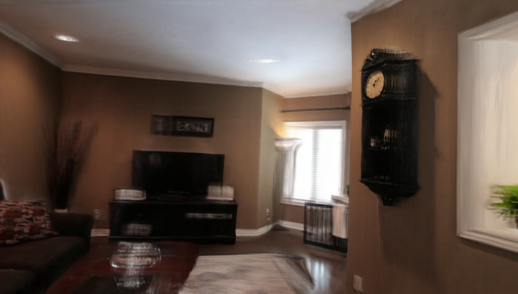} &
    \includegraphics[width=0.25\textwidth]{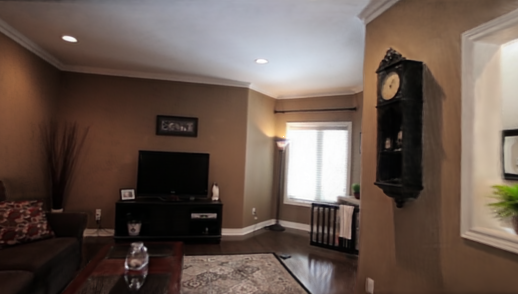} &
    \includegraphics[width=0.25\textwidth]{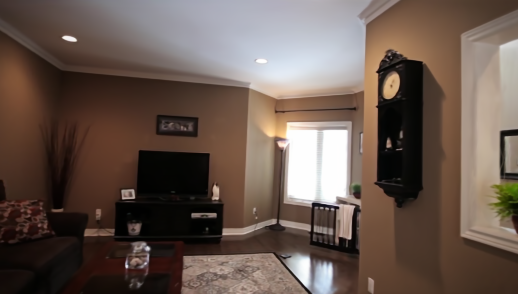} &
    \includegraphics[width=0.25\textwidth]{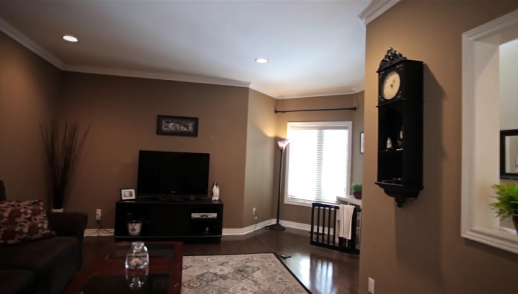}\\
    \includegraphics[width=0.25\textwidth]{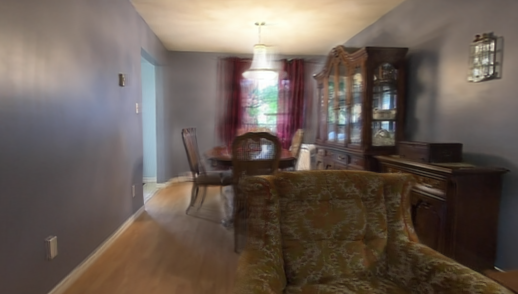} &
    \includegraphics[width=0.25\textwidth]{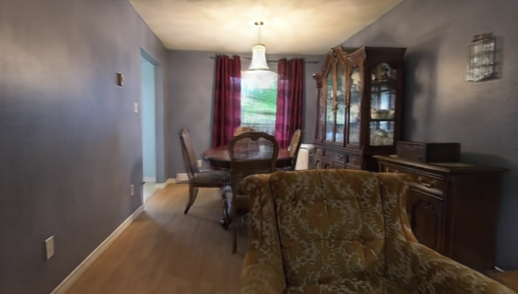} &
    \includegraphics[width=0.25\textwidth]{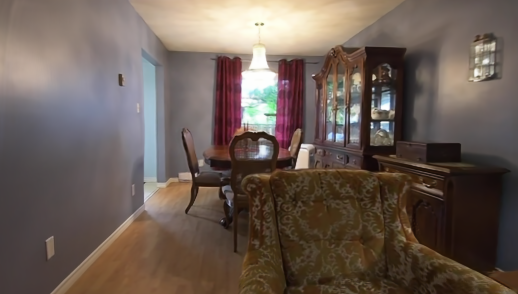} &
    \includegraphics[width=0.25\textwidth]{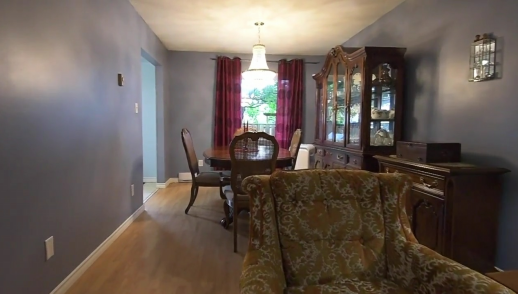}\\
    \includegraphics[width=0.25\textwidth]{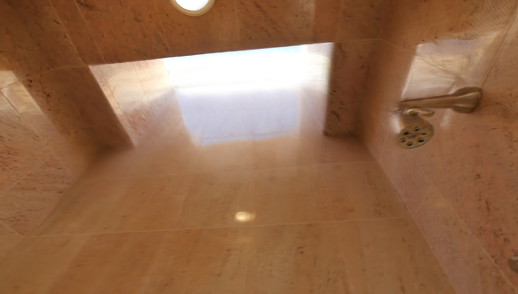} &
    \includegraphics[width=0.25\textwidth]{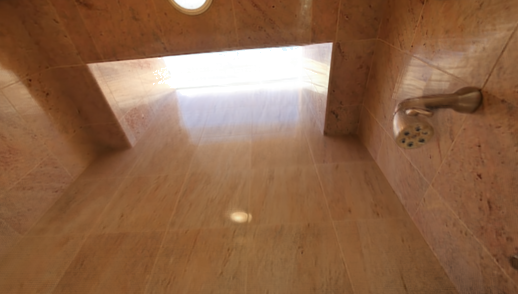} &
    \includegraphics[width=0.25\textwidth]{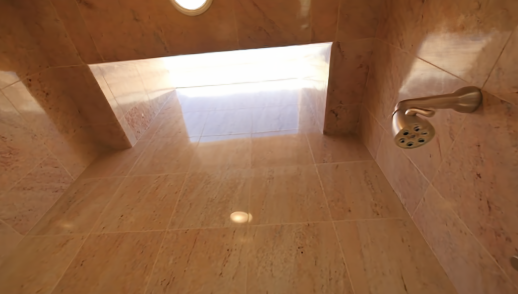} &
    \includegraphics[width=0.25\textwidth]{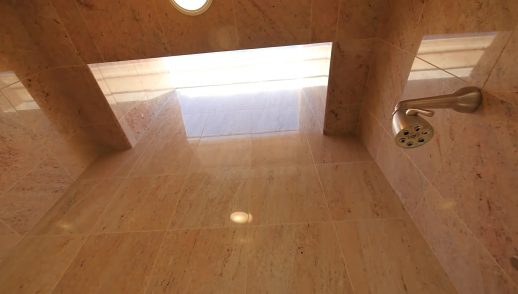}\\
    \includegraphics[width=0.25\textwidth]{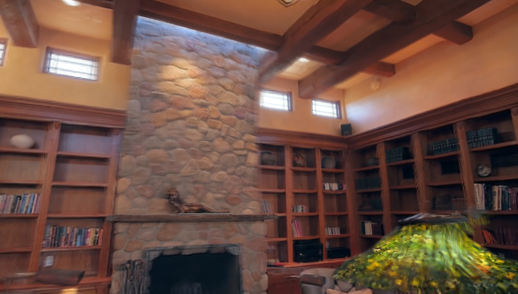} &
    \includegraphics[width=0.25\textwidth]{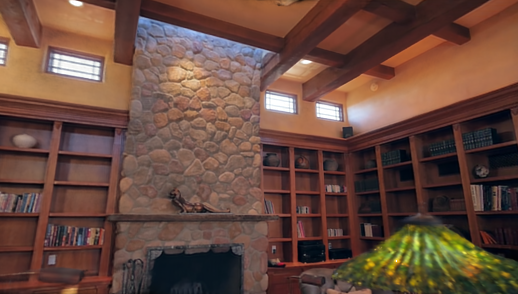} &
    \includegraphics[width=0.25\textwidth]{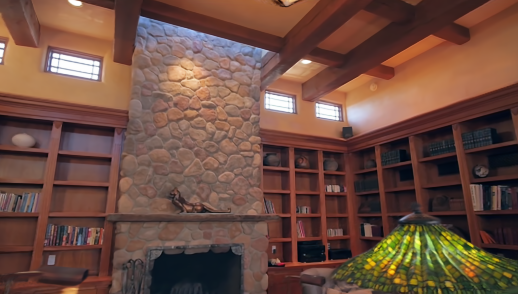} &
    \includegraphics[width=0.25\textwidth]{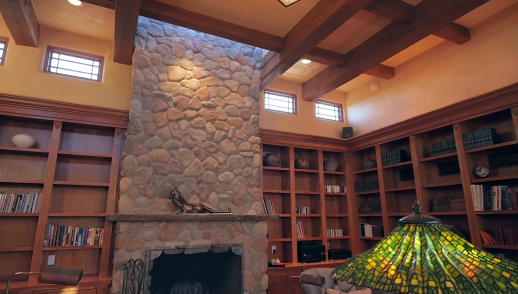}\\
    \includegraphics[width=0.25\textwidth]{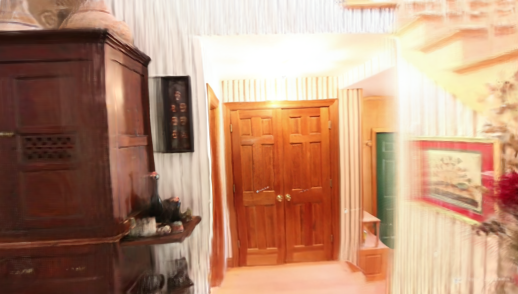} &
    \includegraphics[width=0.25\textwidth]{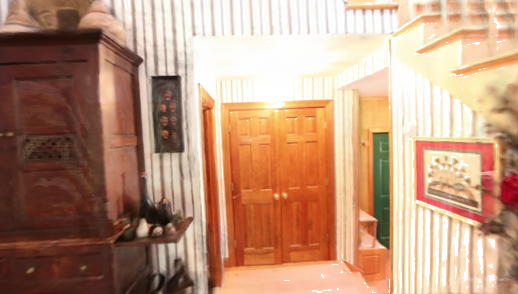} &
    \includegraphics[width=0.25\textwidth]{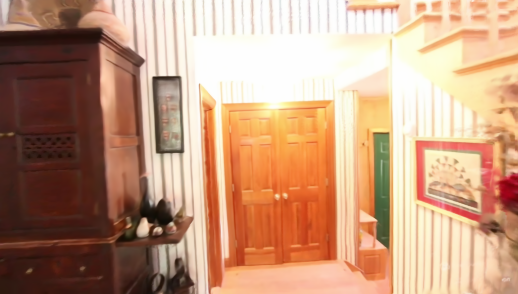} &
    \includegraphics[width=0.25\textwidth]{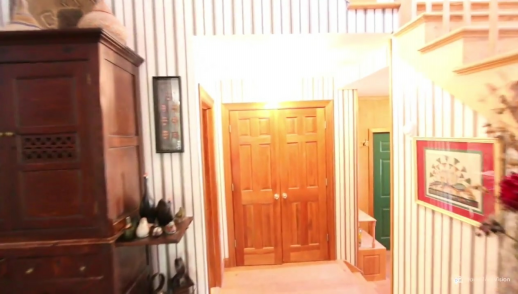}\\
    \includegraphics[width=0.25\textwidth]{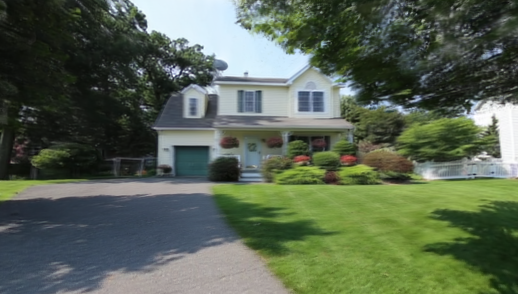} &
    \includegraphics[width=0.25\textwidth]{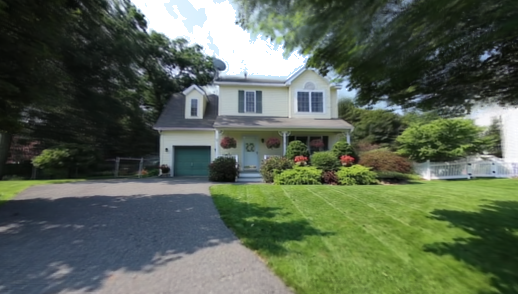} &
    \includegraphics[width=0.25\textwidth]{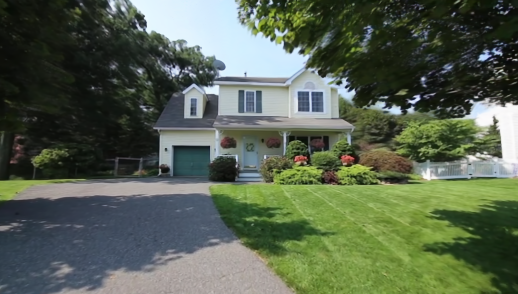} &
    \includegraphics[width=0.25\textwidth]{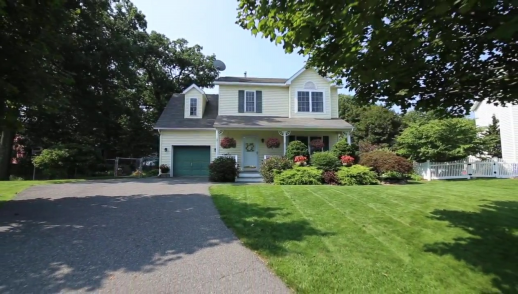}\\
    \includegraphics[width=0.25\textwidth]{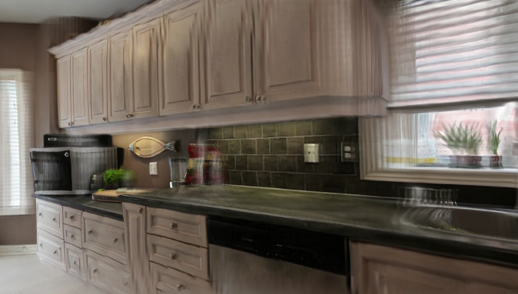} &
    \includegraphics[width=0.25\textwidth]{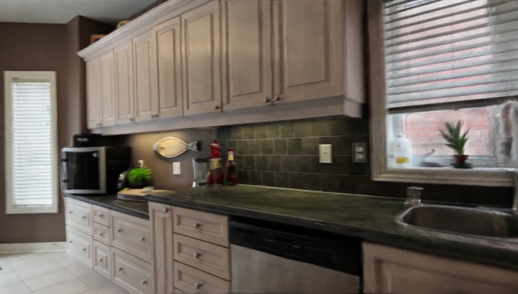} &
    \includegraphics[width=0.25\textwidth]{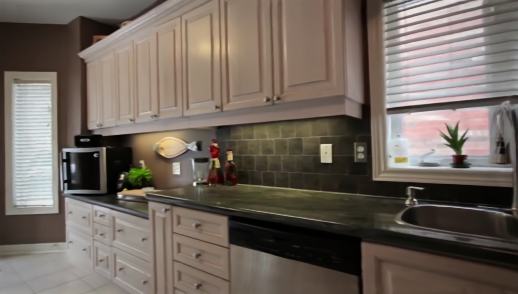} &
    \includegraphics[width=0.25\textwidth]{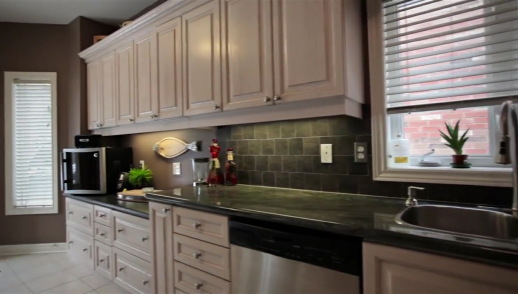}\\
    \includegraphics[width=0.25\textwidth]{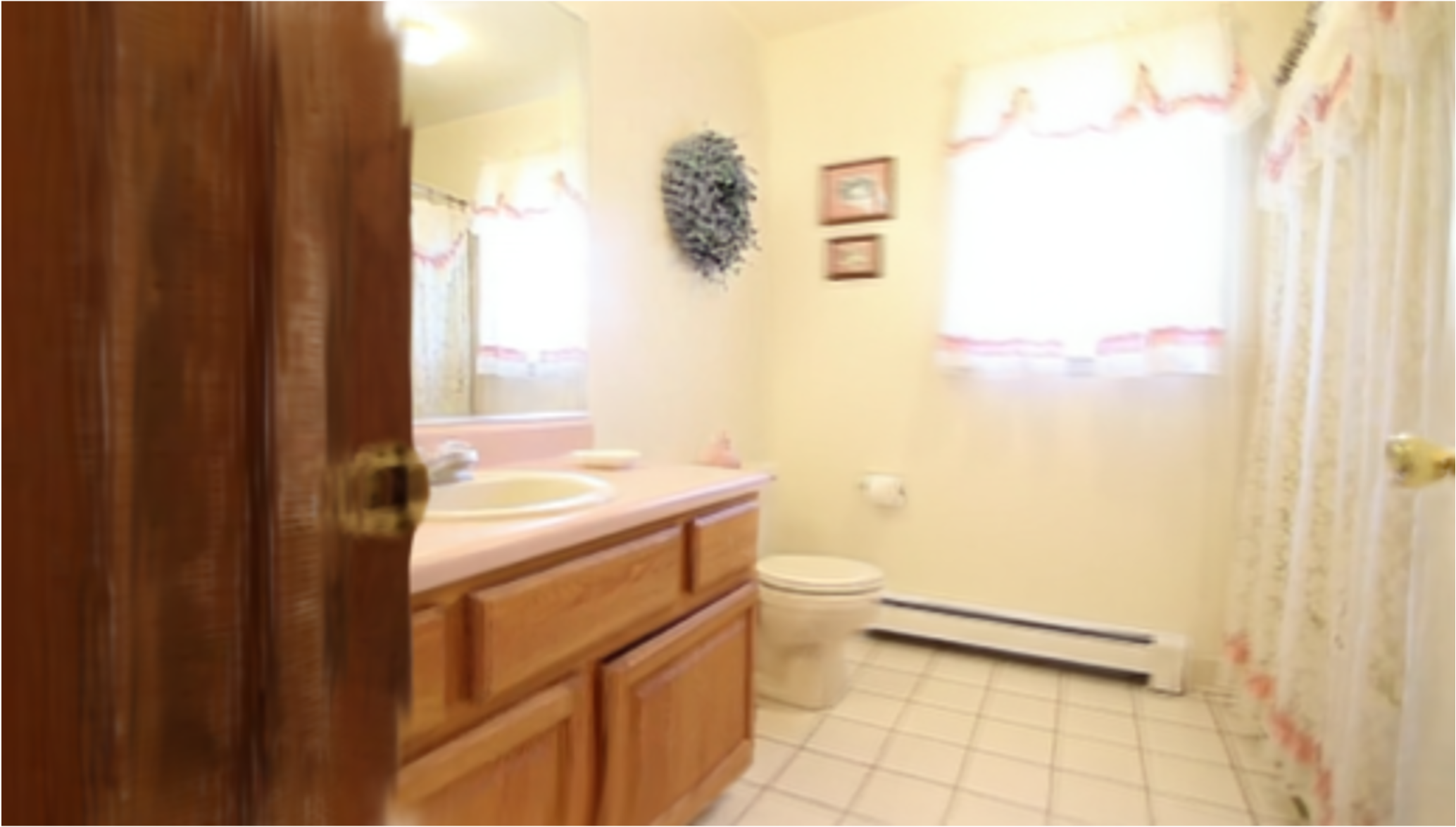} &
    \includegraphics[width=0.25\textwidth]{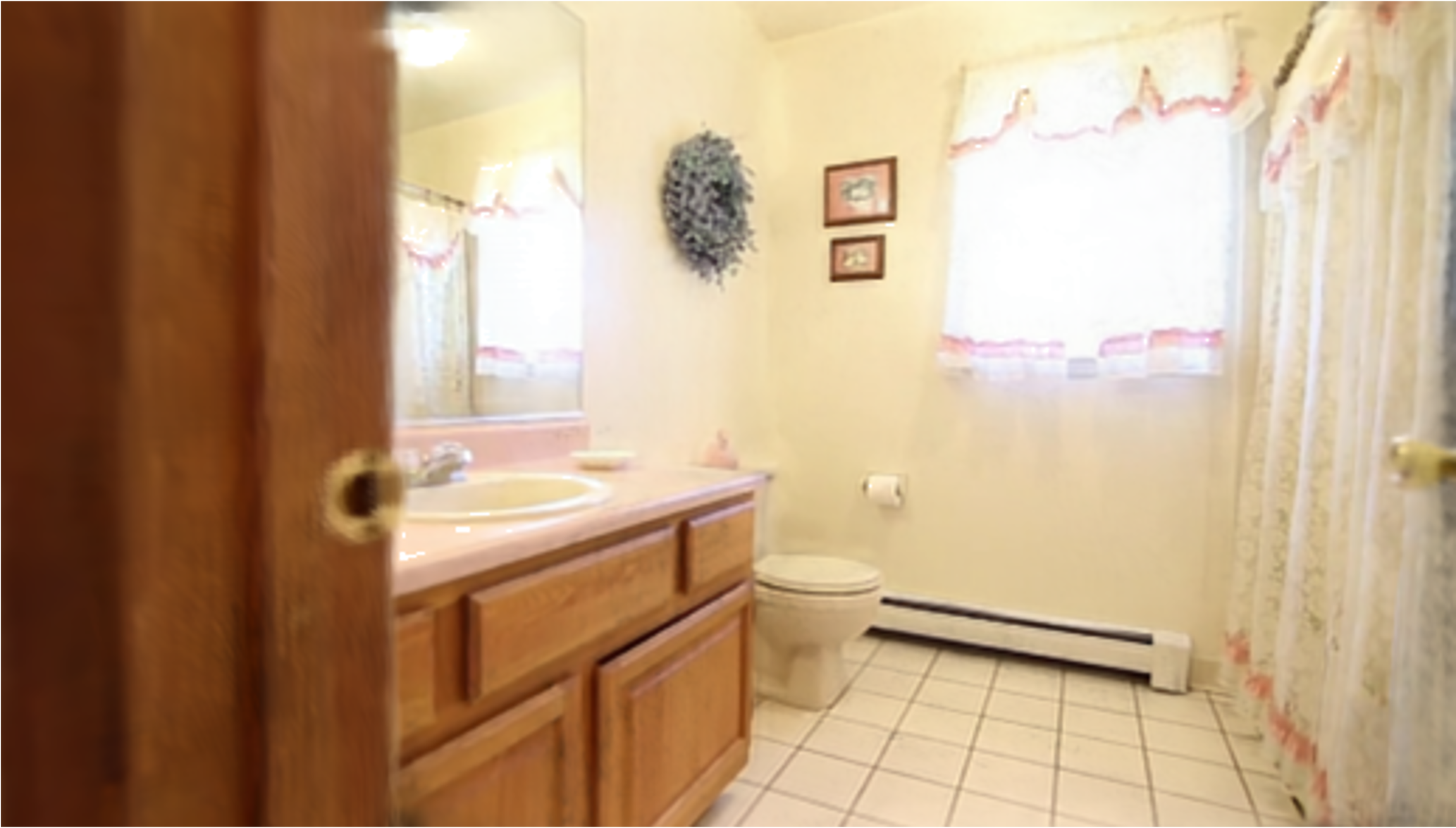} &
    \includegraphics[width=0.25\textwidth]{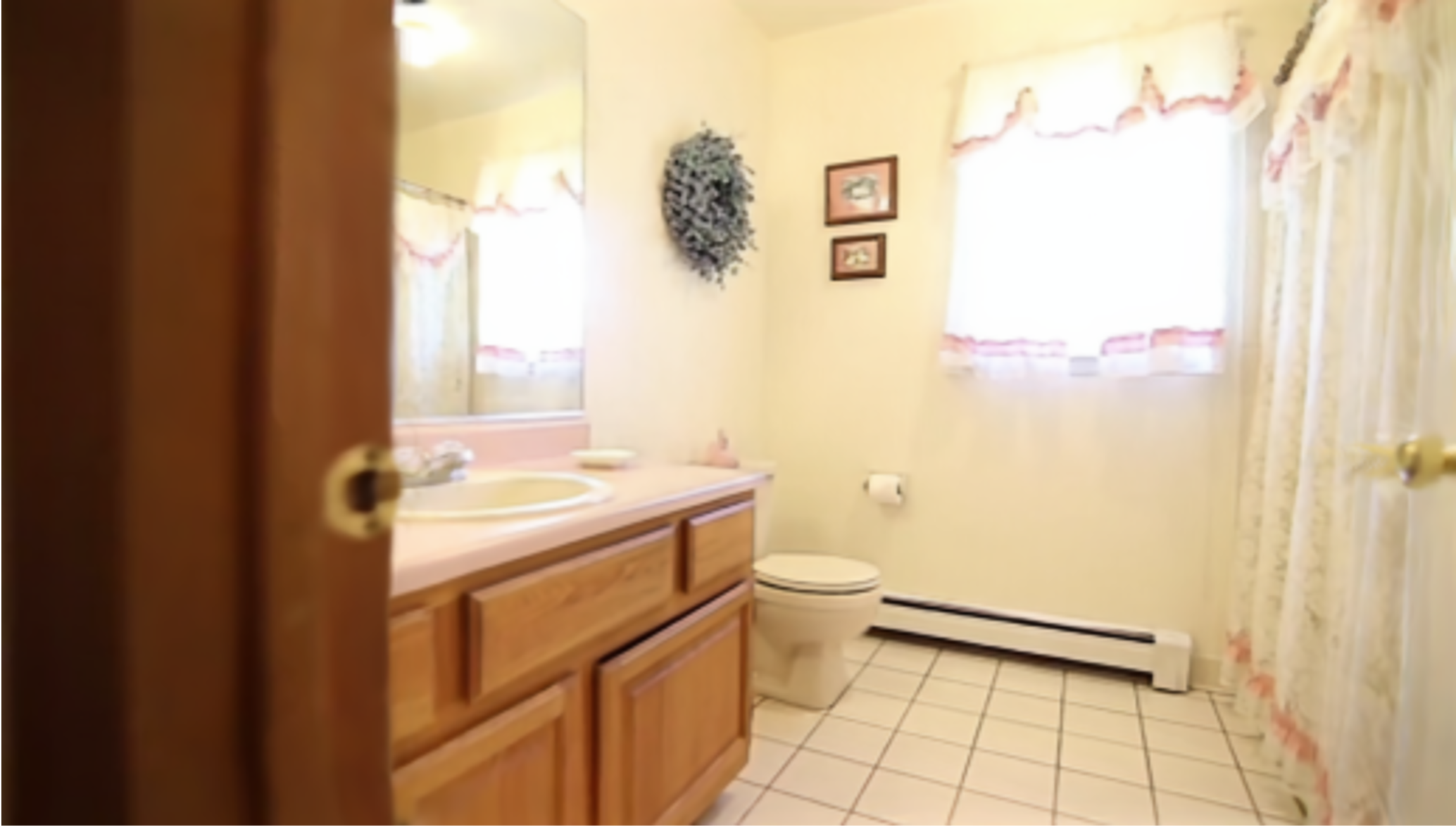} &
    \includegraphics[width=0.25\textwidth]{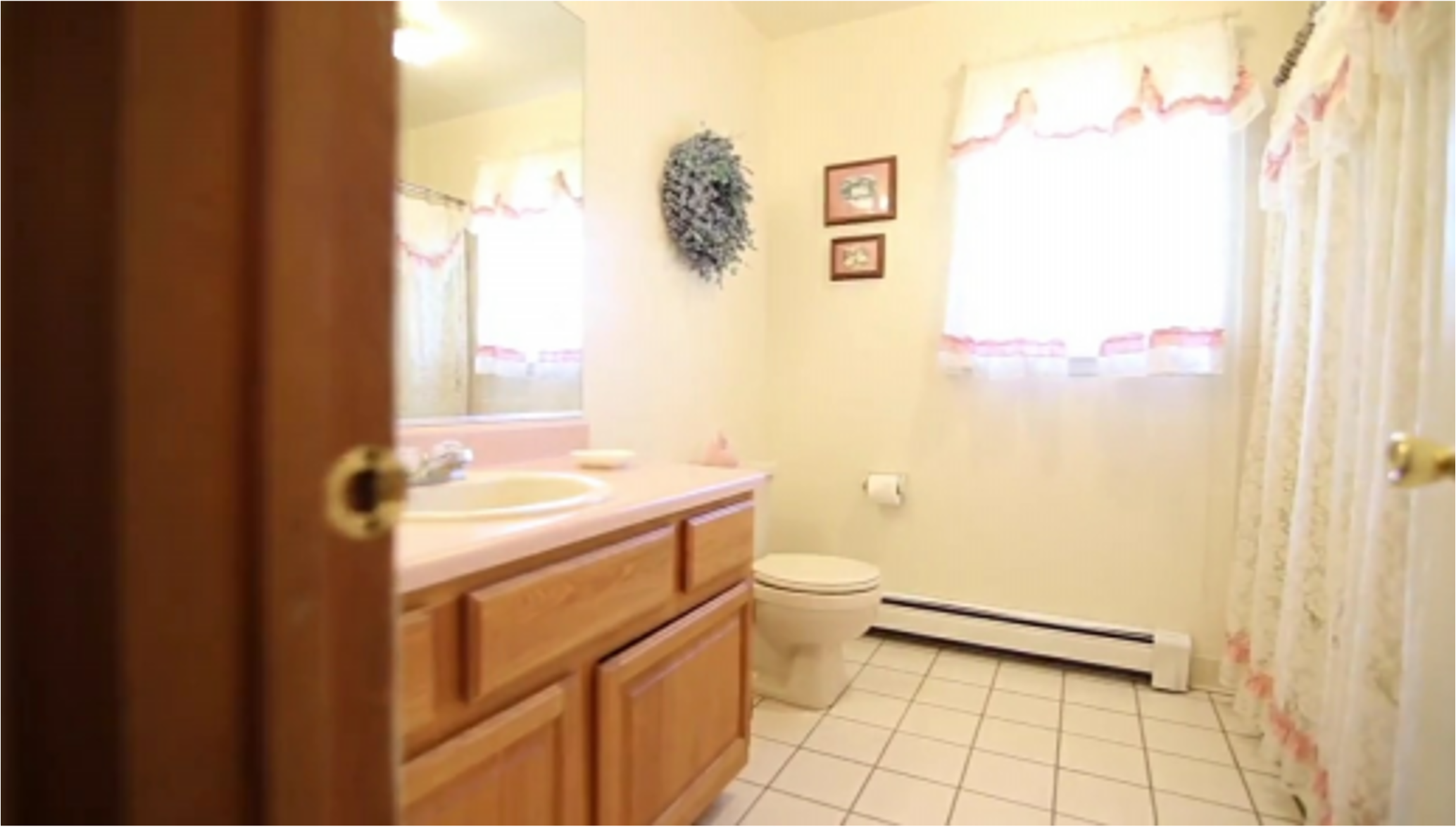}\\
    \multicolumn{1}{c}{(a) AnySplat} &
    \multicolumn{1}{c}{(b) WorldMirror} &
    \multicolumn{1}{c}{(c) \textbf{Ours}} &
    \multicolumn{1}{c}{(d) GT}\\
    \end{tabular}
    }
    \caption{\textbf{Qualitative Comparisons on RE10K}. We visualize additional examples of renderings from AnySplat~\cite{jiang2025anysplat}, WorldMirror~\cite{liu2025worldmirror}, and our method on RealEstate10K.}
    \label{fig:supp_qualitative_re10k}
\end{figure*}

\myparagraph{More Qualitative Comparisons.} We provide additional qualitative  comparisons of our model renderings compared to baselines AnySplat~\cite{jiang2025anysplat} and WorldMirror~\cite{liu2025worldmirror} in~\cref{fig:supp_qualitative_dl3dv} and~\cref{fig:supp_qualitative_re10k}. We include additional video examples on our webpage and recommend that reviewers check them.

\section{Limitations}\label{sec:supp_limitations}
While our experimental results are encouraging, there are limitations that remain. Firstly, because we rely on a pretrained VGGT backbone, We find that depth predictions in certain regions may not be accurate, for instance, in the sky and far-away regions. This is largely because the VGGT is trained with a normalized scale to eliminate ambiguity. As a result, a background with an extremely large depth difference from the foreground cannot be represented in a normalized scene field. While a view-dependent density term can mitigate this effect by reducing low-confidence Gaussians to be transparent when rendering to a novel view, the fundamental artifact stems from incorrect depth and pose estimates. In addition, we find that exposure differences between the input images and ground-truth targets can impact rendering metrics, as our produced rendering will mimic the exposure of the input images which may differ from that of the ground-truth target.
Our model can exhibit failures on dynamic scenes as both the VGGT and our feature alignment head train exclusively on static scenes; we observe that dynamic regions may be incorrectly matched with the parts of the static background they occlude, and improving the featuring matching on dynamic scenes is a promising future direction to investigate.

\clearpage
{
    \small
    \bibliographystyle{ieeenat_fullname}
    \bibliography{main}
}

\end{document}